\documentclass[conference]{IEEEtran}
\usepackage{times}

\usepackage[numbers]{natbib}
\usepackage{multicol}
\usepackage[bookmarks=true]{hyperref}
\usepackage{graphicx}
\usepackage{footnote} 
\usepackage{tabularx}
\usepackage{booktabs}
\usepackage{color}
\usepackage{subcaption}
\usepackage{xcolor}
\usepackage{bigstrut}
\usepackage{bbding}
\usepackage{amssymb}
\usepackage{pifont}
\usepackage{multirow}
\usepackage{graphicx}
\usepackage{amsmath}
\usepackage{cases}
\usepackage{bbding}
\usepackage{utfsym}
\usepackage{amsthm, thmtools}

\usepackage{caption}
\usepackage{multirow}
\usepackage{changes}
\definechangesauthor[name={Xiaohao Xu}, color=orange]{X.}
\definechangesauthor[name={Xiaohao Xu}, color=red]{Y.}

\begin{document}

\title{Customizable Perturbation Synthesis for \\ Robust SLAM Benchmarking}

\author{Xiaohao Xu,  Tianyi Zhang${}^{\dagger}$\thanks{$\dagger$ Equal contribution.}, Sibo Wang${}^{\dagger}$,  Xiang Li,  Yongqi Chen, Ye Li,   \\   Bhiksha Raj,   Matthew Johnson-Roberson,   Xiaonan Huang${}^{*}$\thanks{* Corresponding author.}}



%
\author{\authorblockN{Xiaohao Xu${}^{1}$,
Tianyi Zhang${}^{2}$\authorrefmark{2},
Sibo Wang${}^{1}$\authorrefmark{2}, 
Xiang Li${}^{3}$, 
Yongqi Chen${}^{1}$,
Ye Li${}^{1}$,
\\
Bhiksha Raj${}^{3}$, 
Matthew Johnson-Roberson${}^{2}$, and
Xiaonan Huang${}^{1}$\authorrefmark{1}}

\authorblockA{\authorrefmark{1}Corresponding author \quad \authorrefmark{2}Equal contribution}
\authorblockA{${}^{1}$ Department of Robotics,
University of Michigan,
Ann Arbor}
\authorblockA{${}^{2}$ Robotics Institute, Carnegie Mellon University}
\authorblockA{${}^{3}$ Department of Electrical and Computer Engineering, Carnegie Mellon University}
\authorblockA{Email: \{xiaohaox, xiaonanh\}@umich.edu}
}

\twocolumn[{%
\renewcommand\twocolumn[1][]{#1}%
\maketitle
\IEEEpeerreviewmaketitle
\begin{center}
    \centering 
    \includegraphics[width=\linewidth]{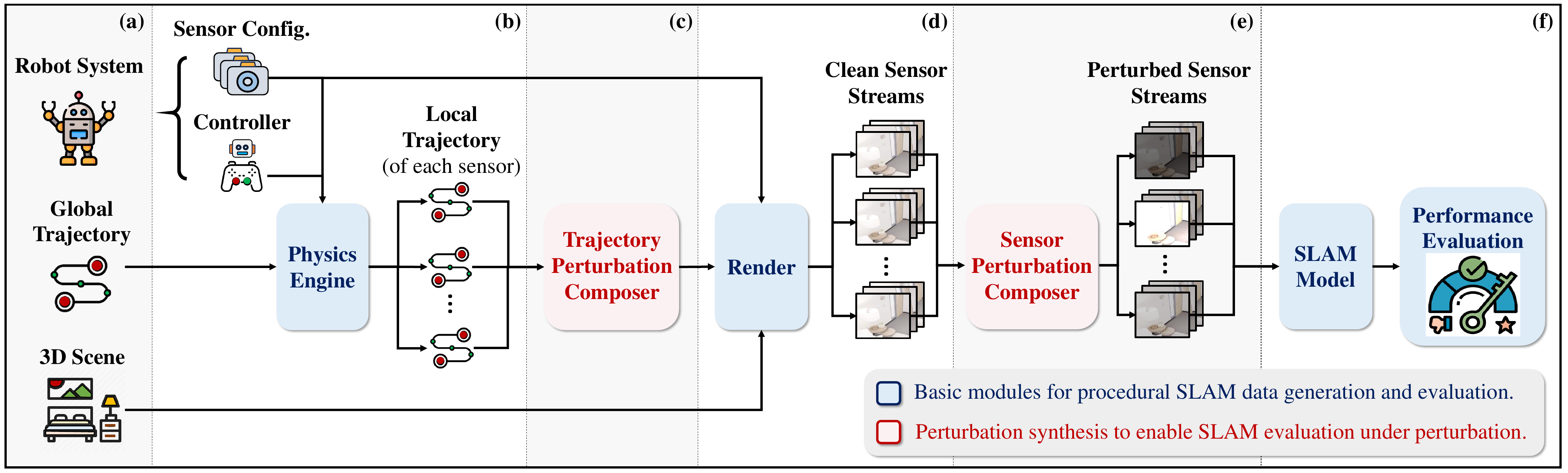}
    \captionof{figure}{\textbf{Overview of the noisy data synthesis pipeline for SLAM evaluation under perturbation.} (\textbf{a}) Given the customizable robot system and global trajectory, (\textbf{b}) the local trajectory of each sensor can be generated via the physics engine. (\textbf{c}) Subsequently,  the trajectory perturbation composer introduces deviations to simulate locomotion perturbations.  (\textbf{d}) Following this,  the render combines sensor configurations, perturbed local trajectories, and 3D scene models to generate sensor streams. (\textbf{e}) Finally,  the sensor perturbation composer introduces corruptions to the clean sensor streams, (\textbf{f}) resulting in perturbed data for  SLAM robustness benchmarking.}
    \label{fig:overview-simulator-pipeline}
\end{center}
\vspace{4mm}
}]

\begin{abstract}
Robustness is a crucial factor for the successful deployment of robots in unstructured environments, particularly in the domain of Simultaneous Localization and Mapping (SLAM).  
Simulation-based benchmarks have emerged as a highly scalable approach for robustness evaluation compared to real-world data collection.
However, crafting a challenging and controllable \textit{noisy world} with diverse perturbations remains relatively under-explored.
To this end, we propose a novel, customizable pipeline for noisy data synthesis, aimed at assessing the resilience of multi-modal SLAM models against various perturbations. This pipeline incorporates customizable hardware setups, software components, and perturbed environments. In particular, we introduce comprehensive perturbation taxonomy along with a perturbation composition toolbox, allowing the transformation of clean simulations into challenging noisy environments. Utilizing the pipeline, we instantiate the \textit{Robust-SLAM} benchmark, which includes diverse perturbation types, to evaluate the risk tolerance of existing advanced multi-modal SLAM models. Our extensive analysis uncovers the susceptibilities of existing SLAM models to real-world disturbance, despite their demonstrated accuracy in standard benchmarks. Our perturbation synthesis toolbox, SLAM robustness evaluation pipeline, and \textit{Robust-SLAM} benchmark will be made publicly available at https://github.com/Xiaohao-Xu/SLAM-under-Perturbation/.

\end{abstract}

\section{Introduction}
\label{sec:introduction}
\IEEEPARstart{T}he growing prevalence of mobile robots deployed in complex and dynamic environments~\cite{kaufmann2023champion,SubT}, \textit{i.e.}, \textit{noisy world}, underscores the critical need for robustness in robotic systems. This robustness, essential for effective operation, is significantly influenced by the robot's ability to withstand disruptive perturbations.
Consequently, robustness evaluation in such settings~\cite{RobustNav, antonante2023task} has emerged as a critical research area.
A key component in this domain is Simultaneous Localization and Mapping (SLAM)~\cite{slam-survey1,TNNLS-SLAM-survey}, which is fundamental to achieving robotic autonomy. The challenge, therefore, lies in developing a comprehensive and reliable benchmark to evaluate the robustness of SLAM systems against various disturbances.

Recent advances in SLAM system assessment have primarily focused on collecting challenging datasets. These datasets expose SLAM systems to domain-specific environmental degradation, broadening our understanding of real-world operational challenges~\cite{oxford-robotcar,Raincouver,Multi-Spectral,schubert2018tum,zhao2023subt}. However, scaling these datasets in size and diversity for holistic and comprehensive evaluation remains a challenge due to the inherent difficulties in data collection and labeling in the wild. 
In addition, the complex interplay of environmental factors makes isolating the impact of specific perturbations on SLAM performance a daunting task.
To overcome these limitations, simulation-based benchmarks~\cite{VI-navigation-with-simulation,dosovitskiy2017carla,RobustNav,tartanvo,driving-matrix} have emerged as a promising direction. 
They offer the advantage of creating infinite `battlefields' with greater data scalability and diversity for survival `testing' of SLAM models. 
Additionally, simulation-based benchmarks allow for adaptive crafting of more customizable and challenging scenarios, 
enabling the continuous advancement of more robust SLAM systems~\cite{tartanvo}. 
While current simulators may not fully replicate real-world fidelity, the rapid advancements of visual content synthesis techniques~\cite{rombach2022high,raistrick2023infinite} are progressively bridging this gap.

Despite the increasing availability of (nearly) photo-realistic 3D scene datasets and simulators ~\cite{straub2019replica,dai2017scannet,deitke2022️,zheng2020structured3d} for SLAM evaluation, they often lack varied and controllable disturbances. As a result, these simulations typically represent idealized, perturbation-free environments, \textit{i.e.}, \textit{{perfect world}}, leaving the simulated perturbed environment, \textit{i.e.}, \textit{{noisy world}}, largely unexplored.
To address this gap and enhance SLAM evaluation under perturbations, we propose a comprehensive taxonomy of perturbations in multi-modal SLAM. This encompasses 16 image corruption types, 4 depth corruption types, 3 trajectory perturbation types, and the multi-sensor misalignment effect. Based on the taxonomy, we build a perturbation composition toolbox that can be seamlessly integrated with existing simulation tools~\cite{denninger2020blenderproc,deitke2022️,straub2019replica} to transform the simulated environment from a \textit{{perfect world}} into a more challenging \textit{{noisy world}} for robustness evaluation. Fig.~\ref{fig:overview-simulator-pipeline} illustrates our customizable noisy data synthesis pipeline, designed to assess the resilience of SLAM under varied perturbations. This pipeline is adaptable to different robot hardware configurations (e.g., sensor placement) and software components (e.g., SLAM model). Moreover, it incorporates both trajectory-level and sensor-level perturbations, each with varying severity levels, to simulate robot locomotion and real-world environmental disturbances.

Utilizing the established perturbation taxonomy and the pipeline of noisy data synthesis, we instantiate a large-scale {{{\textit{Robust-SLAM}}}} benchmark. This benchmark aims to assess the performance of existing SLAM models under a spectrum of perturbations, offering insights into their strengths and weaknesses.
The benchmark includes both classical and neural models, evaluated under single-modal (monocular) and multi-modal (RGBD) input settings.
Our findings indicate
that while existing advanced SLAM models 
excel in standard clean SLAM benchmarks~\cite{TUM-RGBD,straub2019replica}, they exhibit vulnerabilities and a propensity for failure when exposed to sensor-related or trajectory-related perturbations.

Briefly put, the main contributions of this work are that it:
\begin{itemize}
\item proposes a comprehensive taxonomy of perturbations for SLAM in dynamic and unstructured environments.

\item introduces a noisy data synthesis pipeline, utilizing the perturbation composition toolbox, thereby enabling customizable SLAM robustness assessment.

\item creates a large-scale dataset called Robust-SLAM, which includes diverse perturbation types. This dataset serves to evaluate the risk tolerance of SLAM in the presence of various sensor corruptions and dynamic motion.

\item systematically studies and analyzes the robustness of existing SLAM models under isolated perturbation types.

\end{itemize}

\vspace{0.5mm}
The subsequent sections are organized as follows: Sec. \ref{sec:related_work} reviews related work on robustness in SLAM. Sec. \ref{sec:problem_statement} presents the assumptions and formulations in this paper. Sec. \ref{sec:method} introduces the pipeline for customized perturbed SLAM benchmark synthesis and detailed perturbation taxonomies. In Sec. \ref{sec:experiment}, we instantiate a robustness benchmark using our pipeline and assess the resilience of current SLAM models against various perturbations. Finally, Sec. \ref{sec:conclusions} summarizes the study and discusses future research directions.

\section{Related Work}\label{sec:related_work}
This section provides an overview of SLAM methodologies, followed by a discussion on studies addressing corruption robustness in perception tasks, including SLAM.

\subsection{{SLAM Methods}}
This overview highlights visual-related SLAM systems. More comprehensive reviews of SLAM systems can be obtained from various resources such as \cite{slam-survey1,macario2022comprehensive,kazerouni2022survey}. Classical single-modal SLAM methods, like visual-only models exemplified by ORB-SLAM~\cite{orbslam}, have demonstrated remarkable accuracy in `clean' benchmark settings, such as TUM VI~\cite{schubert2018tum} and Replica~\cite{straub2019replica}.
To address the complexities of real-world environments, researchers have explored various techniques \cite{orbslam2,orbslam3,MIMOSA,rosinol2021kimera} that incorporate multi-view sensors and fuse diverse data modalities, such as visual-inertia and RGBD. Additionally, the development of multi-agent SLAM models \cite{Kimera-Multi,covins-g,D2SLAM} has facilitated collaborative localization and mapping among heterogeneous robots, showcasing improved robustness in navigation.
Furthermore, several approaches \cite{teed2021droid,tartanvo,coslam,imap,niceslam,zhang2023goslam,rosinol2022nerf} have utilized neural networks and neural representations to enhance generalization capabilities and improve the quality of 3D map reconstruction. While these models have achieved notable improvements in accuracy, their robustness of these models against sensor corruptions and motion perturbations remains relatively under-explored.

\subsection{{Robustness Benchmark}}
To ensure the reliable deployment of mobile robots, their perception modules must demonstrate resilience to shifts in natural distributions~\cite{zhang2017resilient}. A pioneering benchmark, ImageNet-C \cite{hendrycks2019robustness}, has emerged for systematically studying image corruption robustness by evaluating the performance of image classification methods against common corruptions and perturbations. Building upon this, subsequent research has expanded the scope of investigation to encompass other perception tasks. These tasks include object detection \cite{michaelis2019benchmarking,carlson2018modeling,kong2023robo3d}, segmentation \cite{kamann2020benchmarking,xu2022towards,li2023robust}, and embodied navigation \cite{RobustNav,yokoyama2022benchmarking}.
These studies underscore the significance of evaluating models' robustness to corruptions. In SLAM, the challenges extend beyond just handling image-level corruptions, like those due to camera malfunctions. It is also crucial to account for dynamic variations in sensor corruption and deviations in sensor transformation simultaneously over time. These variations arise from time-variant environmental effects and the diverse motion of robots, respectively. 
In this study, we propose a comprehensive taxonomy of perturbations for SLAM in dynamic environments (e.g., varying illumination) and unstructured environments (e.g., uneven terrains that can cause locomotion deviations for mobile robots).

\subsection{{Robustness Evaluation for SLAM}}
The robustness of SLAM systems is essential for their reliable and accurate operation in dynamic and challenging real-world environments~\cite{slam-survey1}. This robustness is critical not only to handle sensor faults, but also to ensure long-term performance.
To facilitate the robustness evaluation of SLAM models, several datasets \cite{schubert2018tum,helmberger2022hilti,tian2023resilient,SubT,zhao2023subt}  have been collected in degraded environments with perturbations such as low illumination or motion blur.  
Furthermore, SLAMBench \cite{bujanca2021robust} compares the performance of several classical SLAM models across multiple challenging datasets and reveals the vulnerability of these SLAM models. 
Considering that constructing real-world datasets via robot platforms for SLAM can be challenging and unscalable, Wang \textit{et al.} \cite{wang2020tartanair} have utilized photo-realistic simulation environments to create a pioneering simulated SLAM benchmark called TartanAir for robustness evaluation. In this study, we expand the scope of evaluation to include the robustness of multi-modal SLAM models—encompassing both classical and neural SLAM methods—against a broader spectrum of sensor corruptions and motion patterns (\textit{e.g.}, varying speed and trajectory deviations).

\section{Problem Statement}\label{sec:problem_statement}

\subsection{{Assumptions}}

In this work, we develop a comprehensive taxonomy of perturbations in SLAM and conduct robustness analyses of advanced SLAM models under the following assumptions.

\vspace{1.0mm}\noindent\textbf{Task.} We focus on the standard (passive) SLAM setting, assuming the absence of active decision-making  processes.

\vspace{1.0mm}\noindent\textbf{Model.} Our analysis is centered on vision-oriented SLAM scenarios, specifically targeting monocular and RGBD settings. We assume the use of dense depth representation as opposed to sparse depth data obtained from a LiDAR scanner. In addition, the SLAM system is presumed to have known motion and observation models. 

\vspace{1.0mm}\noindent\textbf{Perturbation.} 
Although our noisy data synthesis pipeline is capable of generating SLAM benchmarks with multiple heterogeneous perturbations, we deliberately concentrate on investigating the performance degradation caused by individual sensor or trajectory perturbations. This focused approach is designed to dissect the system's response to isolated perturbations, allowing precise quantification of their specific impacts on SLAM performance. By analyzing the degradation induced by individual perturbations, we can effectively assess the system's robustness in a controlled manner and identify the root causes of performance degradation. This knowledge is crucial for developing targeted mitigation strategies that address the most vulnerable aspects, \textit{i.e.}, \textit{Achilles' Heel}, of the whole SLAM system.
Also, we model these perturbations using simplified linear models (\textit{e.g.}, Gaussian noise assumptions), in line with precedent set by established literature~\cite{hendrycks2019robustness,wang2020tartanair,RobustNav}. While these simplified perturbations may not fully capture the complexity of real-world scenarios, they offer interpretability and facilitate analysis across different perturbation types.

\vspace{1.0mm}\noindent\textbf{3D scene}. We assume that the environment is static, meaning there are no moving or dynamically changing objects within the scene. Also, the scene is bounded, typically referring to an indoor setting with predefined boundaries or limits.

\subsection{{Formulation of Standard SLAM}}
SLAM aims to concurrently construct a map of the scene and estimate the pose (position and orientation) of a robot~\cite{probabilistic_robot}. Formally, given a sequence of observations $\mathbf{z}_{1:t}$ and control inputs or motions $\mathbf{u}_{1:t}$ from timestamp $1$ to $t$, the goal is to estimate the environmental map $\mathbf{m}$ (e.g., 3D occupancy or point cloud) and the robot's trajectory $\mathbf{x}_{1:t}$ over time. Specifically, the robot's pose at a specific timestamp $i$ ($1\leq i \leq t$) can be denoted in terms of spatial coordinates ($x_i, y_i, z_i$) and quaternion orientation ($\mathbf{q}_{w_i}, \mathbf{q}_{x_i}, \mathbf{q}_{y_i}, \mathbf{q}_{z_i}$), as formulated by
\begin{equation}
\mathbf{x}_i = [ x_i \quad y_i \quad z_i \quad \mathbf{q}_{w_i} \quad \mathbf{q}_{x_i} \quad \mathbf{q}_{y_i} \quad \mathbf{q}_{z_i} ]
\label{eq:pose_representation}
\end{equation}

Then, the probabilistic posterior articulates our belief regarding the map and trajectory based on the provided observations and motions, expressed as
\begin{equation}
p(\mathbf{m}, \mathbf{x}_{1:t} | \mathbf{z}_{1:t}, \mathbf{u}_{1:t})
\label{eq:posterior}
\end{equation}

\vspace{1.0mm}\subsection{{Formulation of SLAM under Perturbations}}

Real-world applications often encounter disturbances or perturbations which can affect both the observations taken by the robot's sensors and the robot's trajectory. In the following paragraphs, we illustrate the high-level formulations of SLAM under these perturbations.

\vspace{1.0mm}\noindent{\textbf{Sensor-level perturbation.}}
These corruptions introduce noise to the robot's sensor readings. Representing these perturbations with $\delta_{\mathbf{z}}$, the altered observation model is
\begin{equation}
\mathbf{z}_t' = \mathbf{z}_t + \delta_{\mathbf{z}}(\mathbf{z}_t)
\label{eq:img_perturb}
\end{equation}
With these sensor  corruptions, the perturbed SLAM problem can be formulated as
\begin{equation}
p(\mathbf{m}, \mathbf{x}_{1:t} | \mathbf{z}_{1:t}', \mathbf{u}_{1:t})
\label{eq:slam_img_perturb}
\end{equation}

\vspace{1.0mm}\noindent{\textbf{Trajectory-level perturbations.}}
These deviations affect the estimated pose of the robot. For a pose at time \( t \), represented as \( \mathbf{x}_t = [\mathbf{R}_t, \mathbf{t}_t] \) (with \( \mathbf{R}_t \) being a rotation matrix and \( \mathbf{t}_t \) a translation vector), we can introduce perturbations \( \delta_{\mathbf{R}} \) and \( \delta_{\mathbf{t}} \):
\begin{equation}
\mathbf{x}_t' = [\mathbf{R}_t + \delta_{\mathbf{R}}, \mathbf{t}_t + \delta_{\mathbf{t}}]
\label{eq:traj_perturb}
\end{equation}
With these deviations, the SLAM problem becomes
\begin{equation}
p(\mathbf{m}, \mathbf{x}_{1:t}' | \mathbf{z}_{1:t}, \mathbf{u}_{1:t})
\label{eq:slam_traj_perturb}
\end{equation}

\vspace{2mm}
\section{Customizable Data Synthesis}\label{sec:method}

\subsection{{Pipeline Overview}}

Fig.~\ref{fig:overview-simulator-pipeline} shows the proposed noisy data synthesis pipeline for SLAM robustness benchmarking. The pipeline is highly customizable and incorporates controllable perturbations to simulate sensor noises and locomotion disturbances.

 The initial phase is to configure the robot system, the desired trajectory, and the environmental setup (see Fig.~\ref{fig:overview-simulator-pipeline}a). We can optionally utilize off-the-shelf physics engines (such as MuJoCo~\cite{mujoco}) in conjunction with motion controllers to obtain sensor-specific local poses (see Fig.~\ref{fig:overview-simulator-pipeline}b).  Subsequently, these clean trajectories can be passed to the trajectory perturbation composer to introduce deviations to the pose, thereby better emulating pose inaccuracy in the wild (see Fig.~\ref{fig:overview-simulator-pipeline}c). The render, implemented via the OpenGL~\cite{shreiner2009opengl} library,  derives clean sensor data streams conditional on the trajectory and sensor configurations (see Fig.~\ref{fig:overview-simulator-pipeline}d). Sensor corruptions are introduced into the sensor streams to mimic real-world observational anomalies and sensor failures (see Fig.~\ref{fig:overview-simulator-pipeline}e).   By utilizing the generated noisy data, which encompasses perturbed sensor streams as inputs and perturbed trajectory and 3D scene as ground-truth labels, the robustness of SLAM models to perturbations can be rigorously assessed (see Fig.~\ref{fig:overview-simulator-pipeline}f).

\begin{figure}[t!]
	\centering
	\includegraphics[width=0.48\textwidth]{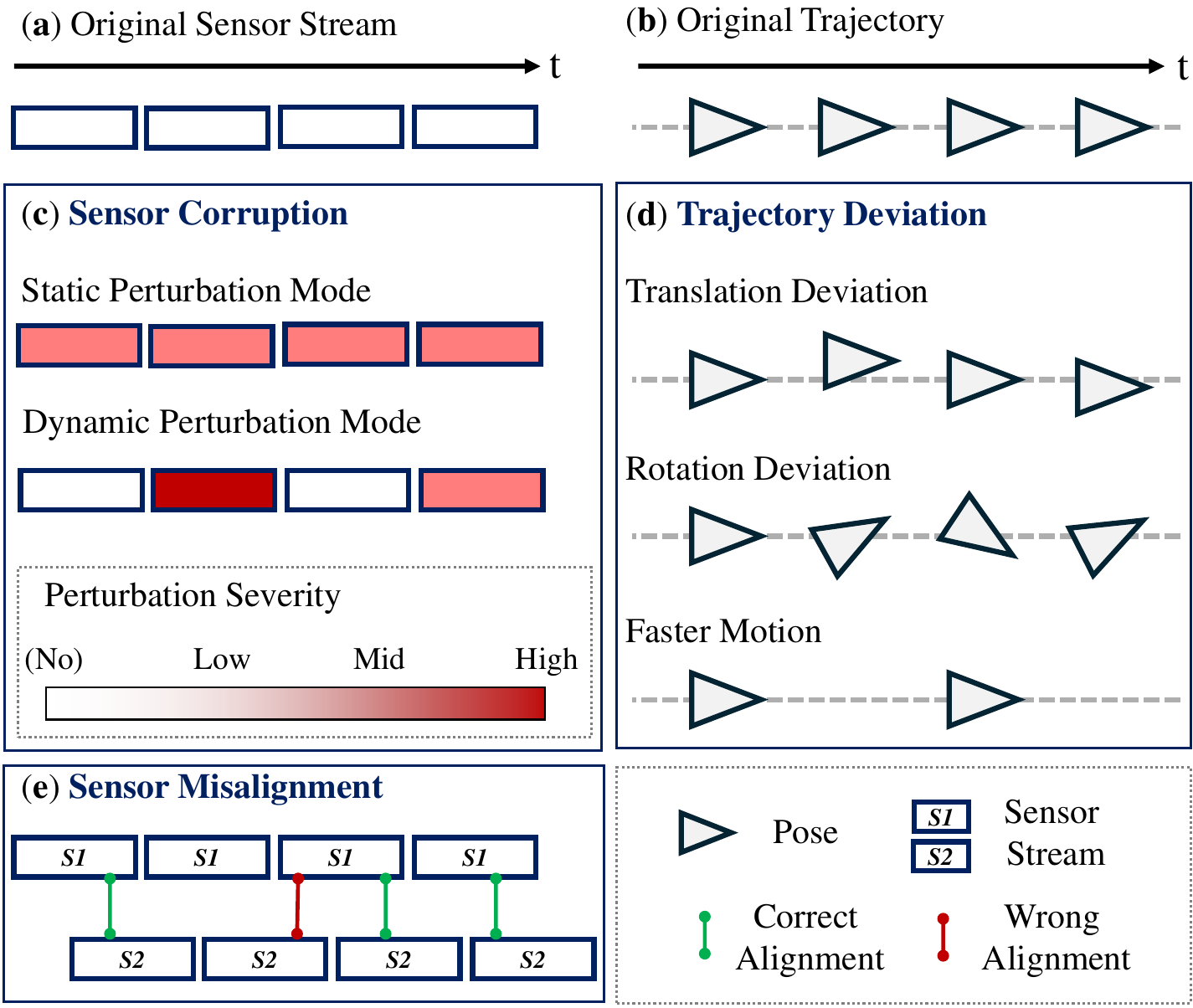} 
	\caption{\textbf{Taxonomy of perturbations for SLAM.} Given (\textbf{a}) the original (clean) sensor stream and (\textbf{b}) trajectory, our noisy data synthesis pipeline enables the simulation of (\textbf{c}) sensor corruption (for both RGB images and depth maps), (\textbf{d}) trajectory perturbations, and (\textbf{e}) multi-sensor misalignment for the multi-modal input setting.}
	\label{fig:all-perturb-taxonomy}
\end{figure}
\begin{figure*}[t]
	\centering
\includegraphics[width=\textwidth]{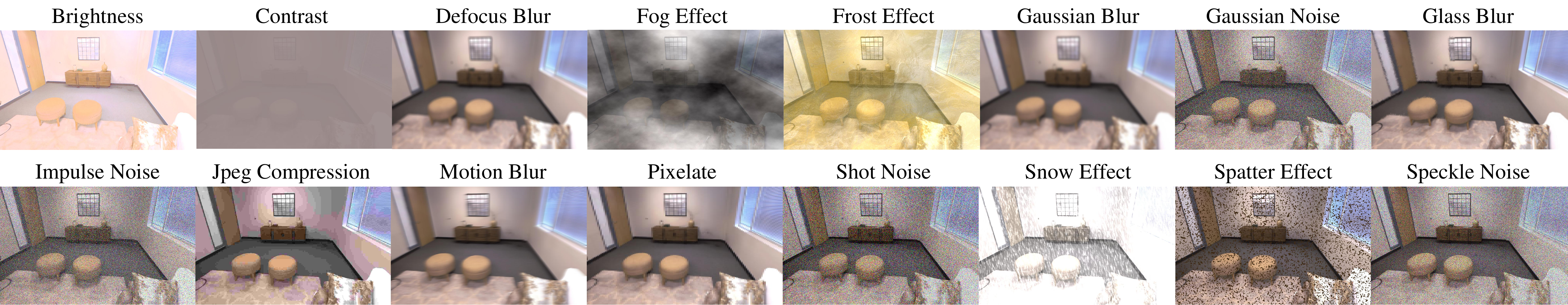} 
	\caption{\textbf{Taxonomy of sensor-level perturbations for RGB imaging}. We consider \textbf{16} common image corruption types~\cite{hendrycks2019robustness} from \textbf{{4}} main categories of perturbations for SLAM robustness evaluation: \textbf{1}) \textbf{{noise-based distortions}}: Gaussian noise, shot noise, impulse noise, and speckle noise; \textbf{2}) \textbf{{blur-based effects}}: defocus blur, glass blur, motion blur, and Gaussian blur; \textbf{3}) \textbf{{environmental interferences}}: snow effect, frost effect, fog effect, and spatter effect; \textbf{4}) \textbf{{post-processing manipulations}}: brightness, contrast, pixelate, and JPEG compression.  } 
	\label{fig:sensor-corruption}
\end{figure*}

\subsection{{Taxonomy of Perturbations for SLAM}}
In Fig.~\ref{fig:all-perturb-taxonomy}, we present common perturbations for SLAM as elaborated in the following paragraphs.

\vspace{1.0mm}\noindent{\textbf{Sensor corruptions for RGB imaging.}}
Prominent image perturbations~\cite{hendrycks2019robustness} (see Fig.~\ref{fig:sensor-corruption}) included in this study are: 
\begin{itemize}
    \item \textbf{Noise-based distortions}. 
    \begin{enumerate}
    \item \textbf{Gaussian noise}. Consider an original clean image denoted by $I$. To simulate the presence of Gaussian noise, we introduce additive noise $\eta$ to create a corrupted version of the image $I'$, expressed as
    \begin{equation}
    I' = I + \eta
    \end{equation}
    where the noise $\eta$ is sampled from a Gaussian distribution with zero mean and variance $\sigma^2$, represented as $\eta \sim \mathcal{N}(0, \sigma^2)$.

    \item \textbf{Shot noise}. 
    Shot noise is associated with the random arrival of photons or particles during the image capture process. It can be modeled using a Poisson distribution~\cite{hasinoff2014photon}.

    \item \textbf{Impulse noise}. For impulse noise, random pixels of the original image $I$ are set to extreme values (\textit{e.g.,} either min or max of the RGB values) with a certain probability $p$. The perturbation process for each pixel at the location $(x, y)$ can be expressed as
        \begin{equation}
        I'(x, y) = 
        \begin{cases} 
          \text{min value} & \text{with probability } p/2 \\
          \text{max value} & \text{with probability } p/2 \\
          I(x, y) & \text{otherwise}
        \end{cases}
        \end{equation}
        \item \textbf{Speckle noise}. Speckle noise~\cite{bianco2018strategies} can be represented as a multiplicative noise model:
        \begin{equation}
        I' = I \times (1 + \rho \times \eta)
        \end{equation}
    where $\rho$ controls the intensity level of the speckle noise, and $\eta$ represents the noise term following a Gaussian distribution.
\end{enumerate}

\item \textbf{Blur-based effects}. 
The blurring process involves convolving ($\ast$) an input image $I$ with a specific blur kernel $K$ to obtain the blurred image $I'$, which is formulated as
\begin{equation}
I' = I \ast K
\end{equation}
\begin{enumerate}
\item  \textbf{Defocus blur}. This effect simulates the out-of-focus visuals caused by camera lens properties. It can be modeled by convolving the input image with a circular disc (bokeh) kernel.
 
\item  \textbf{Glass blur}. Emulating the appearance of viewing through textured or patterned glass, this effect adds complexity to the blurring process. The glass texture is approximated using an irregular kernel.

\item \textbf{Motion blur}. Rapid movement during image capture, either by the camera or objects in the scene, results in motion blur. This effect can be represented by convolving the image with a linear kernel oriented in the direction of motion.

\item \textbf{Gaussian blur}. 
Gaussian blur convolves the image with a Gaussian kernel. The standard deviation of the Gaussian distribution determines the level of blurring.
\end{enumerate}

    \item \textbf{Environmental interference}. Environmental interference or weather effects~\cite{kamann2020increasing,yu2019nighttime} are commonly simulated via alpha blending techniques~\cite{smith1995image}. This involves blending a perturbed weather effect layer $W$ with the original (clean) image $I$ to generate a composite perturbed image $I'$, which can be expressed as
    \begin{equation}
    I' = (1 - \alpha_W) \times I + \alpha_W \times W
    \end{equation}
    where $\alpha_W$ controls the blending proportion.

    \begin{enumerate}
    \item \textbf{Snow effect}. To simulate the disturbances caused by snowfall, we construct a snow effect layer with random regional white values~\cite{von2019simulating}. 
    
    \item \textbf{Frost effect}. The frost effect introduces a semi-transparent whitening overlay on the image~\cite{michaelis2019benchmarking}. This effect is modeled through a weighted combination of the original image and the whitened version of the original image. 

    \item \textbf{Fog effect}. The fog effect results in a hazy observation. A simplified model~\cite{hendrycks2019robustness} to simulate this effect is achieved through linear interpolation between the original image and a constant gray-value image.

    \item \textbf{Spatter effect}. The spatter effect mimics the appearance of droplets on a lens or window. To achieve this effect, a layer comprising semi-transparent dark spots or streaks is blended with the image. Considering the local property of the spatters, we incorporate a hard mask onto the spatter effect layer. This mask designates transparent regions as 0 and perturbed (occupied) regions as 1, thereby controlling the visibility of the spatter effect in specific areas.

    \end{enumerate}

    \item \textbf{Post-processing manipulations}. 
    \begin{enumerate}
    \item \textbf{Brightness}. This effect adjusts the global luminance of the image. For every pixel $(x, y)$ in image $I$, an brightness offset $b$ is added:
    \begin{equation}
    I'(x, y) = I(x, y) + b
    \end{equation}

    \item \textbf{Contrast}. This effect alters the tone variance  of the image by linear scaling about the mean intensity $\mathcal{J}$:
    \begin{equation}
    I' = \alpha \times (I - \mathcal{J}) + \mathcal{J}
    \end{equation}
    where $\alpha$ controls the contrast level.

    \item \textbf{JPEG compression}. This effect simulates lossy compression artifacts when using the JPEG image compression standard. 
    
    \item \textbf{Pixelate}. This effect reduces resolution by dividing the image into blocks and setting all pixels in each block to the block's average value.

    \end{enumerate}
\end{itemize}

\begin{figure}[t!]
	\centering
\includegraphics[width=0.48\textwidth]{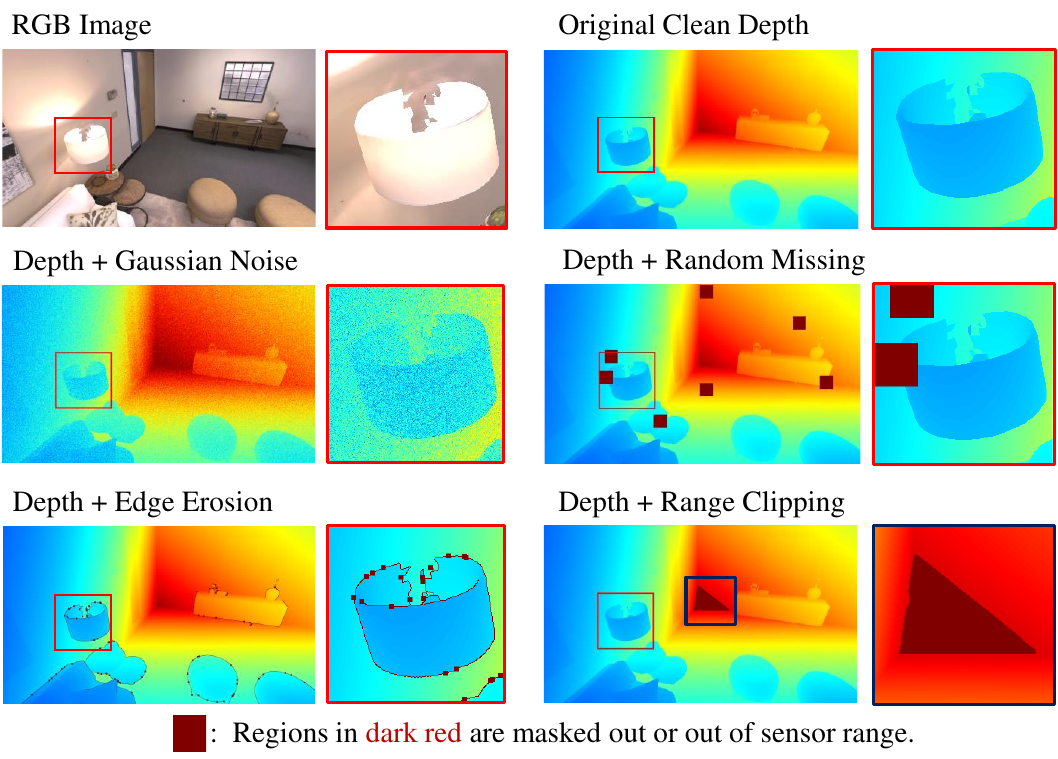} 
	\caption{\textbf{Taxonomy of sensor-level perturbations for depth imaging}. We introduce four perturbations to mimic the perception noises (\textit{i.e.}, random noises and missing) and the limited perception field of depth sensors.
}
	\label{fig:depth-corruption}
\end{figure}

\vspace{1.0mm}\noindent{\textbf{Sensor corruption for depth imaging.}}
As shown in Fig.~\ref{fig:depth-corruption}, we propose a set of generic depth perturbation operations designed specifically for depth images.

\begin{figure}[t!]
	\centering
	\includegraphics[width=0.48\textwidth]{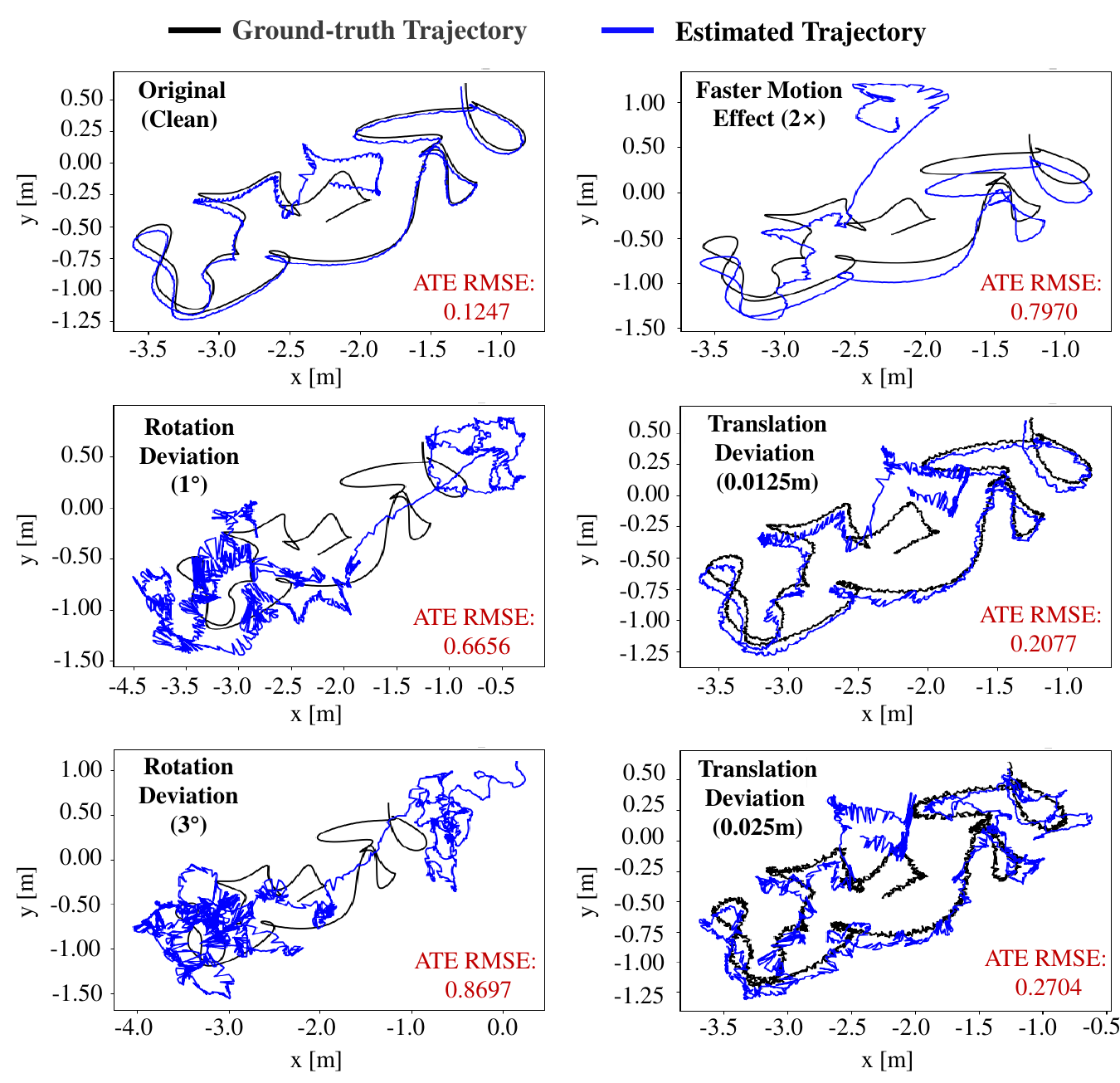} 
	\caption{\textbf{Taxonomy of trajectory deviation and faster motion perturbation.} We present the synthesized ground-truth trajectories (in black) and the estimated trajectories (in blue) obtained using the real-time SLAM model CO-SLAM~\cite{coslam}. For clarity, we visualize the projected trajectory on the horizontal x-y plane derived from the 3D trajectory, which shows that slight trajectory deviations can have a  significant impact on the trajectory estimation performance, measured by ATE.}
	\label{fig:taxonomy-trajectory-misalign-perturbation-simplified}
\end{figure} 

\begin{itemize}
    \item \textbf{{{Gaussian noise}}}. This perturbation mimics sensor random noise $\eta$, which follows Gaussian distribution, in depth measurement.  Let $D$ be the original depth map and $D'$ be the perturbed depth map, then each pixel $(x,y)$ in the depth map is perturbed as
\begin{equation}
D'(x,y) = D(x,y) + \eta
\end{equation}

    \item \textbf{{{Edge erosion}}}. 
    The multi-path interference effect of certain depth sensors (\textit{e.g.}, time-of-flight sensors) can lead to inaccurate depth measurements, particularly for regions with complex geometries. To simulate this perturbation, we first leverage the edge detection algorithm to obtain the edges and then remove a subset of edge pixels $\mathcal{P}$:
    \begin{equation}
    D'(x,y) = \begin{cases}
    $VOID$ & \text{if } (x,y) \in \mathcal{P} \\
    D(x,y) & \text{otherwise}
    \end{cases}
    \end{equation}

    \item \textbf{{{Random missing depth data}}}. This perturbation introduces random masked regions to simulate occlusions or missing depth data. Specifically, a binary mask $M$ is  applied to the depth map, where the masked regions are set to a void value (\textit{e.g.}, 0):
    \begin{equation}
    D' = D  \odot M
    \end{equation}
      where $\odot$ denotes element-wise multiplication.
   
    \item \textbf{{{Range clipping}}}. This perturbation accounts for the limited depth coverage of real-world depth sensors. Objects beyond this range will  appear as missing data in the depth image.  Specifically, any depth value $D(x,y)$ falling outside a specified range  $[D_{min}, D_{max}]$ is replaced with a predefined value $\mathbb{V}$ (\textit{e.g.}, a void value which represents depth missing), expressed as
    \begin{equation}
    D'(x,y) = \begin{cases}
    D(x,y) & \text{if } D_{min} \leq D(x,y) \leq D_{max} \\ \mathbb{V} & \text{otherwise} 
    \end{cases}
    \end{equation}
    \end{itemize}

\vspace{1.0mm}\noindent{\textbf{Trajectory perturbations.}}
As is shown in Fig.~\ref{fig:taxonomy-trajectory-misalign-perturbation-simplified}, we study three types of trajectory perturbations: rotational and translational matrix deviations, and faster motion effects. 

\begin{itemize}
    \item \textbf{Rotation deviation}.  This perturbation induces rotational inconsistencies within a trajectory. By applying the rotation matrix perturbation, denoted as $\Delta \mathbf{R}$, the perturbed rotation matrix $\mathbf{R}'$ is obtained. This process can be formulated as 
    \begin{equation}
        \mathbf{R}' = \mathbf{R} \Delta \mathbf{R} 
    \end{equation}
    \item \textbf{Translation deviation}. This perturbation induces translation inconsistencies within a trajectory. By adding the perturbation translation vector $\Delta \mathbf{t}$ to the original translation vector $\mathbf{t}$, the perturbed translation vector $\mathbf{t}'$ is obtained. This process
can be formulated as
    \begin{equation}
        \mathbf{t}' = \mathbf{t} + \Delta \mathbf{t} 
    \end{equation}
    For our implementation, both perturbations are sampled from Gaussian distribution.
    \item \textbf{Faster motion.}
    Given an existing sensor stream sequence $\mathbf{S}$, a sequence with faster motion ($\mathbf{S}'$) can be synthesized by sampling ($\mathcal{S}$) frames with an interval $k$:
    \begin{equation}
    \mathbf{S}' = \mathcal{S}(\mathbf{S}, k)
    \end{equation}
 
\end{itemize}

\vspace{1.0mm}\noindent{\textbf{Multi-sensor misalignment.}}  
Sensor delays can be emulated by introducing a temporal misalignment between the data streams. Given two temporally-aligned sensor streams, \textit{i.e.}, $\mathbf{S}_{1}(t)$ and $\mathbf{S}_{2}(t)$, a misalignment time delay $\Delta t_m$ is introduced to create misaligned perturbed sensor streams, \textit{i.e.},   $\mathbf{S}_{1}'(t)$ and $\mathbf{S}_{2}'(t)$, which is formulated as
    \begin{equation}
    \mathbf{S}_1'(t) = \mathbf{S}_1(t + \Delta t_m)
    \end{equation}
    \begin{equation}
    \mathbf{S}_2'(t) = \mathbf{S}_2(t)
    \end{equation}

\begin{table}[t]
\caption{Comparison of SLAM benchmarks.}
\centering
\label{tab:datasets}
\setlength{\tabcolsep}{0.1pt}

\resizebox{0.48\textwidth}{!}
{
\begin{tabular}{lcccc}
\hline \hline
\textbf{Benchmark} &
  \textbf{Seq Num} &
  \textbf{Trajectory} &
  \textbf{Sensor}  &
  \textbf{Perturbation} 
 \\ \hline %
 \midrule
   \multicolumn{5}{c}{Real-world SLAM Benchmark}  \\ \midrule
EuRoC~\cite{Burri25012016} & 11 & UAV &  RGB+IMU & \multirow{13}{*}{\begin{tabular}[c]{@{}c@{}}Entangled,\\ Natural\end{tabular}} \\ 
KITTI~\cite{Geiger2012CVPR} &  22 &  Car & RGB+IMU+Lidar & \\ 
Malaga~\cite{blanco2014malaga} &  15 &  Car &  RGB+IMU& \\ 
Newer College~\cite{zhang2021multi} &  3 &  Handheld &  RGB +Lidar+IMU & \\ 
PennCOSYVIO~\cite{pfrommer2017penncosyvio} &  4 &  Handheld &  RGB+IMU &\\ 
TUM-RGBD~\cite{TUM-RGBD} & 19 &  Handheld+UGV&  RGB+Depth&\\
TUM VIO~\cite{schubert2018tum} &  28 &  Handheld & RGB+IMU&\\ 
UMA VI~\cite{zuniga2020vi} &  32 &  Handheld &  RGB+IMU & \\ 
UMich~\cite{carlevaris2016university} &  27 &  UGV &  RGB+IMU&  \\ 
UZH-FPV~\cite{Delmerico19icra} &  28 &  UAV & RGBD$\&$Event+IMU &\\ 
Hilti~\cite{helmberger2022hilti} &  12 &  Handheld &  RGB+Lidar+IMU & \\ 
Campus-Hybrid~\cite{tian2023resilient}   & 3 &  UGV &  RGB+Lidar+IMU& \\ \midrule \midrule
      \multicolumn{5}{c}{Synthesized SLAM Benchmark}  \\ \midrule
Replica~\cite{straub2019replica} & 8 &  Simulation & RGB+Depth & No \\ \midrule
TartanAir${}^{*}$~\cite{wang2020tartanair} & 30 &  Simulation & RGB &   \begin{tabular}[c]{@{}c@{}}Decoupled, \\ 30 settings\end{tabular}   \\ \midrule
\begin{tabular}[c]{@{}c@{}}\textbf{Robust-SLAM}${}^{\dagger}$ (\textbf{Ours})\end{tabular} & {\textbf{1,000}} &  Simulation &  RGB+Depth & \begin{tabular}[c]{@{}c@{}}Decoupled, \\ \textbf{172} settings\end{tabular} 
  \\ \hline \hline
\multicolumn{5}{l}{This table is extended from \cite{helmberger2022hilti}. `Seq Num' indicates sequence number.} \\
\multicolumn{5}{l}{{* Though TartanAir collects diverse 3D scenes, its pioneering case study only  }}\\ 
  \multicolumn{5}{l}{{benchmarks the performance under 30 simulated sequences~\cite{wang2020tartanair}.}}\\
  \multicolumn{5}{l}{{$\dagger$ While our proposed perturbed data synthesizer can enable infinite dataset generation, }}\\
  \multicolumn{5}{l}{{we create a dataset with limited sequence number  here for preliminary analyses. }}\\
\end{tabular}}
\end{table}

\subsection{\textbf{Perturbation Mode and Severity}}

As is illustrated in Fig.~\ref{fig:all-perturb-taxonomy}c, perturbations are examined in two key modes: static and dynamic. In addition, the severity of each perturbation can be categorized into varied levels, \textit{e.g.,} low, medium, and high.

\vspace{1.0mm}\noindent{\textbf{Static perturbation.}}
The severity level of static perturbations remains constant across all sensor frames in a sequence.

\vspace{1.0mm}\noindent{\textbf{Dynamic perturbation.}}
Dynamic perturbations exhibit varying severity from frame to frame, closely resembling real-world disturbances.
\begin{table*}[t]
\centering
\caption{SLAM methods for robustness evaluation${}^{\text{(1)}}$.}
\label{tab:slam_methods}
\resizebox{\textwidth}{!} 
{
\begin{tabular}{l|c|c| l |c|c|r|l|c}
\toprule\toprule
\textbf{Method } & \textbf{Type} & \begin{tabular}[c]{@{}c@{}}\textbf{Modality}\\\textbf{Mono/RGBD}\end{tabular} & \begin{tabular}[c]{@{}c@{}}\textbf{Map} \textbf{Representation}\end{tabular} & \begin{tabular}[c]{@{}c@{}}\textbf{Loop}\\\textbf{Closure}\end{tabular} & \begin{tabular}[c]{@{}c@{}}\textbf{External}\\\textbf{Data}\end{tabular}& \multicolumn{1}{|c|}{\begin{tabular}[c]{@{}c@{}}\textbf{Speed${}^{\text{(3)}}$}\end{tabular}} & \textbf{Processing} & \textbf{Year}\\ 
\midrule 
ORB-SLAM3~\cite{orbslam3}  & Classical, Sparse & $\usym{2713}$\quad/\quad$\usym{2713}$ & Keyframe+ORB, Explicit  &$\usym{2713}$ &  $\usym{2715}$& Real-time & CPU & 2020 \\
iMAP~\cite{imap}  & Neural, Dense  &  $\usym{2715}$\quad/\quad$\usym{2713}$ & NeRF-based${}^{\text{(2)}}$, Implicit &$\usym{2715}$ &  $\usym{2715}$ & Quasi Real-time & CPU+GPU & 2021 \\
Nice-SLAM~\cite{niceslam}  & Neural, Dense & $\usym{2715}$\quad/\quad$\usym{2713}$ & NeRF-based, Implicit &$\usym{2715}$ &  $\usym{2715}$& Quasi Real-time & CPU+GPU & 2022  \\
CO-SLAM~\cite{coslam}  & Neural, Dense & $\usym{2715}$\quad/\quad$\usym{2713}$ & NeRF-based, Implicit &$\usym{2715}$ &  $\usym{2715}$ & Real-time & CPU+GPU & 2023 \\
GO-SLAM~\cite{zhang2023goslam}  & Neural, Dense & $\usym{2713}$\quad/\quad$\usym{2713}$ & NeRF-based, Implicit &$\usym{2713}$ &  $\usym{2713}$${}^{\text{(4)}}$& Quasi Real-time & CPU+GPU & 2023 \\
SplaTAM-S~\cite{keetha2023splatam}  & Neural, Dense & $\usym{2715}$\quad/\quad$\usym{2713}$  & Gaussian~\cite{gaussiansplatting}, Explicit &$\usym{2715}$ &  $\usym{2715}$& Quasi Real-time & CPU+GPU & 2023  \\
\bottomrule\bottomrule
\multicolumn{8}{l}{{$(1)$ We only select several representative multi-modal SLAM models with varied characteristics for evaluation in this work. }}\\
\multicolumn{9}{l}{{$(2)$ `NeRF-based' indicates methods that leverage implicit neural networks to encode the 3D scene, following the philosophy of NeRF~\cite{rosinol2022nerf}.}}\\
\multicolumn{9}{l}{{$(3)$ For the speed comparison, `Real-time' and `Quasi Real-time' are in the speed decreasing rank.}}\\
\multicolumn{9}{l}{{$(4)$ GO-SLAM initializes the model parameters from the DROID-SLAM model~\cite{teed2021droid} which leverages external data~\cite{wang2020tartanair} for model pre-training.}}\\
\end{tabular}}

\end{table*}

\section{Customizable Robustness Benchmark}\label{sec:experiment}

Leveraging our established noisy data synthesis pipeline, we instantiate \textit{{Robust-SLAM}}, a large-scale benchmark designed for the robustness evaluation of cutting-edge monocular and multi-modal SLAM models under diverse perturbations. Robust-SLAM consists of an extensive collection of 1,000 sequences, containing nearly 2 million RGBD frames at a resolution of 1200×680 pixels. Ground-truth trajectories and 3D scenes are also provided for evaluation. In the following sections, we delve into the details of \textit{Robust-SLAM} benchmark and evaluate the performance of some advanced monocular and multi-modal SLAM models under perturbation.

\begin{table*}[t]
\caption{Trajectory estimation error (ATE$\downarrow$ [m]) under static (\textbf{Top}) and dynamic (\textbf{Bottom}) image perturbations. }
\label{tab:image_perturb}
\centering\setlength{\tabcolsep}{0.4mm}
\resizebox{\textwidth}{!}{
\begin{tabular}{l|c|cc|cccc|cccc|cccc|cccc}
    \toprule \toprule
     \multirow{2}{*}{\textbf{Method}} &  \multirow{2}{*}{\textbf{Clean}}&  \multicolumn{2}{|c|}{\textbf{Perturbed}}  & \multicolumn{4}{|c|}{\textbf{Blur Effect}} & \multicolumn{4}{c|}{\textbf{Noise Effect}} & \multicolumn{4}{c|}{\textbf{Environmental Interference}}  & \multicolumn{4}{c}{\textbf{Post-processing Effect}} 
    \\ \cmidrule{5-8} \cmidrule{9-12} \cmidrule{13-16} \cmidrule{17-20}
    &  & \textbf{Mean} & \textbf{Max}  & \textbf{Motion} & \textbf{Defocus} & \textbf{Gaussian} & \textbf{Glass} & \textbf{Gaussian} & \textbf{Shot} & \textbf{Impulse} & \textbf{Speckle} & \textbf{Fog} & \textbf{Frost} & \textbf{Snow} & \textbf{Spatter} & \textbf{Bright} & \textbf{Contra.} & \textbf{Jpeg}  & \textbf{Pixelate}
    \\\midrule\midrule
    \multicolumn{20}{c}{\textbf{Static Perturbation} Mode for \textbf{Monocular} SLAM Models}  \\ \midrule
    ORBSLAM3~\cite{orbslam3} & $0.0145$ & $0.750$\textit{\textbf{F}} & $1.000$\textit{\textbf{F}}  & $0.891$\textit{\textbf{F}} & $0.535$\textit{\textbf{F}} & $0.505$\textit{\textbf{F}} & $0.797$\textit{\textbf{F}} & $0.917$\textit{\textbf{F}} & $0.969$\textit{\textbf{F}} & $1.000$\textit{\textbf{F}} & $0.923$\textit{\textbf{F}} & $0.629$\textit{\textbf{F}} & $0.917$\textit{\textbf{F}} & $1.000$\textit{\textbf{F}} & $0.719$\textit{\textbf{F}} & ${0.0265}$ & $0.612$\textit{\textbf{F}} & $0.7949$ & $0.768$\textit{\textbf{F}} 
    \\    
    GO-SLAM~\cite{zhang2023goslam} & $\textbf{0.0039}$ & $\textbf{0.0903}$  & $\textbf{0.7207}$ & $\textbf{0.0151}$ & $\textbf{0.0052}$ & $\textbf{0.0052}$ & $\textbf{0.0089}$ & $\textbf{0.0776}$ & $\textbf{0.0456}$ & $\textbf{0.0296}$ & $\textbf{0.0190}$ & $\textbf{0.2157}$ & $\textbf{0.7207}$ & $\textbf{0.1921}$ & $\textbf{0.0859}$ & $\textbf{0.0046}$ & $\textbf{0.0047}$ & $\textbf{0.0095}$ & $\textbf{0.0046}$ 
   
    \\ \midrule
        \multicolumn{20}{c}{\textbf{Static Perturbation} Mode for \textbf{RGBD} SLAM Models}  \\ \midrule
    ORBSLAM3~\cite{orbslam3} & $0.0823$ & $0.472$\textit{\textbf{F}} & $0.9317$ & $0.300$\textit{\textbf{F}} & $0.340$\textit{\textbf{F}} & $0.292$\textit{\textbf{F}} & $0.211$\textit{\textbf{F}} & $0.470$\textit{\textbf{F}} & $0.433$\textit{\textbf{F}} & $0.591$\textit{\textbf{F}} & $0.351$\textit{\textbf{F}} & $0.397$\textit{\textbf{F}} & $0.640$\textit{\textbf{F}} & $0.795$\textit{\textbf{F}} & $0.508$\textit{\textbf{F}} & $0.0649$ & $0.527$\textit{\textbf{F}} & $0.9317$ & $0.7034$
    \\ 
    iMAP~\cite{imap} & $0.1209$ & $0.1568$ & $0.3831$ & $0.1424$ & $0.1671$ & $0.1811$ & $0.0672$ & $0.0278$ & $0.0779$ & $0.1710$ & $0.1087$ & $0.1913$ & $0.1316$ & $0.1665$ & $0.1473$ & $0.1903$ & $0.3831$ & $0.1884$ & $0.1669$  
    \\
    Nice-SLAM~\cite{niceslam}& $0.0147$ & $0.0253$ & $0.0654$  & $0.0307$ & $0.0151$ & $0.0161$ & $0.0188$ & $0.0254$ & $0.0377$ & $0.0353$ & $0.0151$ & $0.0186$ & $0.0160$ & $0.0323$ & $0.0320$ & $0.0654$ & $0.0161$ & $0.0150$ & $0.0145$
    \\
    CO-SLAM~\cite{coslam}& $0.0090$ & $0.0104$ &  $\textbf{0.0125}$  & $\textbf{0.0115}$ & $0.0096$ & $0.0097$ & $0.0097$ & $0.0125$ & $0.0101$ & $0.0099$ & $0.0105$ & $0.0118$ & $0.0113$ & $0.0104$ & $0.0098$ & $0.0103$ & $0.0112$ & $0.0094$ & $0.0094$

    \\

    GO-SLAM~\cite{zhang2023goslam} & $0.0046$ & $0.0574$ &$0.6271$ & $0.0135$ & $\textbf{0.0052}$ & ${0.0052}$ & $0.0090$ & $0.0169$ & $0.0140$ & $0.0171$ & $0.0100$ & $0.1211$ & $0.6271$ & $0.0416$ & $0.0164$ & $0.0047$ & $\textbf{0.0054}$ & $0.0065$ & $0.0050$ 
    \\
    SplaTAM-S~\cite{keetha2023splatam}& $\textbf{0.0045}$ & $\textbf{0.0062}$ & $0.0160$ & $0.0160$ & $\textbf{0.0052}$  & $\textbf{0.0049}$ & $\textbf{0.0048}$ & $\textbf{0.0054}$ & $\textbf{0.0050}$ & $\textbf{0.0044}$ & $\textbf{0.0051}$ & $\textbf{0.0085}$ & $\textbf{0.0063}$ & $\textbf{0.0048}$ & $\textbf{0.0051}$ & $\textbf{0.0038}$ & $0.0133$ & $\textbf{0.0044}$ & $\textbf{0.0048}$  
    \\  \midrule
        \multicolumn{20}{c}{\textbf{\textit{Survival Upper-bound:}} The highest expected performance (min ATE) attainable using all monocular and RGBD SLAM models.}  \\ \midrule
      min ATE  & \multirow{3}{*}{$\textbf{0.0039}$} & \multirow{3}{*}{${\textbf{0.0056}}$}   & {${0.0115}$} & $0.0115$ & $0.0052$  & $0.0049$ & $0.0048$ & $0.0054$ & $0.0050$ & $0.0044$ & $0.0051$ & $0.0085$ & $0.0063$ & $0.0048$ & $0.0051$ & $0.0038$ & $0.0054$ & $0.0044$ & $0.0046$  
    \\ \cmidrule{5-8}\cmidrule{9-12}\cmidrule{13-16}\cmidrule{17-20}
     mean(min ATE)& &  & {0.0066}  &  \multicolumn{4}{c|}{${0.0066}$}&  \multicolumn{4}{c|}{$0.0050$}&  \multicolumn{4}{c|}{$0.0062$} &  \multicolumn{4}{c}{$0.0046$}
    \\ 
       \midrule     \midrule
    \multicolumn{20}{c}{\textbf{Dynamic Perturbation} Mode for \textbf{Monocular} SLAM Models}  \\ \midrule
    ORBSLAM3~\cite{orbslam3} & $0.0145$ & $0.827$\textit{\textbf{F}} & $1.000$\textit{\textbf{F}}  & $0.925$\textit{\textbf{F}} & $0.506$\textit{\textbf{F}} & $0.760$\textit{\textbf{F}} & $0.813$\textit{\textbf{F}} & $1.000$\textit{\textbf{F}} & $1.000$\textit{\textbf{F}} & $1.000$\textit{\textbf{F}} & $0.885$\textit{\textbf{F}} & $0.751$\textit{\textbf{F}} & $1.000$\textit{\textbf{F}} & $1.000$\textit{\textbf{F}} & $1.000$\textit{\textbf{F}} & $0.0517$ & $1.000$\textit{\textbf{F}} & $0.8276$ & $0.718$\textit{\textbf{F}} 
    \\    
        GO-SLAM~\cite{zhang2023goslam} & $\textbf{0.0039}$ & $\textbf{0.0933}$ & $\textbf{0.7395}$ & $\textbf{0.0155}$ & $\textbf{0.0065}$   & $\textbf{0.0060}$ & $\textbf{0.0090}$  & $\textbf{0.0509}$ & $\textbf{0.0253}$ & $\textbf{0.0396}$ & $\textbf{0.0158}$ & $\textbf{0.2668}$ & $\textbf{0.7395}$& $\textbf{0.2254}$  & $\textbf{0.0474}$ & $\textbf{0.0066}$ & $\textbf{0.0050}$ & $\textbf{0.0298}$ & $\textbf{0.0044}$ 
    \\
    \midrule
        \multicolumn{20}{c}{\textbf{Dynamic Perturbation} Mode for \textbf{RGBD} SLAM Models}  \\ \midrule
   ORBSLAM3~\cite{orbslam3} & $0.0823$ & $0.497$\textit{\textbf{F}} & $1.000$\textit{\textbf{F}}  & $0.279$\textit{\textbf{F}} & $0.303$\textit{\textbf{F}} & $0.0515$ & $0.1680$ & $0.508$\textit{\textbf{F}} & $0.513$\textit{\textbf{F}} & $0.755$\textit{\textbf{F}} & $0.262$\textit{\textbf{F}} & $0.659$\textit{\textbf{F}} & $0.256$\textit{\textbf{F}} & $0.876$\textit{\textbf{F}} & $0.753$\textit{\textbf{F}} & $0.0660$ & $0.751$\textit{\textbf{F}} & $1.000$\textit{\textbf{F}} & $0.756$\textit{\textbf{F}} 
    \\
    iMAP~\cite{imap} & $0.1209$ & $0.1756$ & $0.2873$ & $0.1243$ & $0.1042$ & $0.2149$  & $0.1221$ & $0.1354$ & $0.1170$ & $0.1967$ & $0.1576$ & $0.2279$ & $0.2873$ & $0.2412$ & $0.1528$ & $0.2141$ & $0.2576$ & $0.1607$ & $0.0955$ 
    \\
    Nice-SLAM~\cite{niceslam}& $0.0147$ & $0.0214$ &$0.0409$ & $0.0157$ & $0.0252$ & $0.0359$ & $0.0211$ & $0.0288$ & $0.0409$ & $0.0146$ & $0.0155$ & $0.0167$ & $0.0211$ & $0.0197$ & $0.0187$ & $0.0206$ & $0.0155$ & $0.0146$ & $0.0170$
    \\
    CO-SLAM~\cite{coslam}& $0.0090$ & $0.0105$ & $\textbf{0.0117}$ & $\textbf{0.0107}$ & $0.0095$ & $0.0115$ & $0.0093$ & $0.0106$ & $0.0103$ & $0.0102$ & $0.0098$ & $0.0117$ & $0.0116$ & $\textbf{0.0111}$ & $0.0109$ & $0.0106$ & $0.0111$ & $0.0095$ & $0.0097$
    \\
 GO-SLAM~\cite{zhang2023goslam} & $0.0046$ & $0.0363$  & $0.2213$ & $0.0130$ & $0.0057$ & $0.0055$ & $0.0078$ & $0.0185$ & $0.0117$ & $0.0139$ & $0.0098$ & $0.1685$ & $0.2213$ & $0.0637$ & $0.0166$ & $\textbf{0.0051}$ & $\textbf{0.0052}$ & $0.0092$ & $0.0049$ 
    \\ 
    SplaTAM-S~\cite{keetha2023splatam}& $\textbf{0.0045}$ & $\textbf{0.008}$\textit{\textbf{G}} & $0.045$\textit{\textbf{G}} & $0.0191$ & $\textbf{0.0053}$  & $\textbf{0.0052}$ & $\textbf{0.0050}$ & $\textbf{0.0058}$ & $\textbf{0.0072}$ & $\textbf{0.0044}$ & $\textbf{0.0067}$ & $\textbf{0.0062}$ & $\textbf{0.0062}$ & $0.045$\textit{\textbf{G}} & $\textbf{0.0041}$ & $0.0054$ & $0.0096$ & $\textbf{0.0046}$ & $\textbf{0.0045}$  
    \\     \midrule
        \multicolumn{20}{c}{\textbf{\textit{Survival Upper-bound:}} The highest expected performance (min ATE) is attainable using all monocular and RGBD SLAM models.}  \\ \midrule
      min ATE  & \multirow{3}{*}{$\textbf{0.0039}$} & \multirow{3}{*}{${\textbf{0.0061}}$}   & {${0.0111}$} & ${0.0107}$ & ${0.0053}$  & ${0.0052}$ & ${0.0050}$ & ${0.0058}$ & ${0.0072}$ & ${0.0044}$ & ${0.0067}$ & ${0.0062}$ & ${0.0062}$ & {${0.0111}$} & ${0.0041}$ & ${0.0051}$ & ${0.0050}$ & ${0.0046}$ & ${0.0044}$  
    \\ \cmidrule{5-8}\cmidrule{9-12}\cmidrule{13-16}\cmidrule{17-20}
     mean(min ATE)& &  & {${0.0069}$}  &  \multicolumn{4}{c|}{0.0066}&  \multicolumn{4}{c|}{0.0060}&  \multicolumn{4}{c|}{0.0069} &  \multicolumn{4}{c}{0.0048}
    \\ 
    \bottomrule  \bottomrule

        \multicolumn{20}{l}{{\textbf{1}) As ORBSLAM3~\cite{orbslam3} stops predicting on subsequent frames after losing track, we follow previous literature~\cite{wang2020tartanair,orbslam3} to calculate the ATE score on tracked partial trajectory. }}\\
        \multicolumn{20}{l}{{\textbf{2})  For neural SLAM models (iMAP, Nice-SLAM, CO-SLAM, GO-SLAM, and SplaTAM-S), we follow the literature~\cite{imap,niceslam,coslam,keetha2023splatam,zhang2023goslam} to calculate ATE on the full trajectory.}}\\
            \multicolumn{20}{l}{{\textbf{3}) \textit{\textbf{F}} and \textit{\textbf{G}} represent settings that include failure sequences where no final trajectory is generated. Specifically, \textit{\textbf{F}} denotes the failure is caused by tracking loss, while \textit{\textbf{G}}  denotes }}\\
        \multicolumn{20}{l}{{ the  failure is due to running out of GPU memory (more than 48GB). The number in front of \textit{\textbf{F}} and \textit{\textbf{G}} represents the average ATE as failure sequences are set as a value of 1.0.}}\\
             \multicolumn{20}{l}{{\textbf{4}) We highlight the best trajectory estimation performance, \textit{i.e.}, the lowest ATE, among different methods under each setting in \textbf{bold}.}}
    \end{tabular}
}
\end{table*}

\begin{table*}[t]
\caption{Success rate (SR$\uparrow$) of pose tracking under image perturbations for ORBSLAM3~\cite{orbslam3}. }
\label{tab:image_perturb_sr_metric}
\centering\setlength{\tabcolsep}{0.6mm}
\resizebox{\textwidth}{!}{
\begin{tabular}{cc|c|cc|cccc|cccc|cccc|cccc}
    \toprule \toprule
  \textbf{Perturb.} &   \textbf{Input} &  \multirow{2}{*}{\textbf{Clean}}&  \multicolumn{2}{|c|}{\textbf{Perturbed}}  & \multicolumn{4}{|c|}{\textbf{Blur Effect}} & \multicolumn{4}{c|}{\textbf{Noise Effect}} & \multicolumn{4}{c|}{\textbf{Environmental Interference}}  & \multicolumn{4}{c}{\textbf{Post-processing Effect}} 
    \\ \cmidrule{6-9} \cmidrule{10-13} \cmidrule{14-17} \cmidrule{18-21}
 \textbf{Mode}  & \textbf{Modality}   &  & \textbf{Mean} & \textbf{Min}  & \textbf{Motion} & \textbf{Defocus} & \textbf{Gaussian} & \textbf{Glass} & \textbf{Gaussian} & \textbf{Shot} & \textbf{Impulse} & \textbf{Speckle} & \textbf{Fog} & \textbf{Frost} & \textbf{Snow} & \textbf{Spatter} & \textbf{Bright} & \textbf{Contra.} & \textbf{JPEG}  & \textbf{Pixelate}
    \\\midrule\midrule
    \multirow{4}{*}{\textbf{Static}} &\multirow{2}{*}{Mono} & ${0.854}$ & $0.220$ & $0.000$  & $0.059$ & $0.331$ & $0.382$ & $0.122$ & $0.055$ & $0.016$ & $0.000$ & $0.052$ & $0.229$ & $0.057$ & $0.000$ & $0.211$ & $0.915$ & $0.320$ & $0.712$ & $0.064$
    \\ 
     &   & ${\pm0.149}$ & ${\pm0.265}$ & $\pm0.000$   & $\pm0.152$ & $\pm0.385$ & $\pm0.415$ & $\pm0.307$ & $\pm0.186$ & $\pm0.076$ & $\pm0.000$ & $\pm0.183$ & $\pm0.362$ & $\pm0.197$ & $\pm0.000$ & $\pm0.352$ & $\pm0.142$ & $\pm0.405$ & $\pm0.349$ & $\pm0.089$
    \\ \cmidrule{2-21}
    &\multirow{2}{*}{RGBD} & $\textbf{0.960}$ & $\textbf{0.460}$ & $\textbf{0.081}$ & $\textbf{0.621}$ & $\textbf{0.606}$ & $\textbf{0.617}$ & $\textbf{0.778}$ & $\textbf{0.388}$ & $\textbf{0.391}$ & $\textbf{0.219}$ & $\textbf{0.443}$ & $\textbf{0.276}$ & $\textbf{0.111}$ & $\textbf{0.081}$ & $\textbf{0.259}$ & $\textbf{0.971}$ & $\textbf{0.412}$ & $\textbf{0.818}$ & $\textbf{0.361}$
    \\
        &  & ${\pm0.046}$ & ${\pm0.256}$ & $\pm0.236$ & $\pm0.423$ & $\pm0.443$ & $\pm0.430$ & $\pm0.319$ & $\pm0.468$ & $\pm0.475$ & $\pm0.409$ & $\pm0.457$ & $\pm0.421$ & $\pm0.282$ & $\pm0.236$ & $\pm0.405$ & $\pm0.030$ & $\pm0.461$ & $\pm0.284$ & $\pm0.232$
    \\
  \midrule\midrule
   \multirow{4}{*}{\textbf{Dynamic}} &\multirow{2}{*}{Mono}  & ${0.854}$ & $0.153$ & $0.000$  & $0.008$ & $0.267$ & $0.158$ & $0.000$ & $0.000$ & $0.000$ & $0.000$ & $0.096$ & $\textbf{0.169}$ & $0.000$ & $0.000$ & $0.000$ & $0.904$ & $0.000$ & $\textbf{0.703}$ & $0.142$
    \\
     &   & ${\pm0.149}$ &${\pm0.270}$ & $\pm0.000$  & $\pm0.022$ & $\pm0.320$ & $\pm0.315$ & $\pm0.000$ & $\pm0.000$ & $\pm0.000$ & $\pm0.000$ & $\pm0.273$ & $\pm0.315$ & $\pm0.000$ & $\pm0.000$ & $\pm0.000$ & $\pm0.046$ & $\pm0.000$ & $\pm0.328$ & $\pm0.168$
    \\ \cmidrule{2-21}
   &\multirow{2}{*}{RGBD}  & $\textbf{0.960}$ & $\textbf{0.323}$ & $0.000$  & $\textbf{0.871}$ & $\textbf{0.354}$ & $\textbf{0.301}$ & $\textbf{0.991}$ & $\textbf{0.207}$ & $\textbf{0.277}$ & $\textbf{0.040}$ & $\textbf{0.303}$ & $0.027$ & $\textbf{0.005}$ & ${0.000}$  & $\textbf{0.009}$ & $\textbf{0.975}$ & $\textbf{0.003}$ & $0.376$ & $\textbf{0.432}$ 
\\  
   &   & $\pm{0.046}$ &$\pm{0.344}$  & $\pm0.000$  & $\pm0.364$ & $\pm0.429$ & $\pm0.437$ & $\pm0.081$ & $\pm0.378$ & $\pm0.473$ & $\pm0.084$ & $\pm0.441$ & $\pm0.045$ & $\pm0.009$ & $\pm0.000$  & $\pm0.026$ & $\pm0.027$ & $\pm0.007$ & $\pm0.519$ & $\pm0.218$ 
\\  
    \bottomrule   \bottomrule  
\multicolumn{21}{l}{{Results are presented with \texttt{mean}$ \pm $\texttt{standard deviation}. We highlight the best performance, \textit{i.e.}, the highest SR, under static and dynamic perturbation mode, respectively, in \textbf{bold}.}}
    \end{tabular}
}
\end{table*}

\subsection{{Benchmark Construction}}

\vspace{1.0mm}\noindent{\textbf{Benchmark base 3D scene selection details}}. We synthesize the perturbed benchmark using 3D scene models sourced from the Replica dataset~\cite{straub2019replica}, which comprises real 3D scans of indoor scenes. We select the same set of eight rooms and offices as the (clean) Replica SLAM dataset rendered by iMAP~\cite{imap} for consistent comparison.

\vspace{1.0mm}\noindent{\textbf{Benchmark perturbation generation details}}. We compose a diverse set of perturbations using our noisy data synthesis pipeline. \textbf{1}) For sensor perturbations, we utilize the same trajectory as~\cite{imap} to render clean sensor streams, and then introduce perturbations for each frame. For the perturbation severity of RGB images, we follow the severity used for the most prevalent image robustness benchmark~\cite{hendrycks2019robustness}; for the perturbation severity of depth maps, we refer to the depth range and distribution in real-world dataset TUM-RGBD~\cite{schubert2018tum} to set up the hyper-parameters. \textbf{2}) For trajectory perturbations, we perturb the original trajectories in~\cite{imap} by introducing additional translation and orientation deviations. \textbf{3}) To simulate faster motion, we down-sample sensor streams from the clean source by 2, 4, and 8 times. \textbf{4}) To mimic sensor stream misalignment, we introduce frame delay (5, 10, and 20 frames) between RGB and depth sensor streams. Please refer to the appendix for more details.

\vspace{1.0mm}\noindent{\textbf{Comparison with existing benchmarks.}}
As shown in Table~\ref{tab:datasets}, we highlight the superiority of our synthesized benchmark, \textit{i.e.}, \textit{Robust-SLAM}, in two aspects:
\begin{itemize}
    \item \textbf{Isolation of perturbations}. Real-world datasets often contain multiple entangled corruptions, making it difficult to discern the impact of individual corruptions. In contrast, our benchmark allows for controllable evaluation of SLAM robustness under isolated perturbations. 
    \item \textbf{Diversity and scale}. Compared to existing simulated SLAM benchmarks, such as Replica~\cite{straub2019replica} and TartanAir~\cite{wang2020tartanair}, our benchmark offers more diverse perturbation types at a large scale (172 perturbation settings). Moreover, our benchmark is large-scale in total sequence number (1,000 long video sequences and nearly 2 million image-depth pairs), which allows for statistically significant evaluation of SLAM algorithms.
\end{itemize}

\begin{figure*}[t]
	\centering
\includegraphics[width=\textwidth]{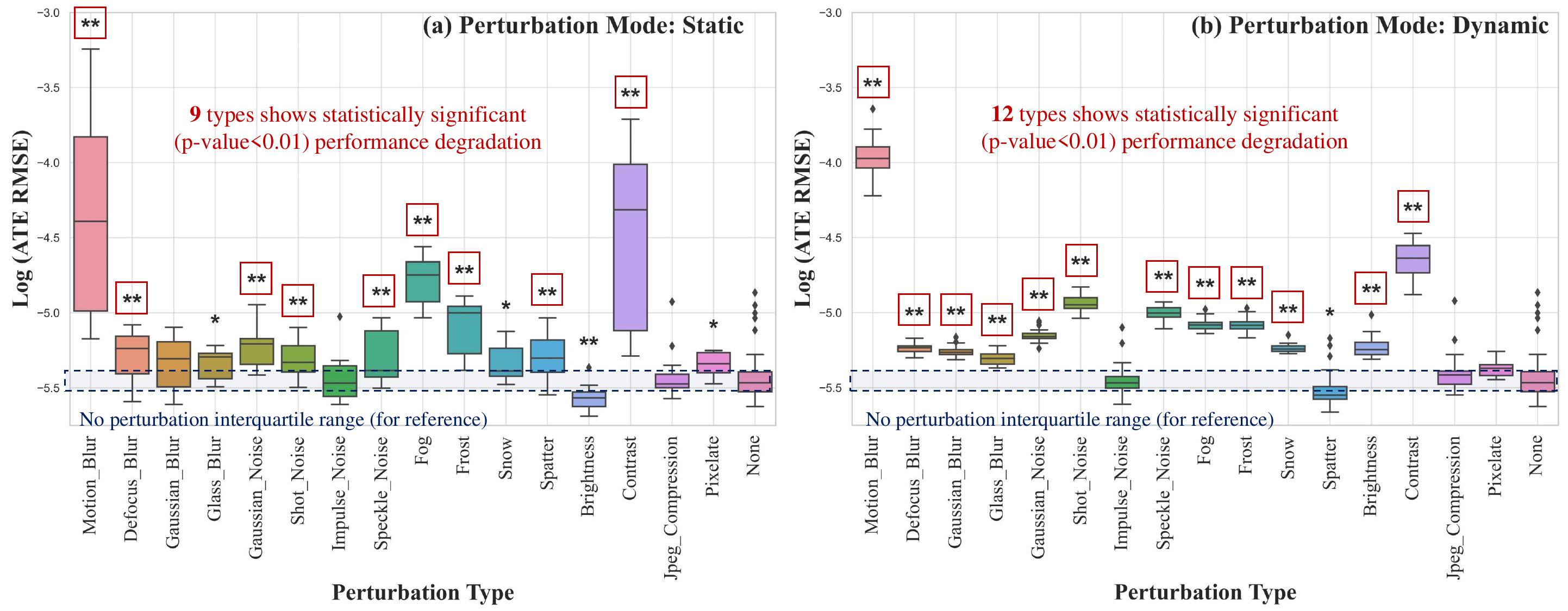}
	\caption{\textbf{Effect of each image perturbation type
 on the trajectory estimation performance}, measured via ATE in the logarithm form, across (\textbf{a}) static and (\textbf{b}) dynamic perturbation mode for SplaTAM-S~\cite{keetha2023splatam} model, which shows the best robustness under image-level corruption among all the benchmark methods. The t-test~\cite{kim2015t} is performed to compare the performance distribution between each perturbed setting and the perturbation-free setting (which is denoted as \textit{None} in the last column of each sub-figure). $**$ and $*$ indicate a significant distribution difference between the pair at the $0.01$ and $0.05$ significance level, respectively.}
	\label{fig:sensor-perturb-splatam-compare}
\end{figure*}

\subsection{{Benchmarking Methods}}

Existing pioneering robustness evaluation for SLAM~\cite{wang2020tartanair,bujanca2021robust} only considers classical SLAM models which mostly leverage heuristic map representation. In contrast, our benchmark considers more advanced multi-modal SLAM models and neural-based models.
Specifically, Table~\ref{tab:slam_methods} presents the evaluated SLAM models: \textbf{1}) classical SLAM: we choose the prevalent and advanced model  ORB-SLAM3~\cite{orbslam3}; \textbf{2}) neural-based SLAM, comprising iMAP~\cite{imap}, Nice-SLAM~\cite{niceslam}, CO-SLAM~\cite{coslam}, GO-SLAM~\cite{zhang2023goslam}, and SplaTAM-S~\cite{keetha2023splatam}. The hyperparameters are set based on the recommendations given in the original papers or use default settings otherwise.

\subsection{{Evaluation Metrics}}
\noindent\textbf{Absolute trajectory error (ATE) of trajectory estimation.} To evaluate the accuracy of trajectory estimation, we adopt the widely-used ATE metric, which measures the disparity between the estimated trajectory and the ground truth trajectory. 
Formally, the ATE metric is calculated as:
\begin{equation}
ATE = \sqrt{\frac{1}{N} \sum_{i=1}^{N} | p_{i}^{\mathrm{est}} - p_{i}^{\mathrm{gt}} |^2}.
\end{equation}
Here, $p_{i}^{\mathrm{est}}$ represents the estimated position,  $p_{i}^{\mathrm{gt}}$ represents the corresponding ground truth position, and $N$ denotes the  trajectory frame number. Lower ATE indicates better performance.

For the classical SLAM model ORBSLAM3~\cite{orbslam3} that stops predicting subsequent frames after losing track, we follow previous literature~\cite{wang2020tartanair,orbslam3} to calculate ATE using the tracked partial trajectory. For neural SLAM models evaluated in this study (included iMAP, Nice-SLAM, CO-SLAM, GO-SLAM, and SplaTAM-S), we follow the literature~\cite{imap,niceslam,coslam,keetha2023splatam,zhang2023goslam} to calculate ATE on the full trajectory. Additionally, we utilize the mean and maximum ATE across multiple perturbation settings of each model to measure the expected average performance and worst-case performance, respectively.

\vspace{1.0mm}\noindent\textbf{Success rate (SR) of pose tracking.} 
Considering the limitations of the widely-used ATE metric in evaluating SLAM performance when only a partial trajectory is tracked~\cite{wang2020tartanair}, especially for classical SLAM models like ORBSLAM3, we leverage the SR metric to evaluate pose tracking performance. SR is defined as the ratio of the cumulative sum of Euclidean distances between consecutive estimated poses to the cumulative sum of Euclidean distances between consecutive ground truth poses. 
Formally, the SR metric is defined as:
\begin{equation}
SR = \frac{\sum_{i=1}^{N} ||p_{i}^{\mathrm{est}} - p_{i-1}^{\mathrm{est}}||}{\sum_{i=1}^{N} ||p_{i}^{\mathrm{gt}} - p_{i-1}^{\mathrm{gt}}||}
\end{equation}
where $|| \cdot ||$ denotes the Euclidean distance. A higher SR value indicates better trajectory tracking performance. 

We also use the mean and minimum SR across multiple perturbation settings to quantify the expected average performance and worst-case performance, respectively.

\subsection{{Benchmarking Results}}\label{sec:results}

This section presents the findings of our comprehensive \textit{Robust-SLAM} benchmark, offering insights into the robustness of existing SLAM models across diverse perturbations. To ensure reliability, the values presented in the following tables represent averages obtained from multiple runs across multiple scenes. In addition, for each perturbed setting conducted under the dynamic perturbation mode, we performed three separate runs to minimize the impact of randomness.


\vspace{1.0mm}\noindent{\textbf{Sensor perturbation on RGB imaging.}}
We analyze the effect of image-level perturbations in Table~\ref{tab:image_perturb}, categorized mainly by the input modality (monocular {versus} RGBD) and the perturbation mode (static {versus} dynamic).  In Table~\ref{tab:image_perturb}, the notations \textit{\textbf{F}} and \textit{\textbf{G}} represent settings that include failure sequences where no final trajectory is generated. \textit{\textbf{F}} represents failure caused by tracking loss, while \textit{\textbf{G}} denotes failure due to GPU memory exhaustion ($\geq$48GB). We assign an ATE value of 1.0 to these failure sequences and calculate the mean performance for each setting. In addition, we present the success ratio of tracked trajectory for ORBSLAM3 in Table~\ref{tab:image_perturb_sr_metric}. We offer the following analyses and insights:

\textbf{\textit{Which perturbation mode is more detrimental: static or dynamic?}} 
Dynamic perturbations present a greater challenge compared to static mode. This is true both in terms of comparing the performance of individual models when perturbed and in determining the overall highest expected performance. The latter is referred to as the `\textit{survival upper bound}', derived from all models.
The obvious performance degradation caused by perturbations in the dynamic mode highlights the increased difficulty of handling real-time dynamic visual disturbances. As SplaTAM-S model shows the highest robustness when compared to other advanced SLAM models, we further investigate the effect of each image perturbation on its performance in Fig.~\ref{fig:sensor-perturb-splatam-compare}. The results reveal that the dynamic perturbation mode causes a more substantial degradation in trajectory estimation performance. Specifically, among the 12 types of perturbations tested, a statistically significant difference exists in the distribution of perturbed ATE compared to the scenario without any perturbation, as shown in Fig.~\ref{fig:sensor-perturb-splatam-compare}b. In contrast, under the static perturbation mode, nine types of perturbations lead to statistically significant performance degradation, as shown in Fig.~\ref{fig:sensor-perturb-splatam-compare}a. It is noteworthy that an increase in brightness results in a slight reduction in trajectory error under static perturbation for SplaTAM-S model, but leads to a large error under the dynamic perturbation mode.

\textbf{\textit{Which type of SLAM model exhibits greater robustness: classical or neural-based?}} Most neural SLAM models typically demonstrate greater robustness to image-level perturbations than classical models. This is attributed to the learning-based components in neural SLAM models, which enhance their adaptability. However, the classical SLAM model we examined, specifically ORBSLAM3, sometimes completely loses tracking when the detection of ORB features becomes impossible in a perturbed image (\textit{i.e.}, $SR=0$ in Table~\ref{tab:image_perturb_sr_metric}).

\textbf{\textit{Which modality configuration yields greater robustness: monocular or RGBD?}} {RGBD} SLAM systems generally surpass monocular methods in terms of robustness. This enhanced robustness can be attributed to the incorporation of (nearly clean) depth information, which provides additional resilience against appearance changes. 

\textbf{\textit{What is the most robust model under image perturbation?}} Among all the competitive SLAM methods evaluated, SplaTAM-S emerges as the most robust model in terms of the ATE after perturbations. It consistently achieves the lowest trajectory error across a range of perturbed conditions for both static and dynamic perturbation modes. However, as depicted in Fig.~\ref{fig:sensor-perturb-splatam-compare}, the performance of SplaTAM-S still encounters challenges in dynamic settings. Specifically, 12 out of 16 image-level perturbations result in statistically significant performance degradation for the SplaTAM-S model.

\textbf{\textit{Which perturbation types have the most significant impact?}} Environmental effects induce the largest errors, highlighting the substantial challenge of deploying SLAM under adverse weather conditions. Motion blur also largely degrades the performance of most advanced SLAM methods.

\textbf{\textit{Which perturbation types have the least significant impact?}} Perturbations introduced during post-processing, such as image compression and degradation artifacts, generally exert a relatively minor influence on SLAM performance. 

\begin{table}[t]
\caption{Trajectory estimation error (ATE$\downarrow$ [m]) under depth perturbation for RGBD SLAM methods. }
\label{tab:depth-perturb}
\centering 
\setlength{\tabcolsep}{1.0mm}
\resizebox{0.48\textwidth}{!}{
\begin{tabular}{l|c|cccc}
    \toprule \toprule   
    \multirow{2}{*}{\textbf{Method}} &    \multirow{2}{*}{\textbf{Clean}}
     &  {\textbf{Gaussian}}   & {\textbf{Edge}} & 
  {\textbf{Random}}  & {\textbf{Range}}      \\  
 & & \textbf{Noise} & \textbf{Erosion} & \textbf{Missing} & \textbf{Clipping}\\ \midrule
    ORBSLAM3~\cite{orbslam3} & $0.0823$ & $0.8026$ & $0.8070$ & $0.7556$ & $0.9940$ 
    \\
    iMAP~\cite{imap} & $0.1209$ & $\usym{2715}$ & $0.0307$ & $0.1083$  & $0.2438$ 
    \\ 
    Nice-SLAM~\cite{niceslam}& $0.0147$ & $\usym{2715}$ & $0.0149$ & $0.0154$  & $0.1183$ 
    \\
    CO-SLAM~\cite{coslam}& $0.0090$ & $0.5794$ & $0.0096$ & $0.0094$  & $0.0122$ 
    \\
    GO-SLAM~\cite{zhang2023goslam} & $0.0046$ & $0.0378$ & $0.0046$ & ${0.0046}$ & ${0.0045}$   
    \\
    SplaTAM-S~\cite{keetha2023splatam}& $\textbf{0.0045}$ & $\textbf{0.0042}$ & $\textbf{{0.0046}}$ & $\textbf{0.0042}$  
  & $\textbf{0.0043}$ 
    \\
    \bottomrule \bottomrule
    \multicolumn{6}{l}{{$\usym{2715}$ indicates  completely unacceptable trajectory estimation performance,}} \\
        \multicolumn{6}{l}{{\textit{i.e.}, ATE $\geq$ 1.0 [m]. Models with the best performance are in \textbf{bold}}}
    \end{tabular}
}

\vspace{2mm}
\caption{Success rate (SR$\uparrow$) of pose tracking under depth perturbation for ORBSLAM3~\cite{orbslam3}  with RGBD input. }
\label{tab:depth-perturb_sr_metric}
\centering 
\setlength{\tabcolsep}{0.5mm}
\resizebox{0.48\textwidth}{!}{
\begin{tabular}{c|cccc}
    \toprule \toprule   
       \multirow{2}{*}{\textbf{Clean}}
     &  {\textbf{Gaussian}}   & {\textbf{Edge}} & 
  {\textbf{Random}}  & {\textbf{Range}}      \\  
  & \textbf{Noise} & \textbf{Erosion} & \textbf{Missing} & \textbf{Clipping}\\ \midrule
    ${0.960}\pm 0.046$ & $0.421\pm 0.331$ & $0.379\pm 0.281$ & $0.322\pm 0.354$ & $0.286\pm 0.181$
    \\
    \bottomrule  \bottomrule 
    \multicolumn{5}{l}{{Results are presented with \texttt{mean}$ \pm $\texttt{standard deviation}. }}
    \end{tabular}
}
\end{table}

\vspace{1.0mm}\noindent{\textbf{Sensor perturbation on depth imaging.}}
In Table~\ref{tab:depth-perturb}, we investigate the impact of depth perturbations. \textbf{1}) When faced with partial depth missing perturbations (including random depth value missing, edge erosion, and range clipping), most neural-based SLAM models exhibit minimal degradation owing to their effective pixel-wise optimization mechanism. In contrast, for the classical SLAM model ORBSLAM3, we observed a notable trajectory estimation performance degradation in the presence of depth-related perturbations. In addition, as shown in Table~\ref{tab:depth-perturb_sr_metric}, ORBSLAM3 model demonstrates a significant loss of track when faced with depth missing. \textbf{2}) Compared to missing depth data, the impact of introducing Gaussian noise to the depth maps is more noticeable. Except for SplaTAM-S, all methods show a considerable increase in trajectory estimation error, attributed to noise directly interfering with the depth distribution of observations.

\vspace{1.0mm}\noindent\textbf{Faster motion perturbation.} In Table~\ref{tab:faster-motion-perturbation} and Table~\ref{tab:faster-motion-perturbation_sr_metric}, we evaluate the effect of faster motion perturbation, revealing the limitations of most approaches in achieving acceptable performance at high speeds. Notably, GO-SLAM excels in handling faster motion scenarios, thanks to the integration of global bundle adjustment mechanism. Besides, the classical SLAM model ORBSLAM3 also demonstrates robustness in tackling high-speed scenarios.

\begin{table}[t]

    \centering
    \setlength{\tabcolsep}{4.5mm}
    \caption{Trajectory estimation error (ATE$\downarrow$ [m]) of mono (\textbf{Top}) and RGBD (\textbf{Bottom}) SLAM under faster motion.}
        \label{tab:faster-motion-perturbation}
\resizebox{0.48\textwidth}{!}
{
\begin{tabular}{l|c|ccc}
    \toprule \toprule
     {\textbf{Speed-up Ratio}}
     & {$\textbf{1}\times$} & \textbf{$\textbf{2}\times$} & \textbf{$\textbf{4}\times$} & \textbf{$\textbf{8}\times$}    \\ \midrule
ORBSLAM3~\cite{orbslam3} & ${0.0145}$ & ${0.0085}$ & ${0.0230}$ & ${0.0231}$  \\  
    GO-SLAM~\cite{zhang2023goslam} & $\textbf{0.0039}$ & $\textbf{{0.0042}}$ & $\textbf{{0.0046}}$ & $\textbf{{0.0048}}$ 
    \\
    \midrule \midrule
    ORBSLAM3~\cite{orbslam3} & $0.0823$ & ${0.0186}$ & ${0.0637}$ & ${0.0774}$ \\
    iMAP~\cite{imap}& $0.1209$ & $0.4675$ & $0.9445$ & $\usym{2715}$ 
    \\
    Nice-SLAM~\cite{niceslam} & $0.0147$ & $0.1702$ & $\usym{2715}$ & $\usym{2715}$  
    \\
    CO-SLAM~\cite{coslam}& $0.0090$ & $0.1062$ & $0.9510$ & $\usym{2715}$ 
    \\
    GO-SLAM~\cite{zhang2023goslam} & ${0.0046}$ & $\textbf{{0.0046}}$ & $\textbf{{0.0046}}$ & $\textbf{{0.0050}}$  
    \\
SplaTAM-S~\cite{keetha2023splatam}& $\textbf{0.0045}$ & $0.0184$ & $\usym{2715}$ & $\usym{2715}$ 
    \\
    \bottomrule \bottomrule
  \multicolumn{5}{l}{{\textbf{1}) Notice that the sensor stream under faster motion effect is synthesized}}  \\
  \multicolumn{5}{l}{{ by downsampling the original sensor stream. Thus, the derived sensor stream}}\\
     \multicolumn{5}{l}{{ under faster motion is shorter than the original one.}} \\
         \multicolumn{5}{l}{{\textbf{2}) $\usym{2715}$ indicates  completely unacceptable trajectory estimation performance,}} \\
        \multicolumn{5}{l}{{\textit{i.e.}, ATE $\geq$ 1.0 [m]. Models with the best performance are in \textbf{bold}}}
    \end{tabular}
    
}\vspace{2mm}
 \caption{Success rate (SR$\uparrow$) of pose tracking under faster motion perturbation for ORBSLAM3~\cite{orbslam3}.}
\centering \setlength{\tabcolsep}{0.5mm}
\label{tab:faster-motion-perturbation_sr_metric}
\resizebox{0.48\textwidth}{!}{
\begin{tabular}{l|c|ccc}
    \toprule \toprule
     {\textbf{Speed-up Ratio}}
     & {$\textbf{1}\times$} & \textbf{$\textbf{2}\times$} & \textbf{$\textbf{4}\times$} & \textbf{$\textbf{8}\times$}    \\ \midrule
Monocular & ${0.854}\pm 0.149$ & ${{0.893}}\pm 0.081$ & ${0.909}\pm 0.049$ & ${{0.837}}\pm 0.115$  \\ 
    RGBD &  $\textbf{0.960}\pm 0.046$  & $\textbf{0.964}\pm 0.012$ & ${\textbf{0.938}}\pm 0.029$ & ${\textbf{0.866}}\pm 0.129$ \\
    \bottomrule \bottomrule
       \multicolumn{5}{l}{{\textbf{1}) Results are presented with \texttt{mean}$ \pm $\texttt{standard deviation}. }}\\
       \multicolumn{5}{l}{\textbf{2}) We highlight the best performance, \textit{i.e.}, the highest SR, for each severity}\\
        \multicolumn{5}{l}{  level of faster motion effect in \textbf{bold}.}
    \end{tabular}
}
\end{table}

\begin{table*}[th]
    \centering
       \caption{Trajectory estimation error (ATE$\downarrow$ [m] ) under trajectory-level deviation.}
\centering\setlength{\tabcolsep}{1.0mm}
\resizebox{\textwidth}{!}{
\begin{tabular}{l|c|ccc|cccc|cccc|cccc}
    \toprule \toprule
     \textbf{Rotate [$\deg$]} & \multirow{2}{*}{\textbf{Clean}}  & \multicolumn{3}{c|}{\textbf{\textit{0}}} & \multicolumn{4}{c|}{\textbf{\textit{1}}} & \multicolumn{4}{c|}{\textbf{\textit{3}}} & \multicolumn{4}{c}{\textbf{\textit{5}}}
    \\ \cmidrule{1-1} \cmidrule{3-5} \cmidrule{6-9} \cmidrule{10-13} \cmidrule{14-17}
     \textbf{Translate  [$m$]} &  & \textit{\textbf{0.0125}} &\textit{\textbf{0.025}} & \textbf{\textit{0.05}} & \textit{\textbf{0}} & \textit{\textbf{0.0125}} &\textit{\textbf{0.025}} & \textbf{\textit{0.05}} & \textit{\textbf{0}}  & \textit{\textbf{0.0125}} &\textit{\textbf{0.025}} & \textbf{\textit{0.05}} & \textit{\textbf{0}} & \textit{\textbf{0.0125}} &\textit{\textbf{0.025}} & \textbf{\textit{0.05}} 
    \\\midrule\midrule
    \multicolumn{17}{c}{Monocular SLAM Models}  \\ \midrule
    ORBSLAM3~\cite{orbslam3} & $0.0145$ & $0.0172$ & $0.136$\textbf{\textit{F}} & $0.190$\textbf{\textit{F}} & $0.0634$ & $0.1463$ & $0.348$\textbf{\textit{F}} & $0.1999$ & $0.170$\textbf{\textit{F}} & $0.0548$ & $0.0532$ & $0.0613$ & $0.0732$ & $0.0227$ & $0.0409$ & $0.0803$ 
    \\    
    GO-SLAM~\cite{zhang2023goslam} & $\textbf{0.0039}$ & $\textbf{0.0084}$ & $\textbf{0.0077}$ & $\textbf{0.0091}$ & $\textbf{0.0083}$ & $\textbf{0.0079}$ & $\textbf{0.0082}$ & $\textbf{0.0094}$ & $\usym{2715}$ & $\usym{2715}$ & $\usym{2715}$ & $\usym{2715}$ & $\usym{2715}$ & $\usym{2715}$ & $\usym{2715}$ & $\usym{2715}$  
    \\ \midrule
        \multicolumn{17}{c}{RGBD SLAM Models}  \\ \midrule
    ORBSLAM3~\cite{orbslam3} & $0.0823$ & $0.1913$ & $0.0851$ & $\usym{2715}$ & $0.0585$ & $0.1629$ & $0.0835$ & $\usym{2715}$ & $0.1483$ & $0.0905$ & $0.1581$ & $0.0853$ & $\usym{2715}$ & $0.0723$ & $0.0497$ & $0.0616$ 
    \\
   iMAP~\cite{imap} & $0.1209$ & $0.0334$ & $0.1386$ & $0.0442$ & $0.2438$ & $0.2135$ & $0.3754$ & $0.2801$ & $\usym{2715}$ & $\usym{2715}$ & $\usym{2715}$ & $\usym{2715}$ & $\usym{2715}$ & $\usym{2715}$ & $\usym{2715}$ & $\usym{2715}$   \\
    Nice-SLAM~\cite{niceslam}& $0.0147$ & $0.5812$ & $\usym{2715}$ & $\usym{2715}$ & $\usym{2715}$ & $\usym{2715}$ & $\usym{2715}$ & $\usym{2715}$ & $\usym{2715}$ & $\usym{2715}$ & $\usym{2715}$ & $\usym{2715}$ & $\usym{2715}$ & $\usym{2715}$ & $\usym{2715}$ & $\usym{2715}$ 
    \\
    CO-SLAM~\cite{coslam}& $0.0090$ & $0.0420$ & $0.0848$ & $0.3087$ & $0.4579$ & $0.5069$ & $0.2998$ & $0.5040$ & $0.6443$ & $0.6630$ & $0.7532$ & $0.5772$ & $0.8457$ & $0.7966$ & $0.8277$ & $\usym{2715}$ 
    \\

    GO-SLAM~\cite{zhang2023goslam} & ${0.0046}$ & $\textbf{0.0082}$ & $\textbf{0.0082}$ & $\textbf{0.0081}$ & $\textbf{0.0080}$ & $\textbf{0.0080}$ & $\textbf{0.0078}$ & $\textbf{0.0077}$ & $\usym{2715}$ & $\usym{2715}$ & $\usym{2715}$ & $\usym{2715}$ & $\usym{2715}$ & $\usym{2715}$ & $\usym{2715}$ & $\usym{2715}$  
    \\
SplaTAM-S~\cite{keetha2023splatam}& $\textbf{0.0045}$ & $0.0545$ & $0.0980$ & $0.2964$ & $0.297F$ & $0.2272$ & $0.2313$ & $\usym{2715}$ & $\usym{2715}$ & $\usym{2715}$ & $\usym{2715}$ & $\usym{2715}$ & $\usym{2715}$ & $\usym{2715}$ & $\usym{2715}$ & $\usym{2715}$ 
    \\ 
    \bottomrule \bottomrule
         \multicolumn{17}{l}{{\textbf{1}) As ORBSLAM3~\cite{orbslam3} fails to predict subsequent frames when losing track, we follow~\cite{wang2020tartanair,orbslam3} to present ATE calculated with partial trajectory.}}\\
       \multicolumn{17}{l}{{\textbf{2})  For neural SLAM models (iMAP, Nice-SLAM, CO-SLAM, GO-SLAM, and SplaTAM-S), we follow~\cite{imap,niceslam,coslam,keetha2023splatam,zhang2023goslam} to calculate ATE on the full trajectory.}}\\
    \multicolumn{17}{l}{\textbf{3}) We highlight the best performance, i.e., the lowest ATE, under each trajectory deviation setting of monocular and RGB SLAM models in \textbf{bold}, respectively.} \\
          \multicolumn{17}{l}{{\textbf{4}) Notation \textit{\textbf{F}} represents settings that include failure sequences where no final trajectory is generated due to tracking loss. The number in front of \textit{\textbf{F}} represents the}}\\
     \multicolumn{17}{l}{{ average ATE as failure sequences are set as a value of 1.0. Notation $\usym{2715}$ indicates completely unacceptable trajectory estimation performance, \textit{i.e.}, ATE $\geq$ 1.0 [m].  }}\\ 

            
    \end{tabular}
}
         \label{tab:trajectory-perturbation_sr_metric}
\end{table*}%

\begin{table*}[th]
    \centering
    \vspace{-2mm}
       \caption{Success rate (SR$\uparrow$) of pose tracking under trajectory-level deviation for ORBSLAM3~\cite{orbslam3}.}
\centering\setlength{\tabcolsep}{1.0mm}
\resizebox{\textwidth}{!}{
\begin{tabular}{l|c|ccc|cccc|cccc|cccc}
    \toprule \toprule
     \textbf{Rotate [$\deg$]} & \multirow{2}{*}{\textbf{Clean}}  & \multicolumn{3}{c|}{\textbf{\textit{0}}} & \multicolumn{4}{c|}{\textbf{\textit{1}}} & \multicolumn{4}{c|}{\textbf{\textit{3}}} & \multicolumn{4}{c}{\textbf{\textit{5}}}
    \\ \cmidrule{1-1} \cmidrule{3-5} \cmidrule{6-9} \cmidrule{10-13} \cmidrule{14-17}
     \textbf{Translate  [$m$]} &  & \textit{\textbf{0.0125}} &\textit{\textbf{0.025}} & \textbf{\textit{0.05}} & \textit{\textbf{0}} & \textit{\textbf{0.0125}} &\textit{\textbf{0.025}} & \textbf{\textit{0.05}} & \textit{\textbf{0}}  & \textit{\textbf{0.0125}} &\textit{\textbf{0.025}} & \textbf{\textit{0.05}} & \textit{\textbf{0}} & \textit{\textbf{0.0125}} &\textit{\textbf{0.025}} & \textbf{\textit{0.05}} 
    \\\midrule\midrule
      \multirow{2}{*}{Monocular} & ${0.854}$ & $\textbf{0.489}$ & $0.247$ & $0.128$ & $\textbf{0.662}$ & $0.340$ & $0.221$ & $\textbf{0.120}$ & $\textbf{0.373}$ & $0.096$ & $0.087$ & $0.059$ & $0.413$ & $0.155$ & $\textbf{0.214}$ & $\textbf{0.144}$ 
     \\ 
        & $\pm0.149$ & $\pm0.107$ & $\pm0.129$ & $\pm0.091$ & $\pm0.329$ & $\pm0.189$ & $\pm0.156$ & $\pm0.069$ & $\pm0.329$ & $\pm0.154$ & $\pm0.081$ & $\pm0.056$ & $\pm0.418$ & $\pm0.160$ & $\pm0.112$ & $\pm0.039$ 
     \\ \midrule
    \multirow{2}{*}{RGBD} & $\textbf{0.960}$ & $0.462$ & $\textbf{0.321}$ & $\textbf{0.152}$ & $0.596$ & $\textbf{0.345}$ & $\textbf{0.228}$ & $0.114$ & $0.325$ & $\textbf{0.259}$ & $\textbf{0.146}$ & $\textbf{0.118}$ & $\textbf{0.422}$ & $\textbf{0.158}$ & $0.180$ & $0.129$ 
    \\
        & $\pm0.046$ & $\pm0.213$ & $\pm0.085$ & $\pm0.100$ & $\pm0.491$ & $\pm0.218$ & $\pm0.151$ & $\pm0.115$ & $\pm0.270$ & $\pm0.226$ & $\pm0.121$ & $\pm0.087$ & $\pm0.428$ & $\pm0.156$ & $\pm0.148$ & $\pm0.092$ 
    \\
    \bottomrule \bottomrule
    \multicolumn{17}{l}{{Results are presented with \texttt{mean}$ \pm $\texttt{standard deviation}. We highlight the best performance, \textit{i.e.}, the highest SR, under each trajectory deviation setting in \textbf{bold}.}}
    \end{tabular}
}
         \label{tab:trajectory-perturbation}
  \vspace{-2mm}
\end{table*}%

\begin{table}[ht]
        \caption{Trajectory estimation error (ATE$\downarrow$ [m]) under sensor misalignment  for RGBD SLAM methods. }
\label{tab:sensor-misalign-perturb}
\centering \setlength{\tabcolsep}{0.8mm}
\resizebox{0.48\textwidth}{!}{
\begin{tabular}{l|c|ccc|ccc}
    \toprule \toprule
     \multirow{2}{*}{\textbf{Method}} & \textbf{Clean} &  \multicolumn{3}{c|}{\textbf{Static Mode}}& \multicolumn{3}{c}{\textbf{Dynamic Mode}} \\ \cmidrule{2-5} \cmidrule{6-8}
     & \textbf{$\Delta=0$} & \textbf{$\Delta=5$} & \textbf{$\Delta=10$} & \textbf{$\Delta=20$} & \textbf{$\Delta=5$} & \textbf{$\Delta=10$} & \textbf{$\Delta=20$}      \\ \midrule
    ORBSLAM3~\cite{orbslam3} & $0.0823$ & $0.0689$ & $0.0661$  & $0.0647$ & $0.0703$ & $0.0773$ & $0.0826$ 
    \\
    iMAP~\cite{imap} & $0.1209$ & $0.4672$ & $0.5344$ & $0.6345$  & $0.5104$ & $0.6803$ & $0.6745$ 
    \\
    Nice-SLAM~\cite{niceslam}& $0.0147$ & $0.3820$ & $0.4062$ & $0.5216$  & $0.5433$ & $0.5548$ & $0.7020$
    \\
     CO-SLAM~\cite{coslam}& $0.0090$ & $0.0520$ & $0.1005$ & $0.1939$ & $0.0740$ & $0.1164$ & $0.2108$ \\

    GO-SLAM~\cite{zhang2023goslam} & $0.0046$ & $\textbf{0.0148}$ & $\textbf{0.0292}$ & $\textbf{0.0646}$  & {$\textbf{0.0151}$} & {$\textbf{0.0297}$} & {$\textbf{0.0650}$} 
    \\
    SplaTAM-S~\cite{keetha2023splatam}& $\textbf{0.0045}$ & $0.0554$ & $0.0629$ & $0.0880$  & $0.0402$ & $0.0645$ & $0.0850$ 
    \\
    \bottomrule  \bottomrule
      \multicolumn{8}{l}{{\textbf{1})  $\Delta$ denotes the misaligned frame interval between RGB and depth streams. }}\\
    \multicolumn{8}{l}{{\textbf{2}) We highlight the best performance, \textit{i.e.}, the lowest ATE, under each }}\\
    \multicolumn{8}{l}{{ severity level of multi-sensor misalignment in \textbf{bold}.}}
    \end{tabular}
}
\vspace{3mm}
\caption{Success rate (SR$\uparrow$) of pose tracking under sensor
misalignment for  RGBD-based ORBSLAM3~\cite{orbslam3}. }
\centering \setlength{\tabcolsep}{0.8mm}
\label{tab:sensor-misalign-perturb_sr_metric}
\resizebox{0.48\textwidth}{!}{
\begin{tabular}{l|c|ccc}
    \toprule \toprule
     \multirow{2}{*}{\textbf{Perturb}} & \textbf{Clean} &  \multicolumn{3}{c}{\textbf{Misaligned Frame Interval ($\Delta$)}} \\ \cmidrule{2-5} 
   \textbf{Mode}  & \textbf{$\Delta=0$} & \textbf{$\Delta=5$} & \textbf{$\Delta=10$} & \textbf{$\Delta=20$}       \\ \midrule
    Static &   \multirow{2}{*}{$\textbf{0.960}\pm 0.046$} & $\textbf{0.960}\pm 0.036$ & $\textbf{0.958}\pm 0.029$  & $\textbf{0.954}\pm 0.030$ 
    \\
    Dynamic &  & $0.955\pm 0.039$ & $0.942\pm 0.050$ & ${0.948}\pm 0.041$ 
    \\
    \bottomrule  \bottomrule
\multicolumn{5}{l}{{\textbf{1}) Results are presented with \texttt{mean}$ \pm $\texttt{standard deviation}.}}\\
\multicolumn{5}{l}{{\textbf{2}) We highlight the best performance, \textit{i.e.}, the highest SR, under each }}\\
\multicolumn{5}{l}{{ severity level of multi-sensor misalignment in \textbf{bold}.}}
    \end{tabular}
}\vspace{-2mm}
\end{table}

\vspace{1.0mm}\noindent{\textbf{Trajectory deviations.}}
In Table~\ref{tab:trajectory-perturbation_sr_metric} and Table~\ref{tab:trajectory-perturbation}, we present the performance analysis under trajectory-level deviations. The degradation caused by translation-only deviation is evident for most advanced models, \textit{e.g.}, a random deviation of 2.5 cm in the translation vector leads to a significant increase in trajectory estimation error. Similarly, for rotation-only deviation, a mere 3-degree random deviation in camera pose leads to an unacceptable ATE score and nearly complete loss of tracking capability for most SLAM models evaluated.
Furthermore, simultaneous introduction of translation and rotation deviations amplifies trajectory estimation error, resulting in the failure of nearly all models. Notably, SplaTAM-S, which generally exhibits robustness under sensor corruptions, encounters failures in 9 out of 15 settings under trajectory-level deviations. Alternately, we observe that ORBSLAM3 and GO-SLAM demonstrate better robustness to trajectory deviation, which can be attributed to the incorporation of loop closure and global bundle adjustment techniques not utilized by other benchmarking methods.
Overall, the accuracy of trajectory estimation in most advanced SLAM models is significantly influenced by locomotion deviations, particularly under high severity of perturbations.

\vspace{1.0mm}\noindent{\textbf{Multi-sensor misalignment perturbation.}}
Table \ref{tab:sensor-misalign-perturb} and Table \ref{tab:sensor-misalign-perturb_sr_metric} present the comparison of trajectory estimation errors and success ratio of tracked trajectory under different severity levels of multi-sensor misalignment, characterized by the frame interval between multiple sensor streams. The performance of iMAP and Nice-SLAM significantly deteriorates as the misalignment intervals increase. In contrast, CO-SLAM, GO-SLAM, and SplaTAM-S demonstrate a certain degree of tolerance towards misalignment. Generally, increasing dis-synchronization frames leads to larger performance drop.

\begin{figure}[t!]
\includegraphics[width=0.48\textwidth]{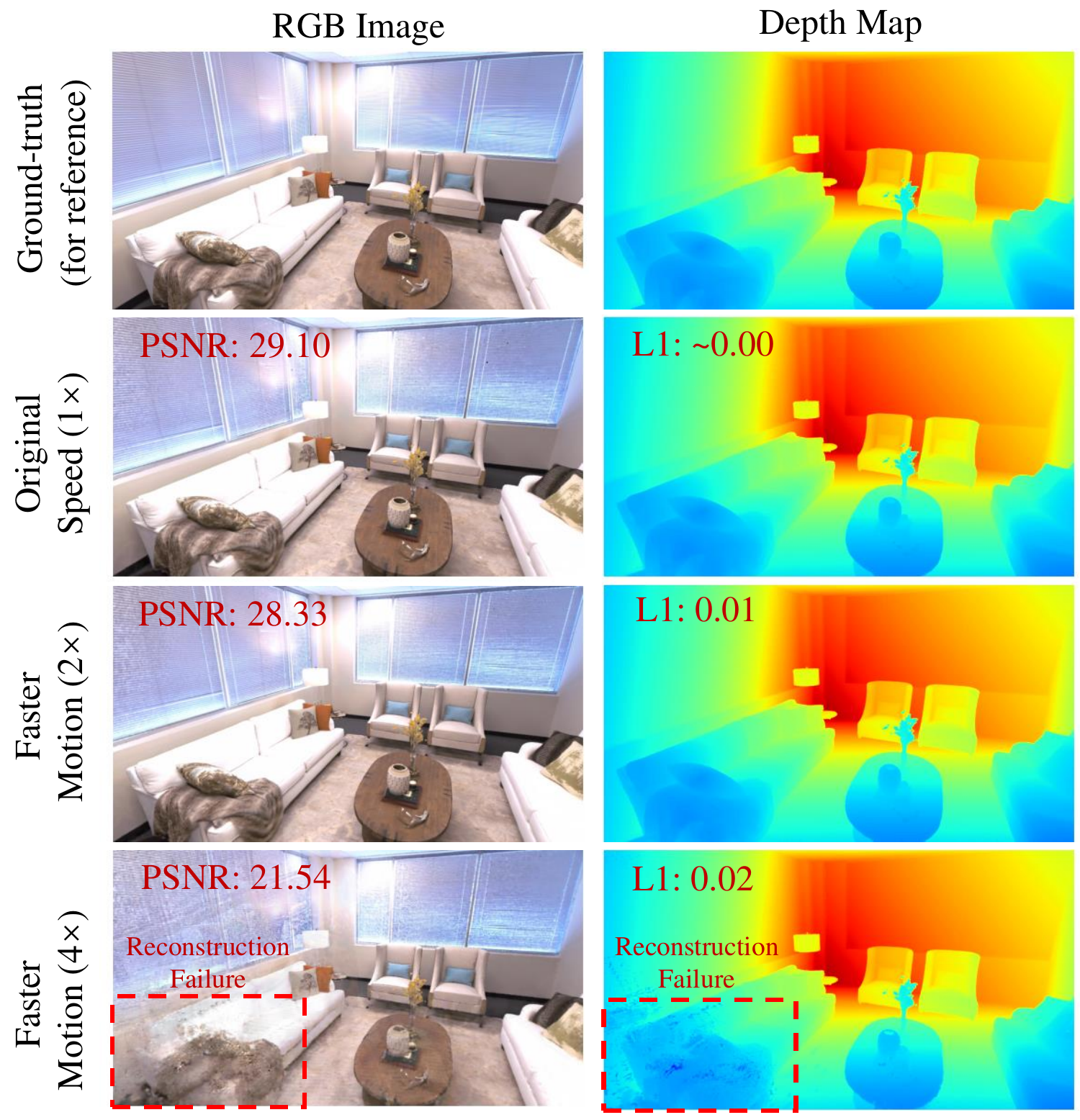} 
	\caption{{Effect of {faster motion}}  {on the reconstruction quality} of RGB images (\textbf{Top}) and depth maps (\textbf{Bottom}), which are measured via PSNR and Depth L1 loss, for SplaTAM-S~\cite{keetha2023splatam}.}
	\label{fig:recon-fast-motion}
 \vspace{3mm}
     \centering
     \includegraphics[width=0.48\textwidth]{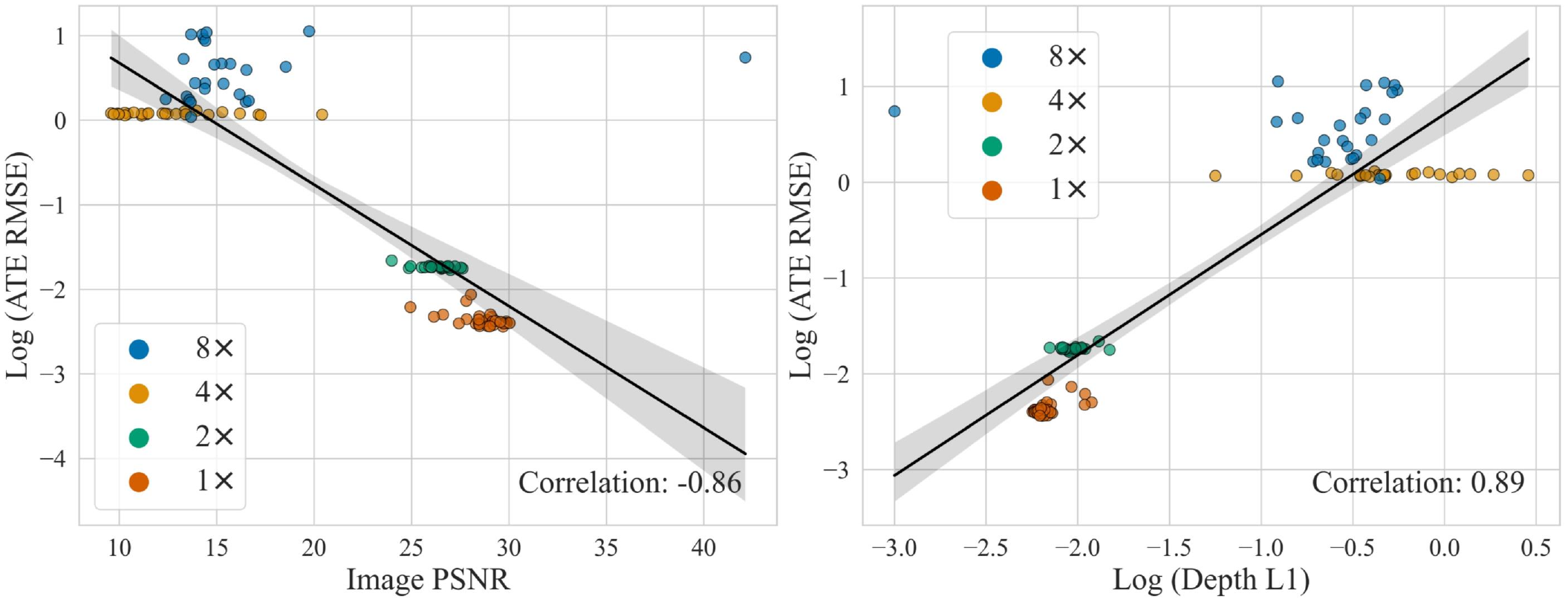} 
	\caption{Correlation between ATE (logarithm form) and reconstruction quality of  {on the reconstruction quality} of RGB images (\textbf{Left}) and depth maps (\textbf{Right}) under faster motion for SplaTAM-S~\cite{keetha2023splatam}. Pearson correlation coefficient~\cite{cohen2009pearson} is reported in the bottom-right corner of each sub-figure. }\vspace{-4mm}
	\label{fig:correlation-fast-motion}\centering
\end{figure}

\subsection{{Discussion}}
{\textbf{\textit{Can one model rule out all perturbed settings?}}} 
In our robustness analysis of existing SLAM models, some methods exhibit specialization in achieving robustness under specific perturbed settings. 
However, a general model that demonstrates robustness against both sensor and trajectory perturbations remains elusive.  Different SLAM methods demonstrate varying performance across different settings, emphasizing the need for tailored approaches for specific scenarios. This finding underscores the importance of selecting the most suitable SLAM method based on the specific requirements and characteristics of the application at hand.

\textit{\textbf{Is there a correlation between a model's performance in clean and perturbed settings?}} We observe a lack of correlation in our analysis, as certain methods that exhibit strong performance under standard clean conditions demonstrate notably inferior performance when subject to specific perturbations. The lack of correlation highlights the importance of evaluating SLAM methods across more diverse perturbed environments to obtain a thorough and robust assessment.

\textit{\textbf{Can a current SLAM model detect degraded observations?}}
We conduct a case study to explore the ability of an advanced RGBD SLAM model SplaTAM-S, to perceive perturbation severity. In Fig.\ref{fig:recon-fast-motion}, we assess the response of SplaTAM-S to faster motion perturbations. It can be observed that more severe perturbations lead to poorer reconstruction quality for the model. Additionally, in Fig.\ref{fig:correlation-fast-motion}, we observe a strong correlation between the accuracy of the final trajectory estimation and the quality of RGBD reconstruction. These findings suggest that the model can potentially identify perturbations during inference, as lower reconstruction loss corresponds to better trajectory accuracy; that is, the reconstruction quality can potentially serve as a useful indicator to predict the final trajectory estimation  error.

\section{Conclusion and Future Work}
\label{sec:conclusions}
\subsection{{Conclusion}}
In this work, we have made several contributions to advance the field of SLAM evaluation under perturbation. 
We presented a comprehensive taxonomy of perturbations in SLAM and introduced a versatile data synthesis pipeline. This pipeline can be utilized to generate customizable perturbed datasets, laying the ground work for rigorous robustness benchmarking. Leveraging this innovative pipeline, we created the \textit{Robust-SLAM} benchmark -- a large-scale initiative designed to assess the resilience of state-of-the-art multi-modal SLAM models against a wide range of perturbations.

Our extensive evaluation has revealed vulnerabilities in current SLAM systems when exposed to various perturbations. These findings not only highlight the limitations of existing models and their potential failures in real-world, unstructured environments, but also offer valuable insights for future research aimed at overcoming these challenges. Through our contributions, we aim to pave the way for the development and evaluation of the next generation of robust SLAM systems, enhancing their performance and reliability in complex environments.

\subsection{{Future Work}}

Future research efforts could aim to unravel the complex interplay between multiple perturbations and evaluate the SLAM model's robustness under mixed perturbations. In addition, a deep dive into regional sensor corruptions~\cite{moseley2021extreme} and adversarial perturbations~\cite{moosavi2017universal} would be valuable as well. Moreover, leveraging generative models~\cite{ramesh2021zero} to enhance the quality of perturbation synthesis could significantly advance our capability to create realistic and challenging testing environments. Furthermore, further exploration on robust SLAM designs, such as sensor correction~\cite{nerfwater} in the front-end of SLAM systems, would provide valuable insights. Such approaches are crucial for mitigating the impact of sensor noise and enhancing the accuracy of the SLAM system. Additionally, exploring the robustness of active SLAM~\cite{actiev-slam} and multi-agent SLAM~\cite{Kimera-Multi} represents an exciting frontier. The active coordination and sensor fusion among multiple robots have the potential to effectively handle noise and elevate SLAM performance.

\setcounter{table}{0}  
\setcounter{figure}{0}  
\setcounter{section}{0}  
\renewcommand{\thetable}{\Alph{table}}
\renewcommand{\thefigure}{\Alph{figure}}
\renewcommand{\thesection}{\Alph{section}}
\begin{figure*}[t]
	\centering
\includegraphics[width=\textwidth]{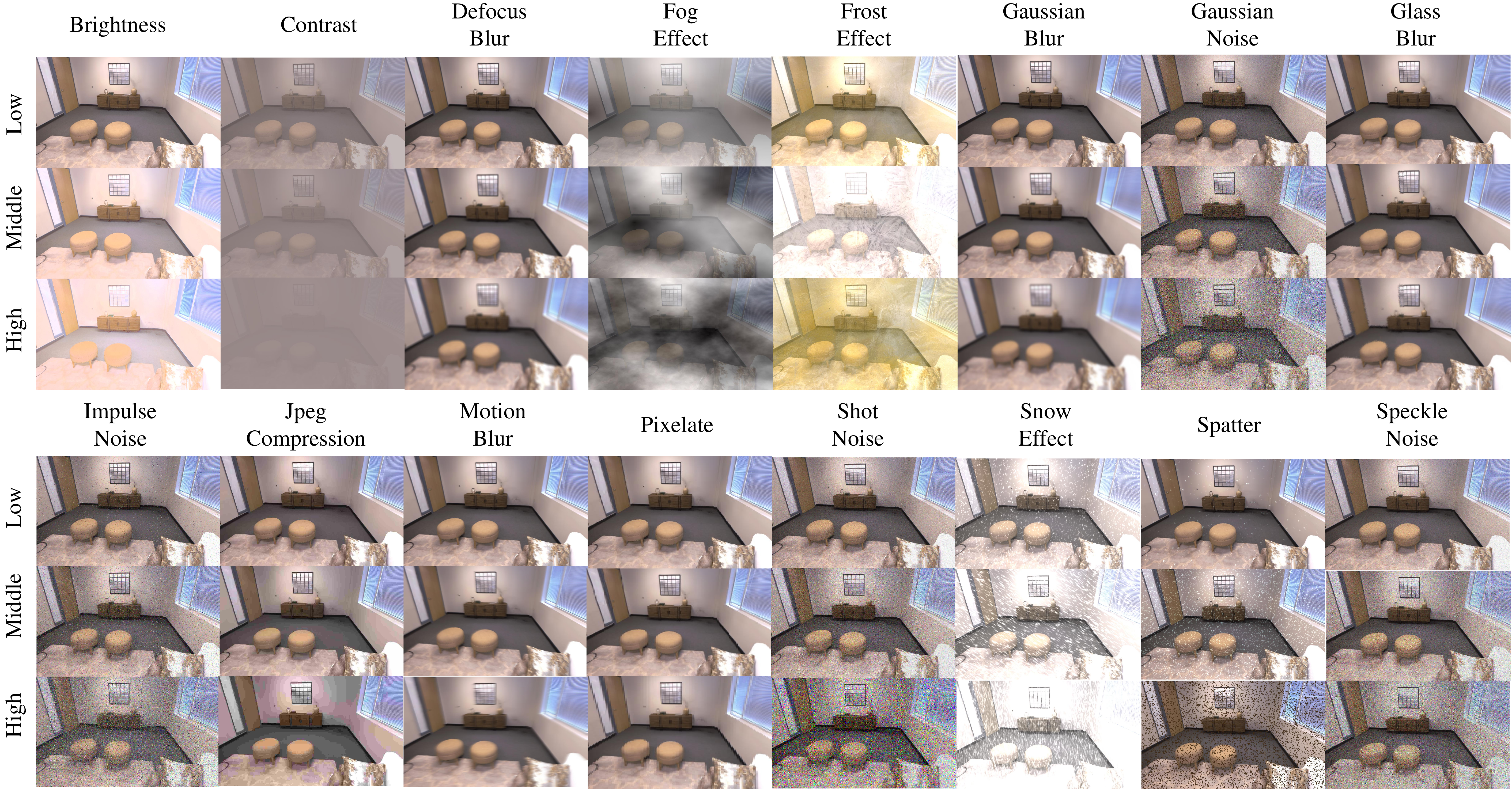} 
	\caption{
\textbf{Illustration of sensor-level perturbations for the RGB imaging under different severity levels}. We consider \textbf{16} common image corruption types~\cite{hendrycks2019robustness} from \textbf{\textit{4}} main categories of perturbations for SLAM robustness evaluation: \textbf{(1)} \textbf{{noise-based distortions}}: Gaussian noise, shot noise, impulse noise, and speckle noise; \textbf{(2)} \textbf{{blur-based effects}}: defocus blur, glass blur, motion blur, and Gaussian blur; \textbf{(3)} \textbf{{environmental interferences}}: snow effect, frost effect, fog effect, and spatter effect. \textbf{(4)} \textbf{{post-processing manipulations}}: brightness, contrast, pixelate, and JPEG compression. Each perturbation type is further split into \textbf{\textit{3}} severity levels (low, middle, and high) for the instantiated SLAM robustness benchmark, \textit{i.e.}, \textit{Robust-SLAM}.  } 
	\label{fig:sensor-corruption-full}
\end{figure*}

\section*{Acknowledgment}
The authors would like to thank lab-mates at Hybrid Dynamic Robotics Lab of UMich Robotics for help and support. The authors appreciate the GPU resources provided by Prof. Ram Vasudevan from UMich Robotics. The authors are grateful to Dr. Lu Li from CMU Biorobotics Lab, Prof. Siheng Chen from SJTU, Dr. Wenshan Wang from CMU AirLab, and Dr. Youmin Zhang from University of Bologna for valuable discussions. The authors express their gratitude to Prof. Sara Nezami Nav, Prof. Pamela Bogart,   Mrs. Lucy Kates, Mr. Todd Maslyk from UMich ELI; Prof. Julie Babcock, Prof. Raymond McDaniel, and Prof. Dave Karczynski from UMich Sweetland Center for proofreading.



\appendices

\section{Appendix Overview}\label{sec:appendix}

\noindent\textbf{Appendix overview. }The appendix is organized as follows:

\begin{itemize}
\item Sec.~\ref{sec:perturbation-taxonomy} provides additional details about the perturbation taxonomy for the evaluation of SLAM under perturbation.
\item Sec.~\ref{sec:benchmark-setup} presents more information about the setup and statistics of the instantiated \textit{Robust-SLAM} benchmark.
\item Sec.~\ref{sec:additional-results} offers supplementary results and discussions based on the experiments conducted on the \textit{Robust-SLAM} benchmark.
\item Sec.~\ref{sec:stereo-slam-case-study} presents the case study of the robustness analysis on the multi-view SLAM system, specifically under the stereo and stereo-inertia settings. This case study highlights the importance of considering the robustness of both the hardware and software aspects of the SLAM system in a holistic manner.
\item Sec.~\ref{sec:multi-agent-slam-robustness} provides preliminary robustness analyses on the multi-agent SLAM system, aiming to provide insights into extending the robustness evaluation to more heterogeneous and diverse SLAM systems.
\item Sec. \ref{sec:qualitative_results} presents the qualitative results of SLAM models under perturbations, including the trajectory estimation and 3D reconstruction prediction results of both successful cases and failure cases.
\item Sec.~\ref{sec:more future work} provides a brief introduction to additional future directions that can advance the frontier of robustness evaluation for SLAM systems under perturbation.
\end{itemize}

\noindent\textbf{Video demo}. We provide a {{video demo}} in the supplementary material that exhibits the visualization of the synthesized noisy data for SLAM robustness evaluation, which includes both sensor and trajectory perturbations. The video also showcases qualitative per-frame prediction results of advanced SLAM models when subjected to perturbations, illustrating both successful cases and failure cases.

\begin{figure*}[ht!]
	\centering
	\includegraphics[width=\textwidth]{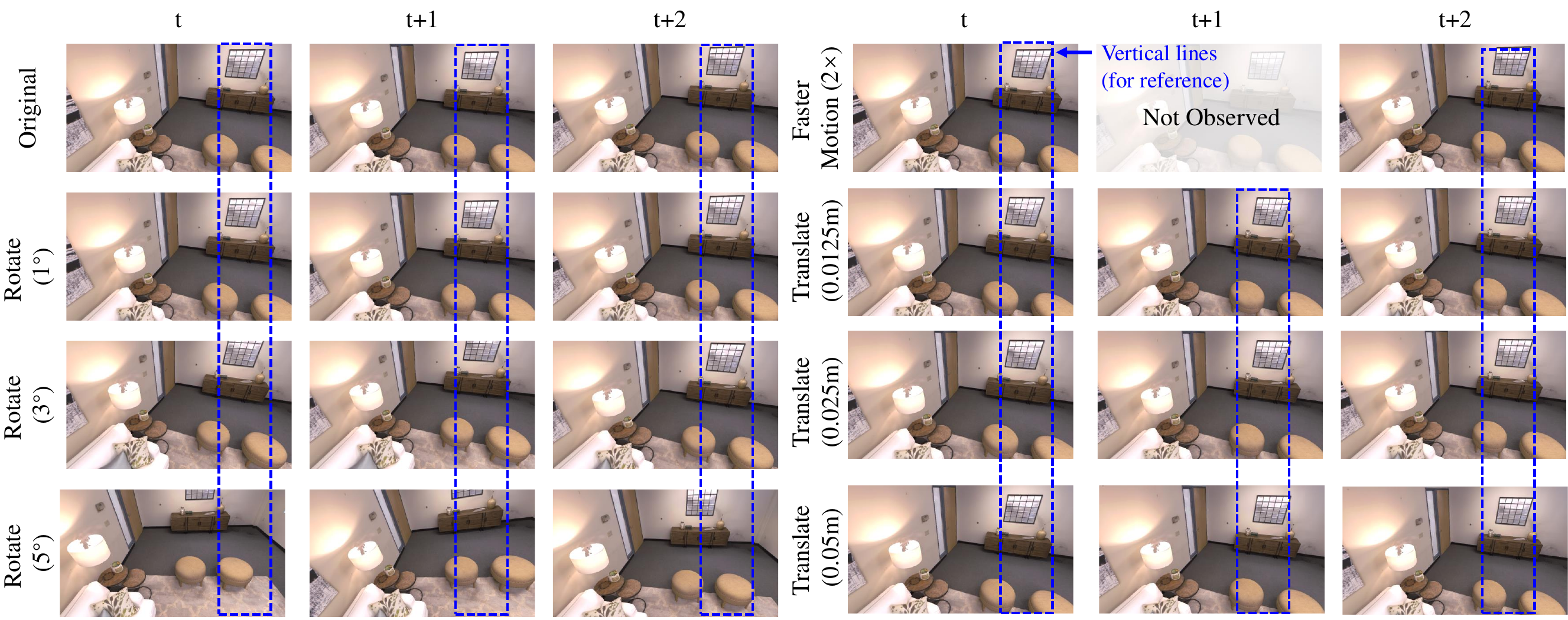} \caption{\textbf{{Rendered RGB image streams under trajectory-level perturbations}}, including translation deviations (Translate), rotation deviations (Rotate), and the faster motion effect. } 
	\label{fig:trajectory-perturbation-illustration}
\end{figure*}
\begin{figure*}[t!]
	\centering
	\includegraphics[width=\textwidth]{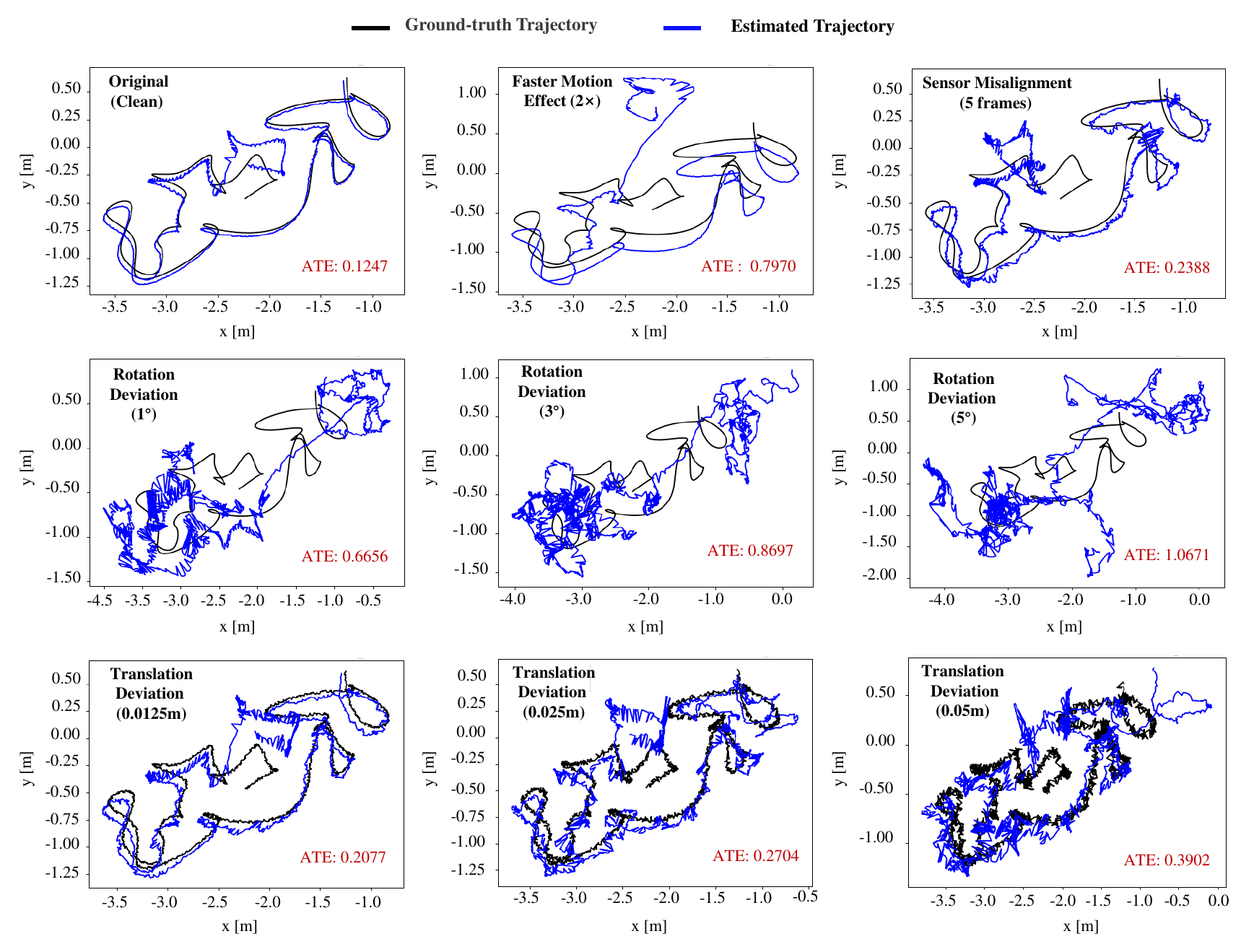} 
	\caption{\textbf{Illustration of trajectory deviations under different severity levels, the faster motion effect, and the multi-sensor misalignment effect.} We also show qualitative results of an advanced real-time SLAM model, \textit{i.e.}, CO-SLAM~\cite{coslam}. The impact of translation deviation, rotation deviation, {faster motion}, and sensor misalignment perturbations is assessed. For clarity, we visualize the projected trajectory on the horizontal x-y plane derived from the 3D trajectory, which shows that slight trajectory deviations can have a  significant impact on the trajectory estimation performance, measured by ATE.}
	\label{fig:taxonomy-trajectory-misalign-perturbation}
\end{figure*} 

\section{More Details about Perturbation Taxonomy  }\label{sec:perturbation-taxonomy}

\subsection{{Sensor-level Perturbation on RGB Imaging }}
\noindent\textbf{Illustration of varied severity levels of sensor corruptions on RGB Imaging}. 
Our work considers a comprehensive taxonomy of image-level corruptions specifically designed to evaluate the robustness of SLAM algorithms. The taxonomy encompasses 16 common image corruption types, categorized into four distinct groups: noise-based distortions, blur-based effects, environmental interferences, and post-processing manipulations. To facilitate a rigorous evaluation, we follow the robustness evaluation setup for vision tasks~\cite{hendrycks2019robustness} to further split each perturbation type into three severity levels in our instantiated \textit{Robust-SLAM} benchmark. In Fig.~\ref{fig:sensor-corruption-full}, we show the perturbed RGB images under different severity levels of image corruptions. Although certain types of image corruption (\textit{e.g.}, motion blur) only lead to slight changes in appearance, they might lead to significant performance degradation for several advanced multi-modal SLAM models, as demonstrated in the benchmarking results section of the main paper.

\subsection{{Trajectory-level Perturbation}}

\noindent{\textbf{Illustration of the rendered sensor streams under trajectory perturbations. }} In Fig.~\ref{fig:trajectory-perturbation-illustration}, we present the rendered image streams under varying severity levels of trajectory-level perturbations, encompassing translational deviations, rotational deviations, and faster motion effects. Although the rotational and translational deviations we examined result in minor changes in observations between adjacent frames, these perturbations lead to significant performance degradation across the majority of benchmarking SLAM models. As depicted in Fig.~\ref{fig:taxonomy-trajectory-misalign-perturbation}, even slight trajectory-level deviations can have a substantial impact on trajectory estimation performance.

\begin{figure}[t]
	\centering
\includegraphics[width=0.28\textwidth]{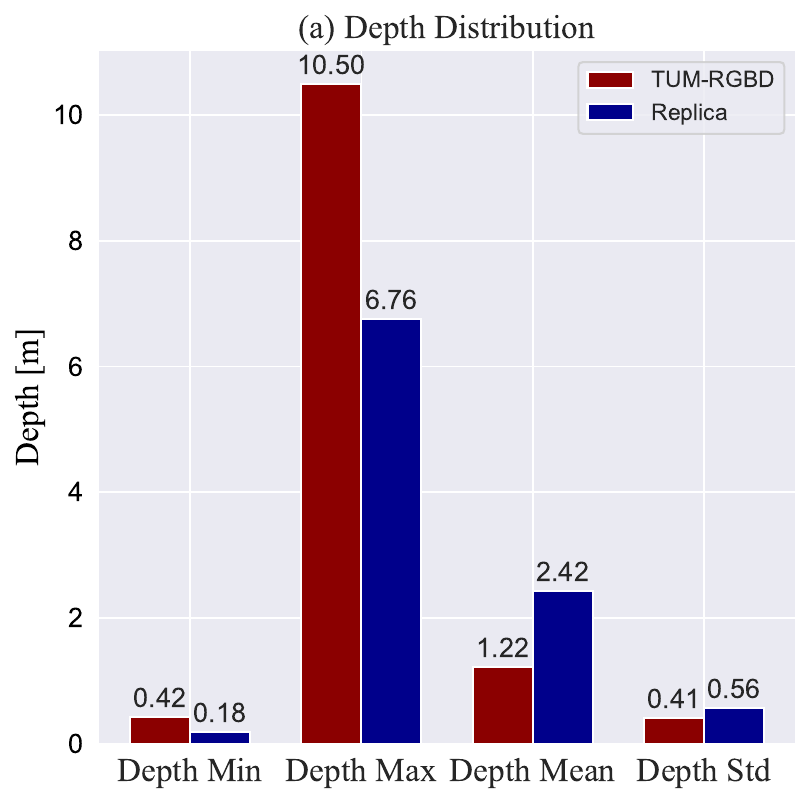}  \hfill 
\includegraphics[width=0.2\textwidth]{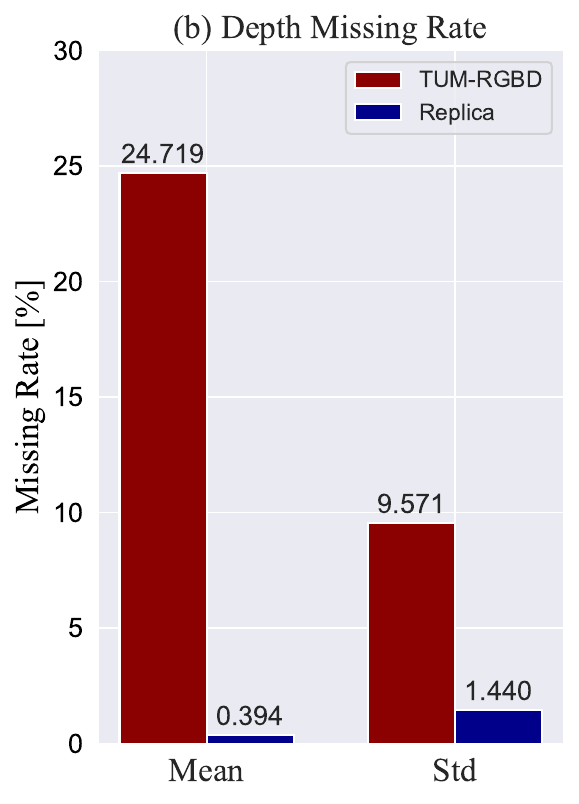} 
	\caption{\textbf{Discrepancy in depth characteristics between current simulated and real-world SLAM datasets.} Here, we compare the depth (\textbf{a}) distribution and (\textbf{b})  missing rate between real-world collected depth from the TUM-RGBD~\cite{TUM-RGBD} dataset and simulated depth from the Replica~\cite{straub2019replica} dataset.}
	\label{fig:depth_distribution_comparison}
\end{figure}
\subsection{{Sensor-level Perturbation on Depth Imaging }}

\noindent{\textbf{Motivation for depth perturbation.}} As depicted in Fig.~\ref{fig:depth_distribution_comparison}, there exists a noticeable disparity between the current simulated clean depth distribution obtained from the Replica~\cite{straub2019replica} SLAM benchmark and the real noisy depth data derived from the TUM-RGBD~\cite{schubert2018tum} SLAM benchmark. In Replica, the minimum depth measures approximately 0.18 m, whereas the TUM-RGBD data exhibits a minimum depth value of 0.4 m, reflecting the limitations of real-world depth sensors. Notably, we observe a significant discrepancy in the depth missing rates, with TUM-RGBD demonstrating an approximate 25\% missing rate compared to nearly zero (0.39\%) in Replica. These observations underscore the necessity of exploring perturbation strategies for depth imaging to bridge the gap between simulated depth and real-world depth.

\section{More Details of \textit{Robust-SLAM} Benchmark}\label{sec:benchmark-setup}
\subsection{\textbf{\textit{Robust-SLAM} Benchmark Statistics}}

\begin{figure*}[t!]
	\centering
\includegraphics[width=0.495\textwidth]{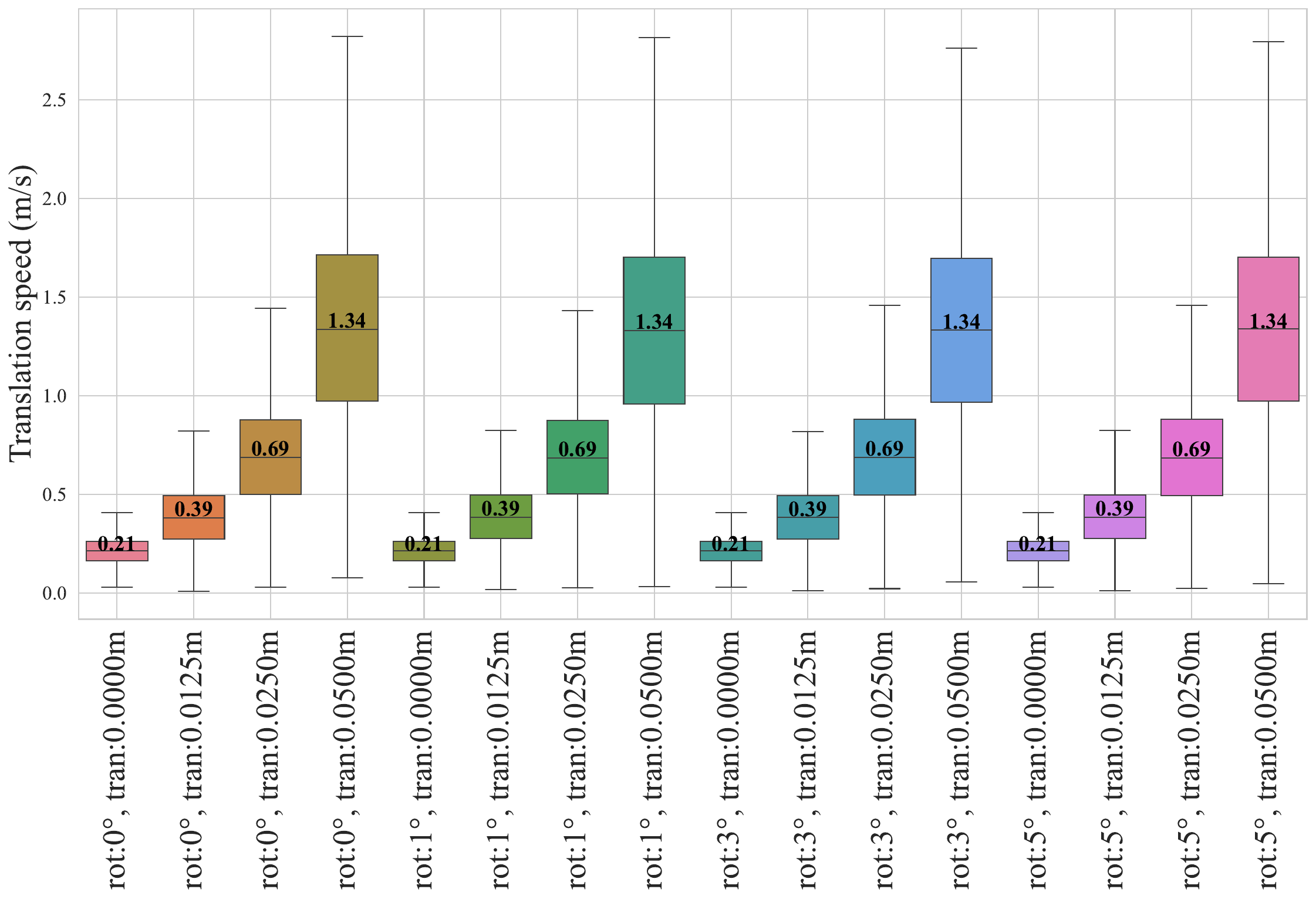}
\includegraphics[width=0.495\textwidth]{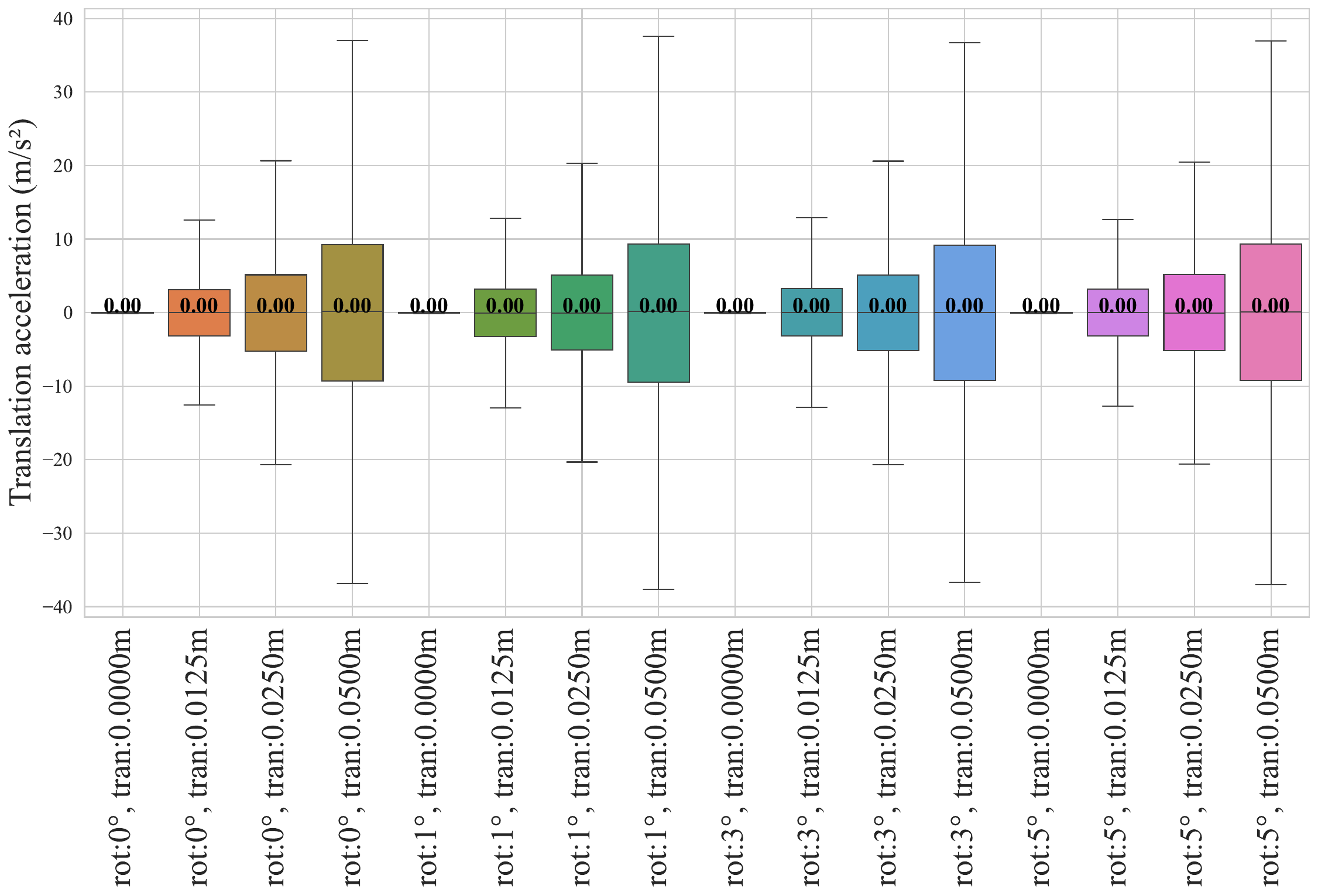}
\includegraphics[width=0.495\textwidth]{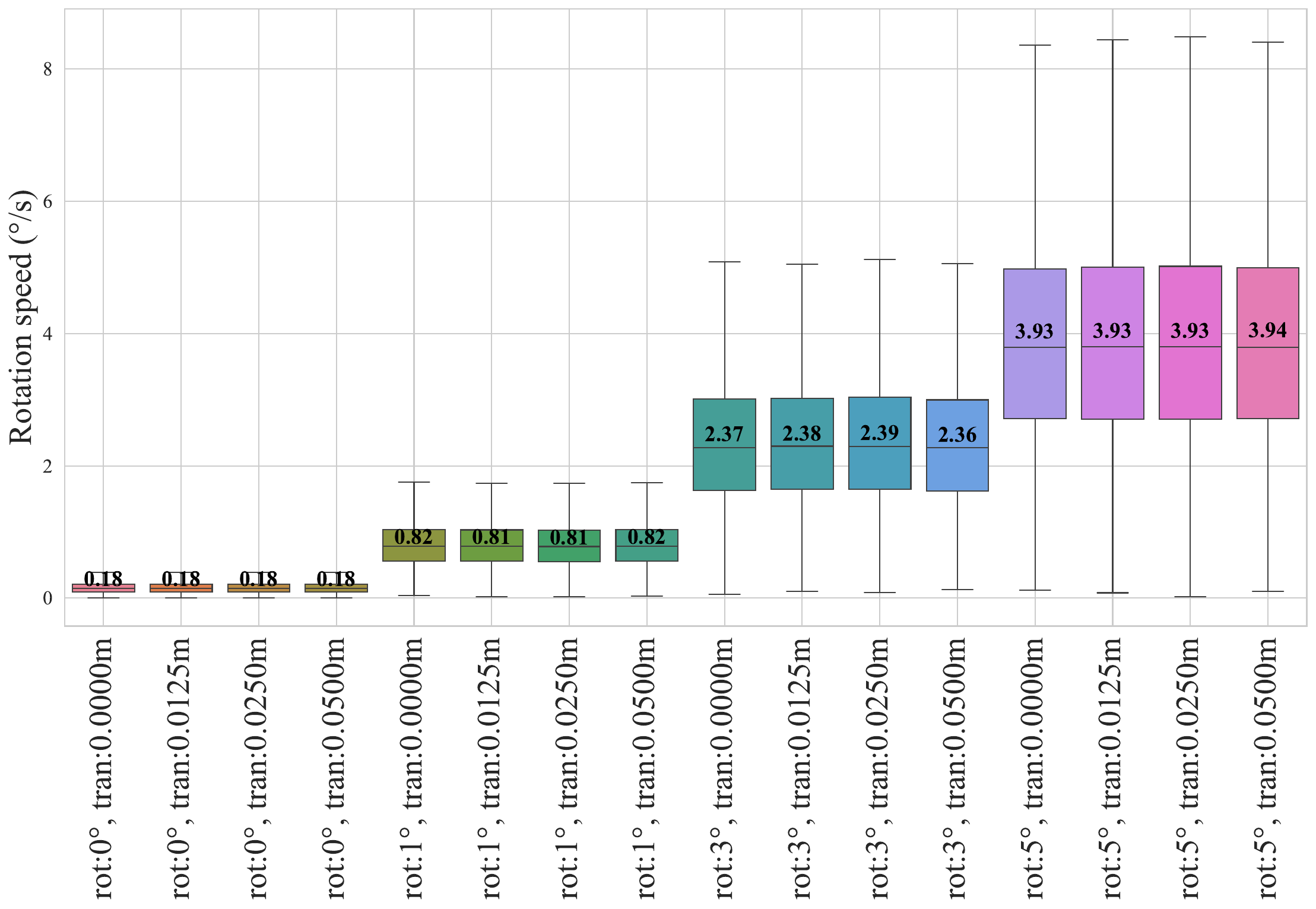}
\includegraphics[width=0.495\textwidth]{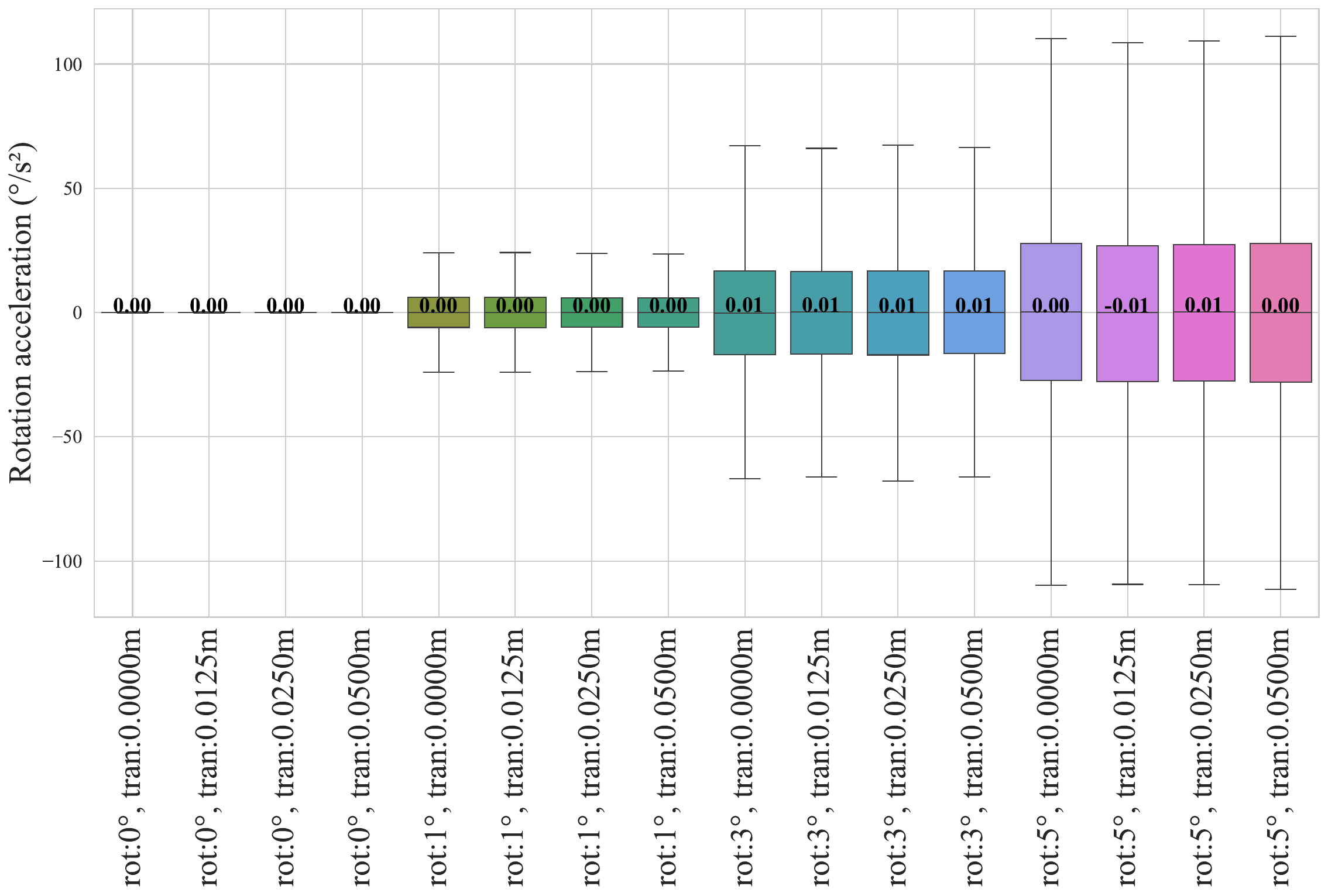}
\caption{\textbf{Motion statistics of trajectory distribution under varying combinations of translation and rotation deviations}. Assuming a frame rate of 20 frames per second for the SLAM system, \textit{i.e.}, a time interval of 0.05 seconds between neighboring pose frames, we present the motion distribution of perturbed trajectories in our \textit{Robust-SLAM} benchmark. The figures show the distribution of translation speed (\textbf{Top Left}), translation acceleration (\textbf{Top Right}), rotation speed (\textbf{Bottom Left}), and rotation acceleration (\textbf{Bottom Right}). We report the mean value of each setting.}
	\label{fig:traj-motion-distribution}
\end{figure*}

\noindent{\textbf{Benchmark sequence number distribution.}} Using our established taxonomy of perturbations for SLAM and the noisy data synthesis pipeline, we have created a large-scale SLAM robustness benchmark called \textit{Robust-SLAM} to evaluate the robustness of monocular and multi-modal (RGBD) SLAM methods by incorporating various perturbations that mimic real-world sensor and action anomalies. We provide details about the specific distribution of sequence numbers for each perturbation type as follows:

\begin{enumerate}
  \item \textbf{8 original clean sequences:} These sequences replicate the quality and the sequence number of the original Replica SLAM dataset~\cite{imap}.
  \item \textbf{768 sequences with image-level perturbations:} We apply 16 different types of image-level perturbations at 3 severity levels, both under static and dynamic conditions.
    \item \textbf{32 sequences with depth-level perturbations:} This category consists of 4 types of perturbations. For the depth noise, we adopt the hyperparameters of the Gaussian noise distribution as specified in previous literature~\cite{hendrycks2019robustness}. Moreover, we set the depth missing rate to 10\% and establish the depth clipping range based on the real-world depth distribution of the TUM-RGBD dataset~\cite{TUM-RGBD}, with a minimum value of 0.42 meters and a maximum value of 10 meters.
  \item \textbf{24 sequences with increased motion levels:} These sequences involve faster speed than the original sequences, with variations of two, four, and eight times the original speed.
  \item \textbf{120 sequences with trajectory-level deviations:} This category includes pure rotation deviation, pure translation deviation, and combined transformation matrix deviation. We define three severity levels for both rotation and translation deviations, and sample the deviation from a Gaussian distribution. Specifically, for rotation deviation, we introduce random deviations in rotation around the x, y, and z axes, with mean values of zero and standard deviations of 1, 3, and 5 degrees at each pose frame. For translation deviation, we introduce random deviations in the x, y, and z axes, with mean values of zero and standard deviations of 0.0125, 0.025, and 0.05 meters at each pose frame. In Fig. \ref{fig:traj-motion-distribution}, we show the motion statistics of the perturbed trajectory sequences under varying combinations of translation and rotation deviations. This category of trajectory-deviated sequences encompasses a broad spectrum of motion speeds and accelerations, enabling a progressive evaluation of the robustness of SLAM models against increasingly challenging motion types. These insights are especially valuable for evaluating the implementation of SLAM systems in high-speed scenarios or on agile robot platforms exposed to significant vibrations.
  \item \textbf{48 sequences with multi-sensor misalignment:} We consider both static and dynamic perturbation models for multi-sensor misalignment. In the static mode, a constant time delay is synthesized between the two sensor streams, while in the dynamic perturbation model, there is a varying time delay between the streams. Specifically, the multi-sensor misalignment perturbation sequences consist of 24 sequences with a fixed cross-sensor frame delay interval ($k$) of 5, 10, and 20 frames, as well as 24 sequences with dynamic perturbation where $k$ deviates by 1 frame from the fixed intervals of 5, 10, and 20 frames.
\end{enumerate}

Overall, this benchmark dataset enables a comprehensive evaluation of existing SLAM algorithms under simulated perturbations, providing a thorough assessment of the robustness of multi-modal SLAM systems in a wide range of challenges.

\subsection{\textbf{ SLAM Models for Benchmarking}}
Below, we offer additional descriptions of SLAM models that have been benchmarked on our \textit{Robust-SLAM} benchmark.

\textbf{1}) ORB-SLAM3~\cite{orbslam3} is an extension of ORB-SLAM2~\cite{orbslam2} that incorporates a multi-map system and visual-inertial odometry, enhancing robustness and performance.

\textbf{2}) iMAP~\cite{imap} is a neural RGBD SLAM system that utilizes the MLP representation to achieve joint tracking and mapping.

\textbf{3}) Nice-SLAM~\cite{niceslam} is a neural RGBD SLAM model that employs a multi-level feature grid for scene representation, reducing computational overhead and improving scalability.

\textbf{4}) CO-SLAM~\cite{coslam} is an advanced neural RGBD SLAM system with a hybrid representation, enabling robust camera tracking and high-fidelity surface reconstruction in real time.

\textbf{5}) GO-SLAM~\cite{zhang2023goslam} is a neural visual SLAM framework for real-time optimization of poses and 3D reconstruction. It supports both monocular and RGBD input settings.

\textbf{6}) SplaTAM~\cite{keetha2023splatam} is a neural RGBD SLAM model that follows Gaussian Splatting~\cite{gaussiansplatting} to construct an adaptive map representation based on Gaussian kernels. Due to the time and computational constraints, we evaluate the relatively more efficient SplaTAM-S model variant in our benchmark.

\section{More Results and Discussions}\label{sec:additional-results}

\subsection{\textbf{More Evaluation Metrics}}

\vspace{1.0mm}\noindent\textbf{Localization: cumulative success rate (CSR).} 
Cumulative success rate  metric is used to evaluate the adherence to predefined trajectory estimation accuracy for SLAM systems, defined as:
\begin{equation}
\text{CSR}(\xi) = \frac{\#(\text{ATE} \leq \xi)}{\#(\text{Total})} \times 100 \quad
\end{equation}
Here, $\text{CSR}(\xi)$ represents the cumulative success rate [\%] for a specific threshold $\xi$ of ATE metric; $\#(\text{ATE} \leq \xi)$ counts the trajectories with an ATE within the defined threshold, and $\#(\text{Total})$ represents the total number of evaluated settings.

\vspace{1.0mm}\noindent\textbf{Mapping: metrics for 3D Reconstruction.} We follow the mapping quality evaluation protocol in~\cite{coslam} to assess 3D reconstruction using Accuracy (ACC) [cm], Completion (Comp.) [cm], and Completion Ratio (Comp. R.) [\%] with a 5 cm threshold. Table~\ref{tab:3d-metric} details the definition for each of these metrics. Note that only certain dense SLAM models can produce 3D reconstruction results for further evaluation of mapping quality.

\begin{table}[t]
\caption{\textbf{Definitions of 3D metrics for evaluation of mesh reconstruction quality} of the reconstructed 3D mesh $P$ when given the ground-truth 3D mesh $Q$ (in the scale of meter [m]). We follows the 3D reconstruction metrics defined in the CO-SLAM~\cite{coslam} paper.}
\begin{center}
\resizebox{0.48\textwidth}{!}{
\begin{tabular}{l|c}
\toprule
\textbf{3D Reconstruction Metric} & \textbf{Definition} \\
\midrule
Accuracy (ACC) & $\frac{1}{|P|} \sum_{p \in P} \left(\min_{q \in Q} {{||p - q||{}^{2}}}\right)$ \\
Completion (Comp.) & $\frac{1}{|Q|} \sum_{q \in Q} \left(\min_{p \in P} {{||p - q||{}^{2}}}\right)$ \\
Completion Ratio (Comp. R.) & $\frac{1}{|Q|} \sum_{q \in Q} \left(\min_{p \in P} {{||p - q||{}^{2}}} \leq {0.05}\right)$ \\
\bottomrule
\end{tabular}\label{tab:3d-metric}}
\end{center}
\end{table}

\begin{figure*}[ht]
	\centering
 \includegraphics[width=0.495\textwidth]{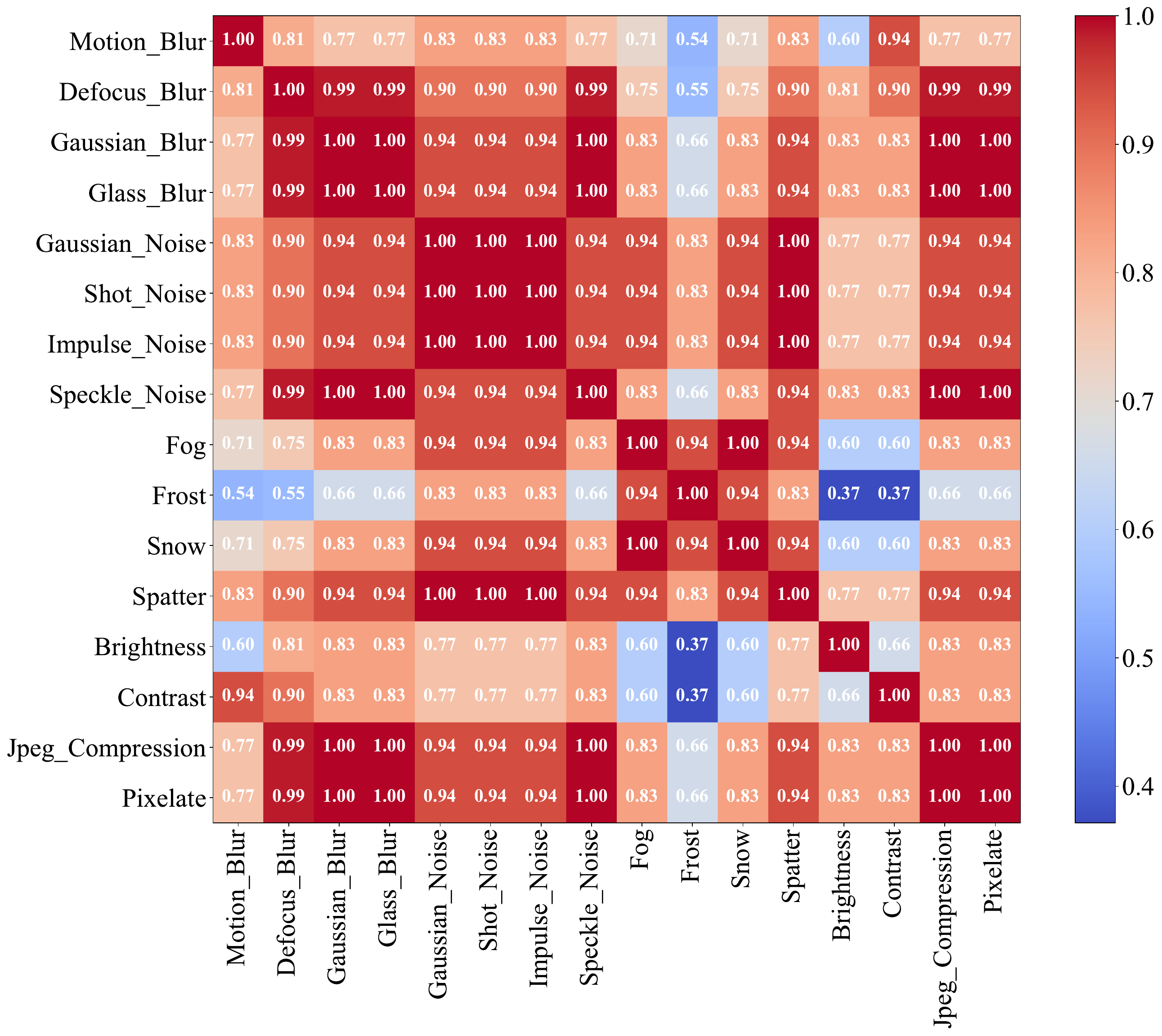} 
 \hfill 
\includegraphics[width=0.495\textwidth]{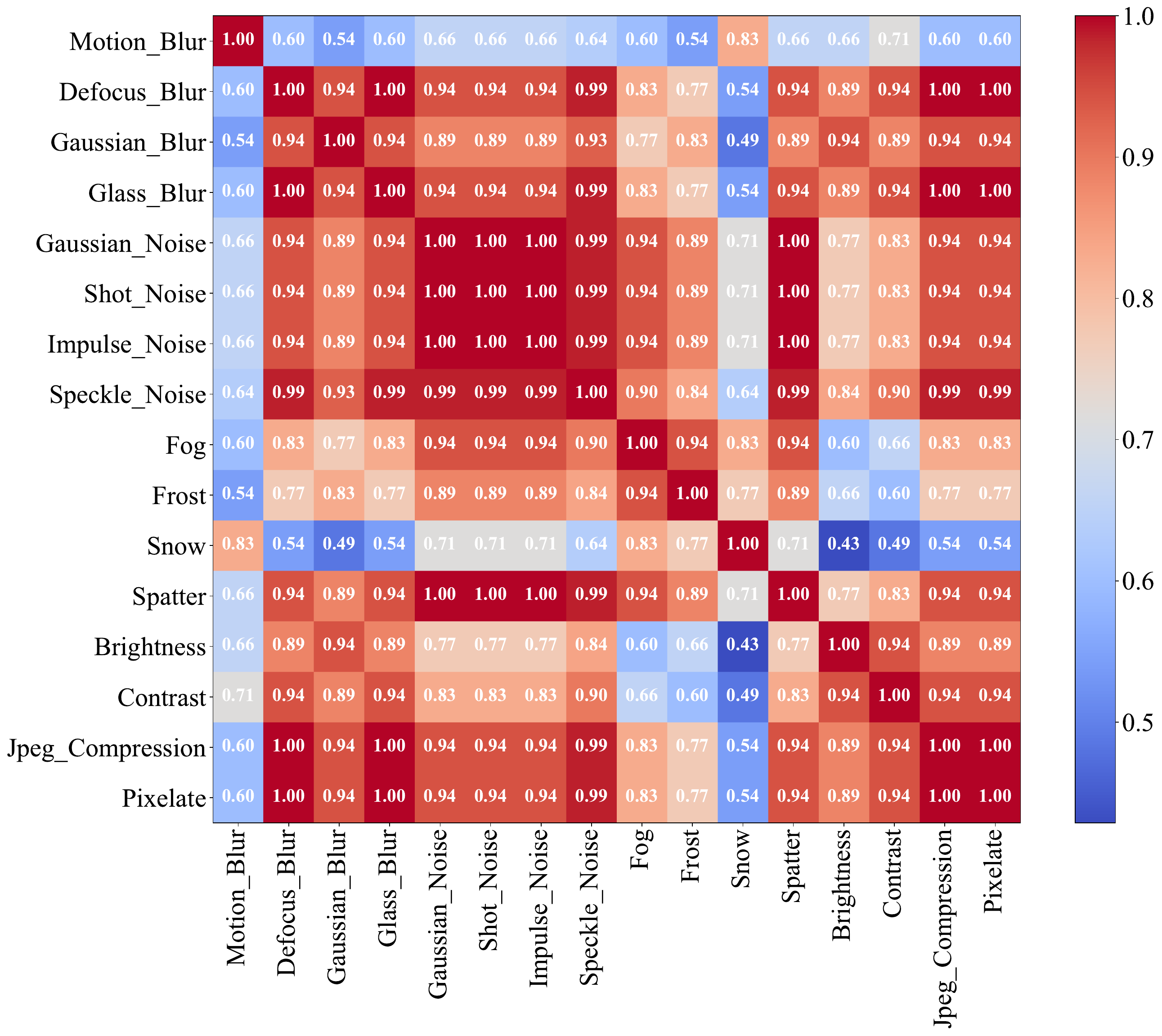} 
	\caption{\textbf{Correlation of perturbed performance (ATE) of multi-modal (RGBD) SLAM models across different image perturbation types} under static (\textbf{Left}) and dynamic (\textbf{Right}) perturbation mode. The pair-wise correlation strength is quantified via Spearman’s rank correlation coefficient~\cite{spearman1961proof}.}
	\label{fig:correlation_perturbed_image_ate-type}
\end{figure*}

\begin{figure*}[t]
	\centering
\includegraphics[width=0.44\textwidth]{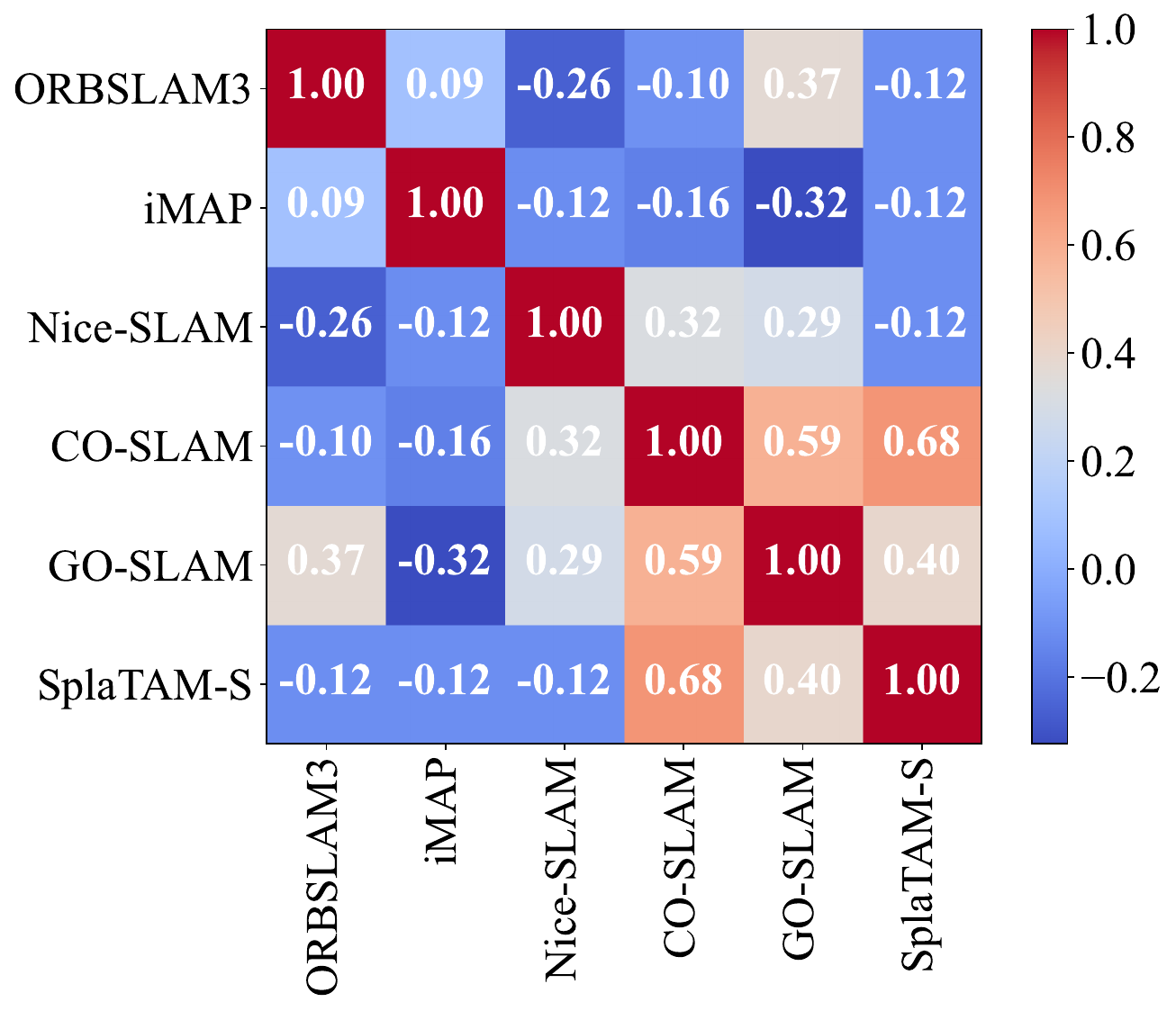} 
\includegraphics[width=0.44\textwidth]{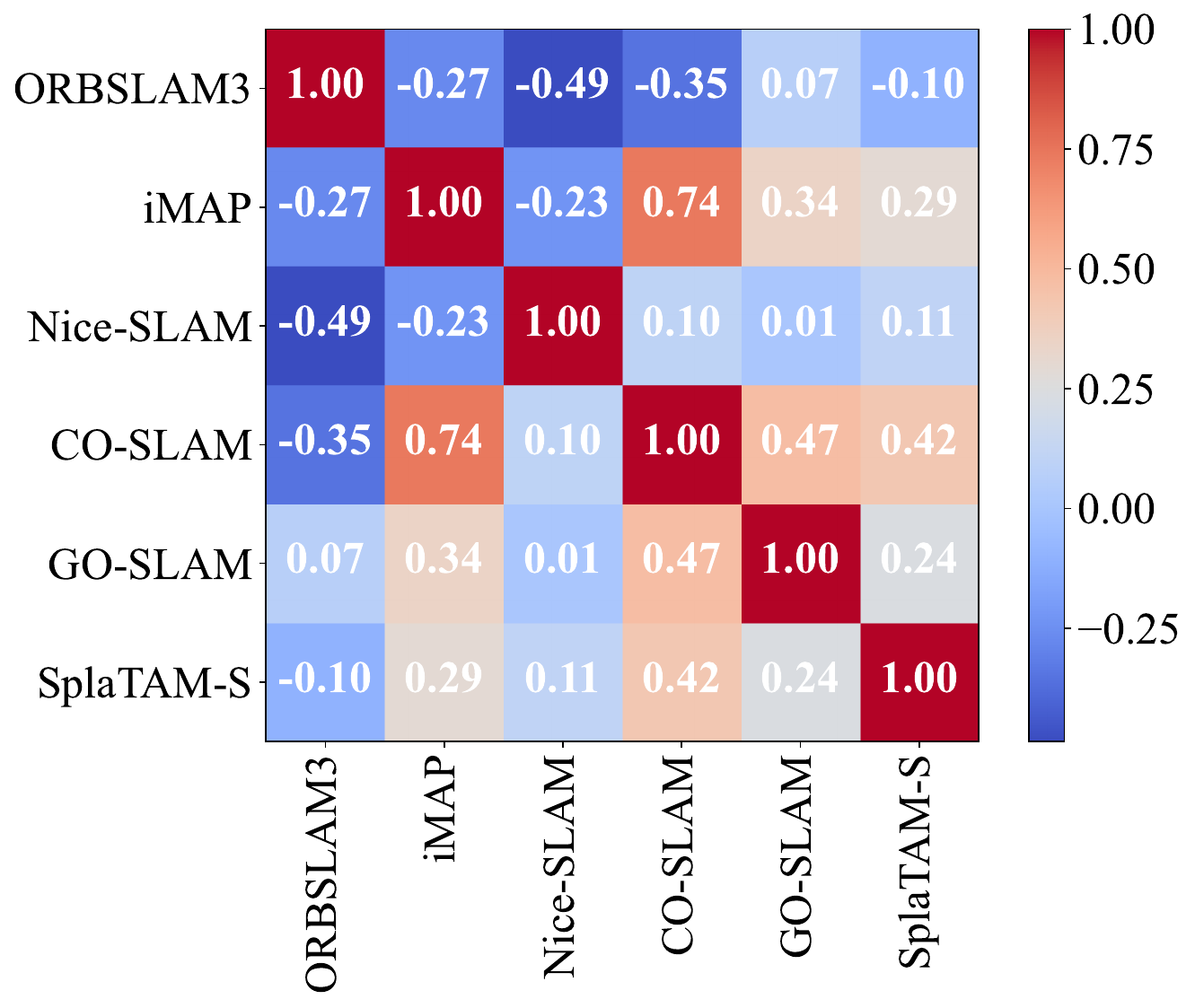} 

	\caption{\textbf{Correlation of perturbed performance (ATE) across different multi-modal (RGBD) SLAM models} under static (\textbf{Left}) and dynamic (\textbf{Right}) image perturbation. The pair-wise correlation strength is quantified via Spearman’s rank correlation coefficient~\cite{spearman1961proof}.}
	\label{fig:correlation_perturbed_image_ate-method}
\end{figure*}

\subsection{\textbf{More Benchmarking Analyses}}

\begin{table}[t]
\caption{\textbf{Effects of image perturbation on 3d reconstruction (mapping) quality} for CO-SLAM~\cite{coslam} model.}
\label{tab:mapping-coslam}
\centering 
\setlength{\tabcolsep}{1.0mm}
\resizebox{0.48\textwidth}{!}{
\begin{tabular}{l|c|ccc|c}
    \toprule \toprule   
    \multirow{2}{*}{\textbf{Metrics}} &     {\textbf{Clean}}
     &  {\textbf{Low}}   & {\textbf{Middle}} & 
  {\textbf{High}}  & {\textbf{Perturb.}}     \\  
 &  \textbf{Mean}& \textbf{Severity} & \textbf{Severity} & \textbf{Severity} & \textbf{Mean}  \\ \midrule
   ACC$\downarrow$ [cm] & $\textbf{2.08}$ & $2.11$ & $2.12$ & $2.39$ & $2.21$ 
    \\
    Comp.$\downarrow$ [cm] & $\textbf{2.17}$ & $2.19$ & $2.20$ & $2.89$  & $2.43$ 
    \\
    Comp. R.$\uparrow$ [\%] & $\textbf{93.13}$ & $93.07$ & $93.04$ & $92.34$  & $92.82$ \\
    \bottomrule \bottomrule
    \multicolumn{6}{l}{{\textbf{1}) The setting with the best performance for each metric is in \textbf{bold}.}}\\
    \multicolumn{6}{l}{\textbf{2}) We compare the performance under no perturbation (clean) and  }\\
    \multicolumn{6}{l}{static image perturbations with different severity levels.}
    \end{tabular}
}
\end{table}

\textbf{\textit{Is there a correlation in the performance after perturbation among different image perturbation types?}} 
In Fig.~\ref{fig:correlation_perturbed_image_ate-type}, a strong correlation is observed in the combined perturbed performance vector of all evaluated RGBD SLAM models for the majority of perturbation types. This finding suggests that the models' performance remains consistent across certain perturbation scenarios. Additionally, the correlation suggests the presence of underlying similarities in the effects of some sub-categories of image perturbation types, \textit{e.g.}, noise effects.

\textbf{\textit{Is there a correlation in the performance under RGB image perturbation among different methods?}} In Fig.~\ref{fig:correlation_perturbed_image_ate-method}, a weak correlation is observed in the combined perturbed performance vector, which encompasses sixteen image perturbation types across six SLAM models with the RGBD input setting. This suggests a large divergence in the distribution of perturbed performance among the different SLAM models.

\textbf{\textit{How do image perturbations influence the mapping quality?}} In Fig.~\ref{tab:mapping-coslam}, we evaluate the impact of image perturbations on the mapping quality of the CO-SLAM~\cite{coslam} model, which shows a strong robustness to most of the image-level corruptions. The results reveal a direct correlation between perturbation severity and both 3D reconstruction error and completion error. Specifically, the clean setting achieves the highest accuracy (2.08 cm) and the lowest completeness score (2.17 cm), while the high perturbation severity setting exhibits the highest errors in ACC (2.39 cm) and completion (2.89 cm). Overall, our analysis shows that increasing severity levels of perturbation lead to larger errors in the reconstructed 3D map.

\begin{figure*}[t]
	\centering
 \vspace{1mm}
\includegraphics[width=\textwidth]{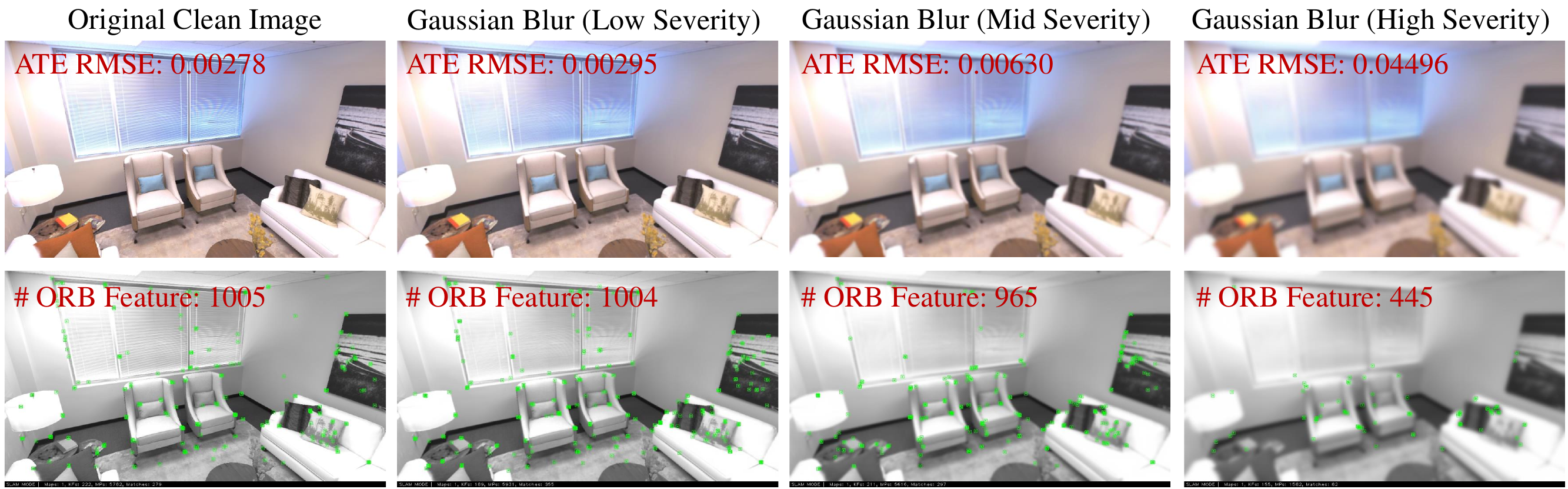} 
	\caption{\textbf{Effect of \textit{Gaussian Blur} image-level perturbation under different severity (\textbf{Top}) on the quality of detected ORB features} (\textbf{Bottom}), which are marked as green dots, for the classical SLAM model ORBSLAM3~\cite{orbslam3}. We report the average trajectory accuracy via ATE RMSE and the average number of ORB features detected in various perturbed settings. }
	\label{fig:orbslam-gaussian-blur} 
\end{figure*}
\begin{figure*}[t]
	\centering
\includegraphics[width=0.48\textwidth]{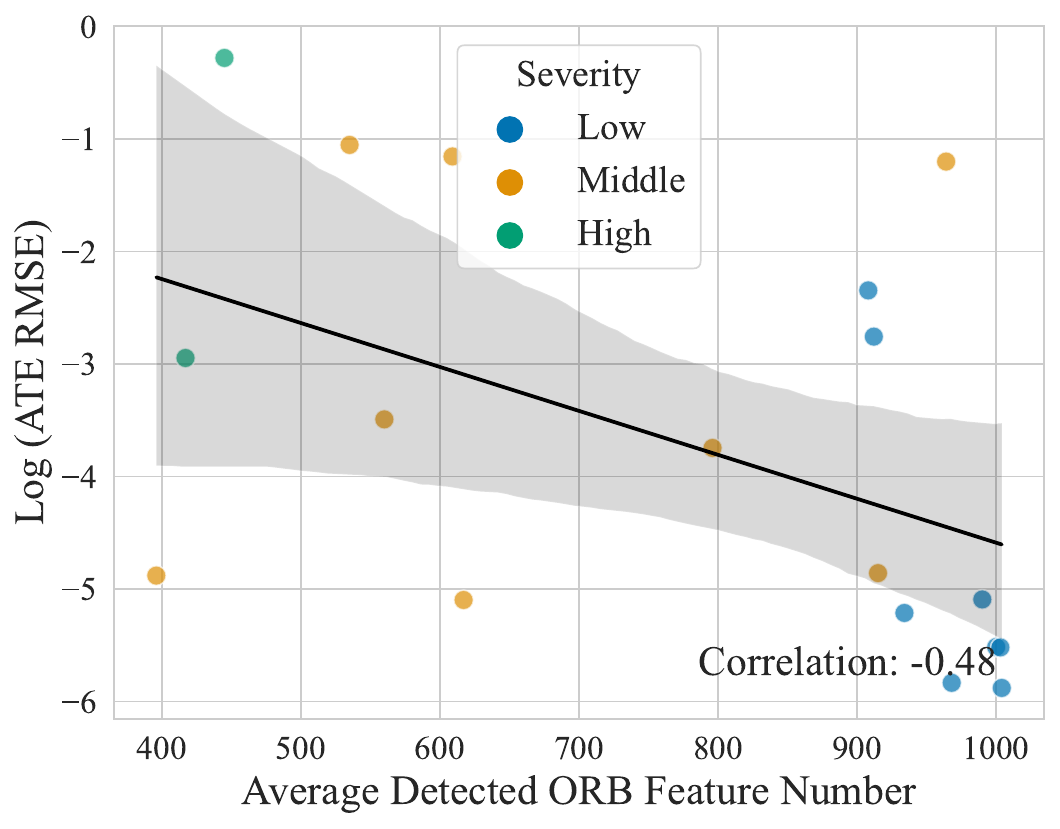} 
\includegraphics[width=0.48\textwidth]{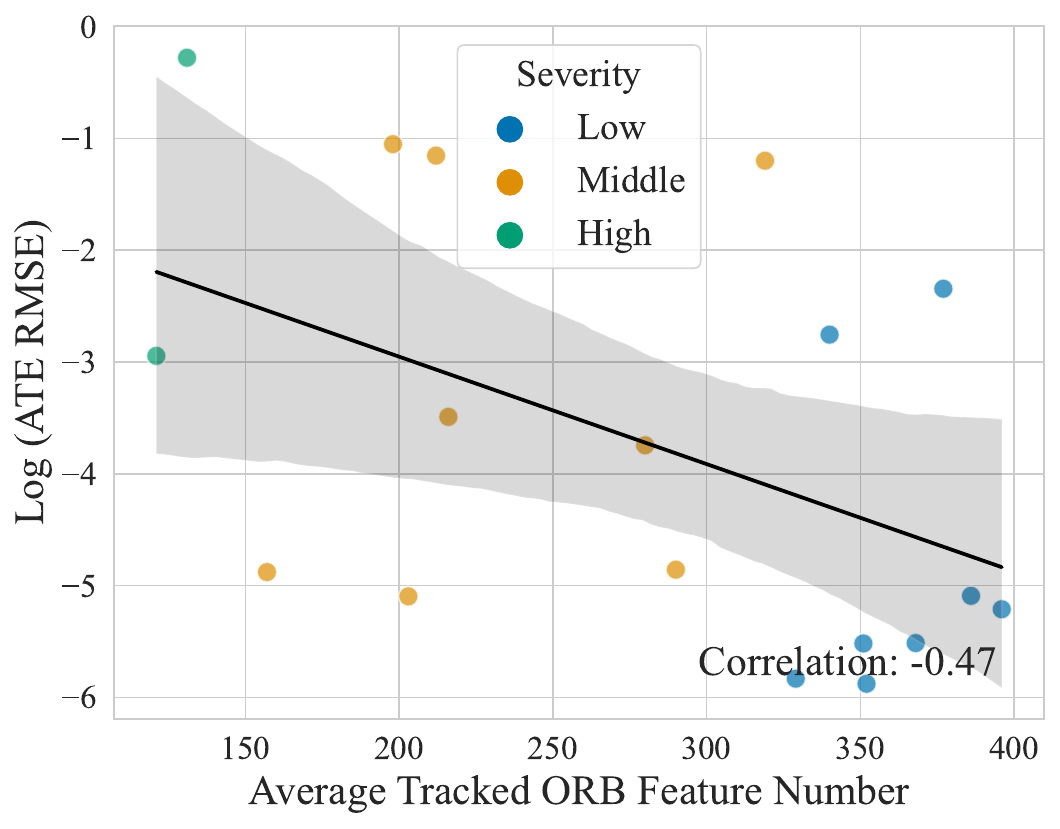} 
	\caption{\textbf{Correlation between trajectory estimation accuracy} \textbf{and the average number of detected} (\textbf{Left}) \textbf{and tracked} (\textbf{Right}) \textbf{ORB features for ORBSLAM3}~\cite{orbslam3} model (RGBD setting) under different severity level of Gaussian Blur image-level perturbation. We report the Pearson correlation coefficient~\cite{cohen2009pearson} at the bottom right corner. While the correlation coefficient does not indicate a significant linear correlation, there is a noticeable trend of increased trajectory estimation when fewer ORB features are detected or tracked.}
	\label{fig:correlation-orb-accuracy}
\end{figure*}
\subsection{More Discussions}

\textit{\textbf{Can a classical SLAM model be aware of degraded observations?}}
In the main paper, we showcase that, for the neural SLAM model SplaTAM-S~\cite{keetha2023splatam}, there is an obvious correlation between the RGBD reconstruction quality and the final trajectory estimation accuracy under the faster motion perturbation.
Here, we explore the ability of a classical SLAM model, \textit{i.e.}, ORB-SLAM3~\cite{orbslam3}, to `perceive' perturbation severity. Specifically, we aim to explore the correlation between the quality of ORB feature detection and the resulting overall performance.
In Fig.~\ref{fig:orbslam-gaussian-blur}, we present qualitative comparisons of the ORB feature detection results of ORB-SLAM3 under the influence of varying severity levels of Gaussian blur image-level perturbations. It is evident that more severe perturbations result in a lower number of detected ORB features. In addition, Fig.\ref{fig:correlation-orb-accuracy} depicts the correlation between the number of (detected or matched) ORB feature descriptors and the accuracy of trajectory estimation. With increasing severity levels, a noticeable reduction in the number of ORB features is observed, accompanied by a subsequent rise in trajectory estimation error, \textit{i.e.}, ATE. This trend indicates that the deterioration of detected feature descriptors could serve as an informative indicator for identifying degraded and anomalous observations. Furthermore, it has the potential to provide a rough estimation of the overall performance of ORB-SLAM3 in situations where ground-truth annotation is unavailable.

\begin{figure*}[ht!]
	\centering
\includegraphics[width=\textwidth]{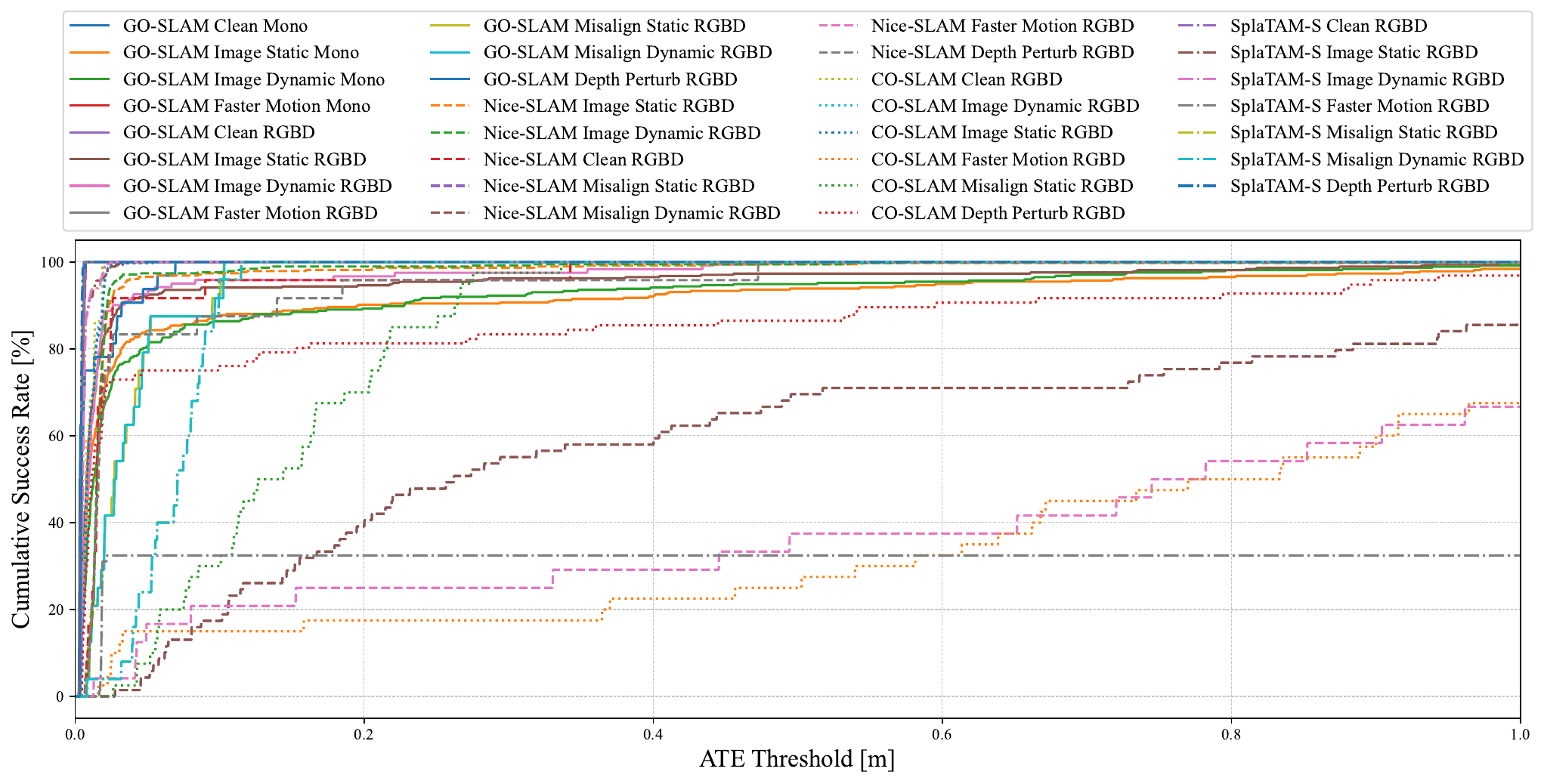} 
\includegraphics[width=\textwidth]{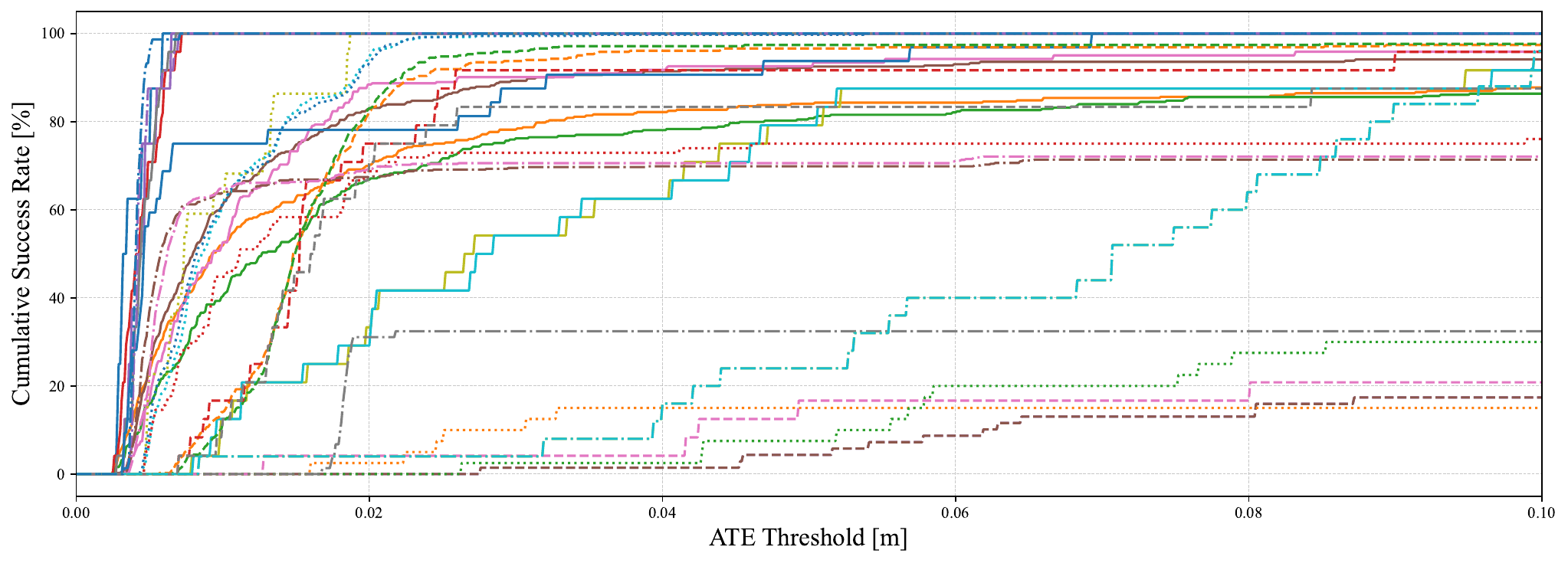} 
\includegraphics[width=\textwidth]{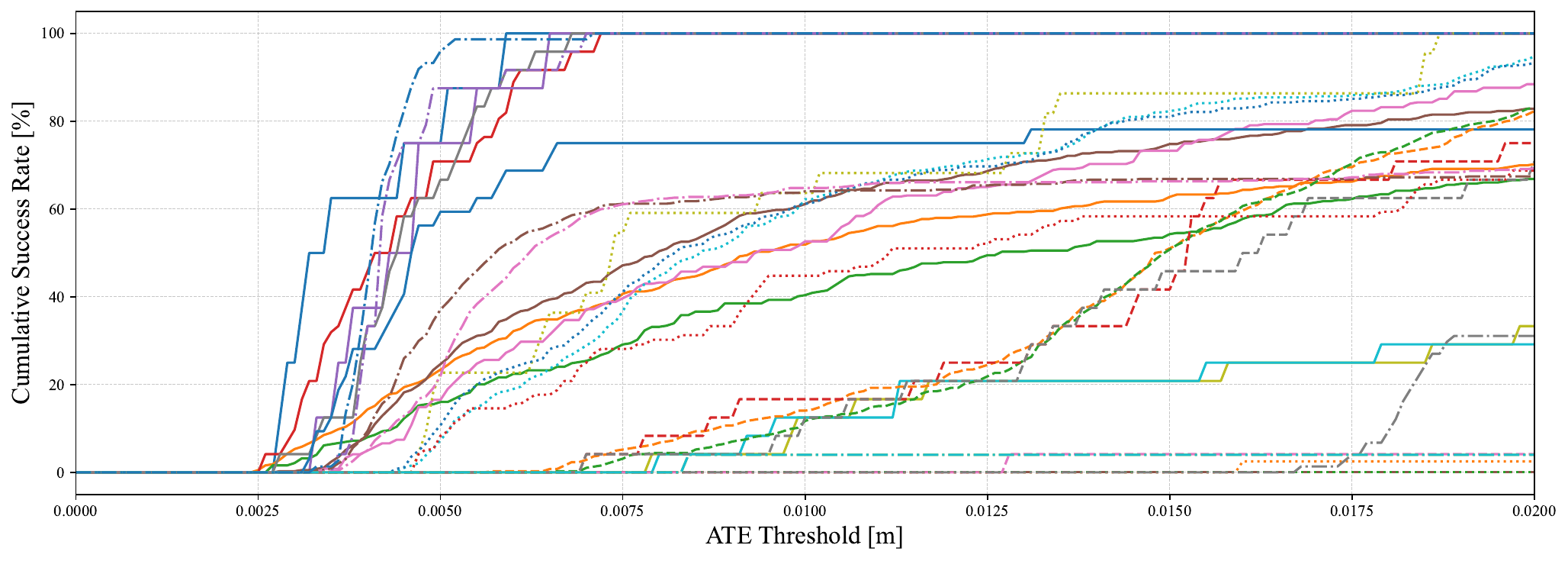} 
\caption{\textbf{Cumulative success rate measured at various ATE threshold ranges for multiple types of perturbations.} We include the results within the range of $\leq 1.0$ [m] (\textbf{Top}), $\leq 0.1$ [m] (\textbf{Middle}), and $\leq 0.02$ [m] (\textbf{Bottom}). We present the results of five SLAM models, including monocular\&RGBD-based GO-SLAM~\cite{zhang2023goslam}, RGBD-based Nice-SLAM~\cite{niceslam}, RGBD-based CO-SLAM~\cite{coslam}, and RGBD-based SplaTAM-S~\cite{keetha2023splatam}, which highlights the vulnerability of current models to diverse types of perturbations.}
\label{fig:cumulative_density_comparison} 
\end{figure*}

\begin{figure*}[ht!]
	\centering
 \includegraphics[width=\textwidth]{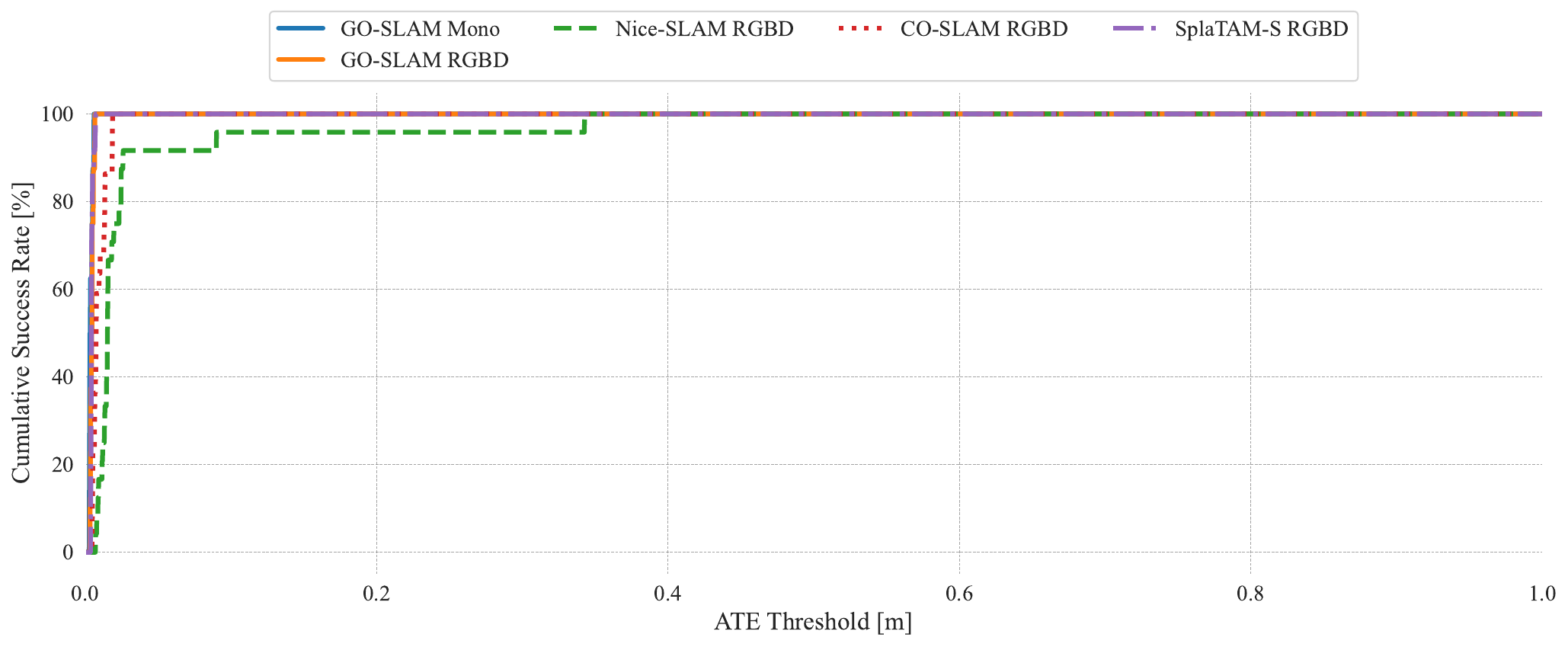}
 \vspace{4mm}
\includegraphics[width=\textwidth]{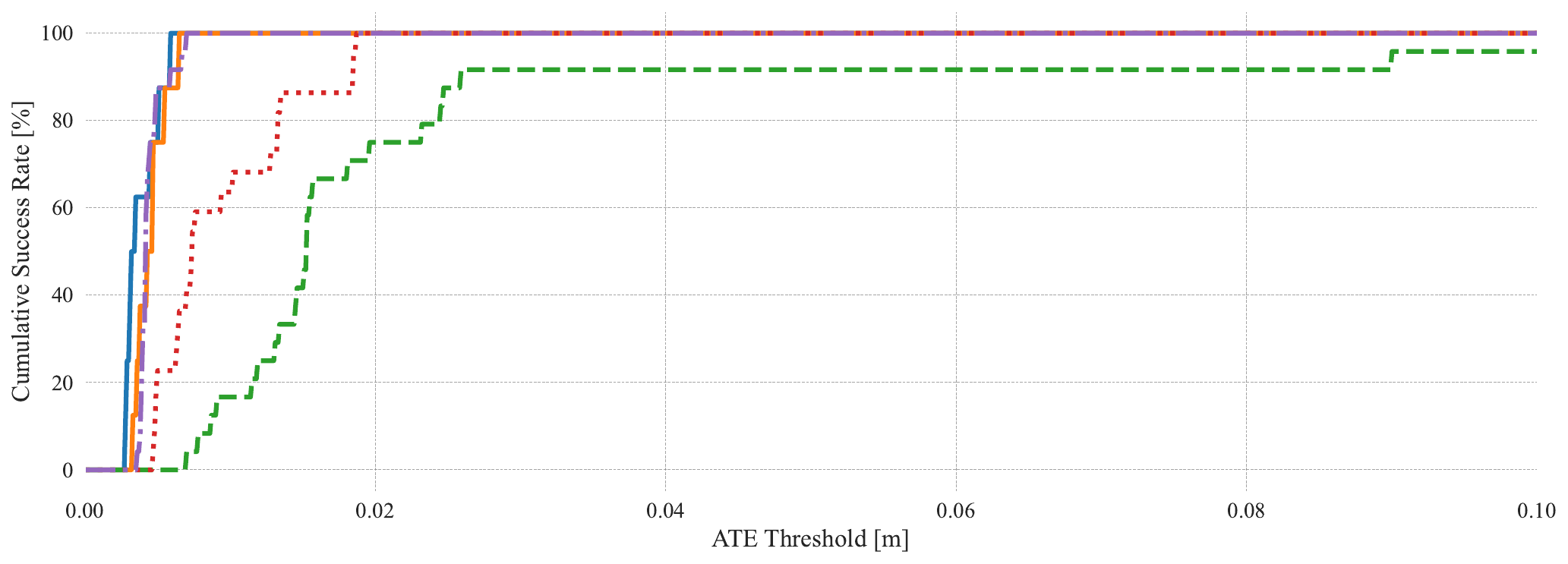}  \vspace{4mm}
\includegraphics[width=\textwidth]{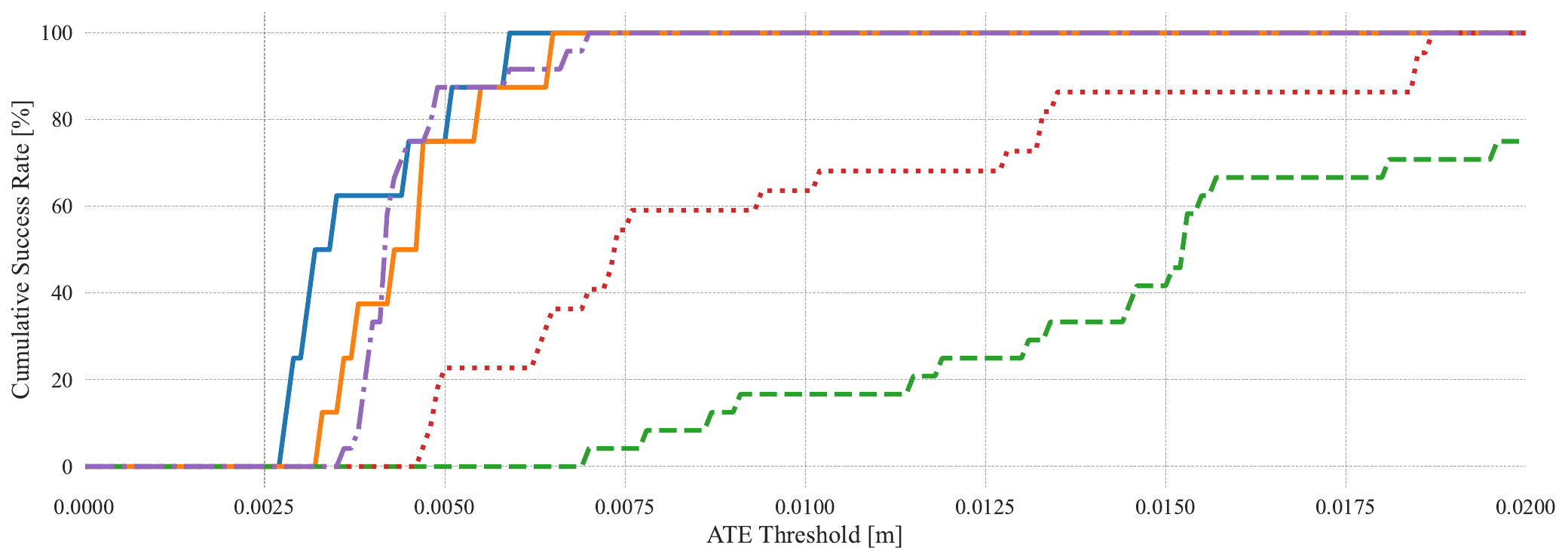} 
\caption{\textbf{Cumulative success rate measured at various ATE threshold ranges under the original clean setting.} We include the results within the ATE range of $\leq 1.0$ [m] (\textbf{Top}), $\leq 0.1$ [m] (\textbf{Middle}), and $\leq 0.02$ [m] (\textbf{Bottom}). We present the results of five SLAM models, including monocular\&RGBD-based GO-SLAM~\cite{zhang2023goslam}, RGBD-based Nice-SLAM~\cite{niceslam}, RGBD-based CO-SLAM~\cite{coslam}, and RGBD-based SplaTAM-S~\cite{keetha2023splatam}.}
\label{fig:cumulative_density_comparison_clean} 
\end{figure*}

\begin{figure*}[ht!]
	\centering
 \includegraphics[width=\textwidth]{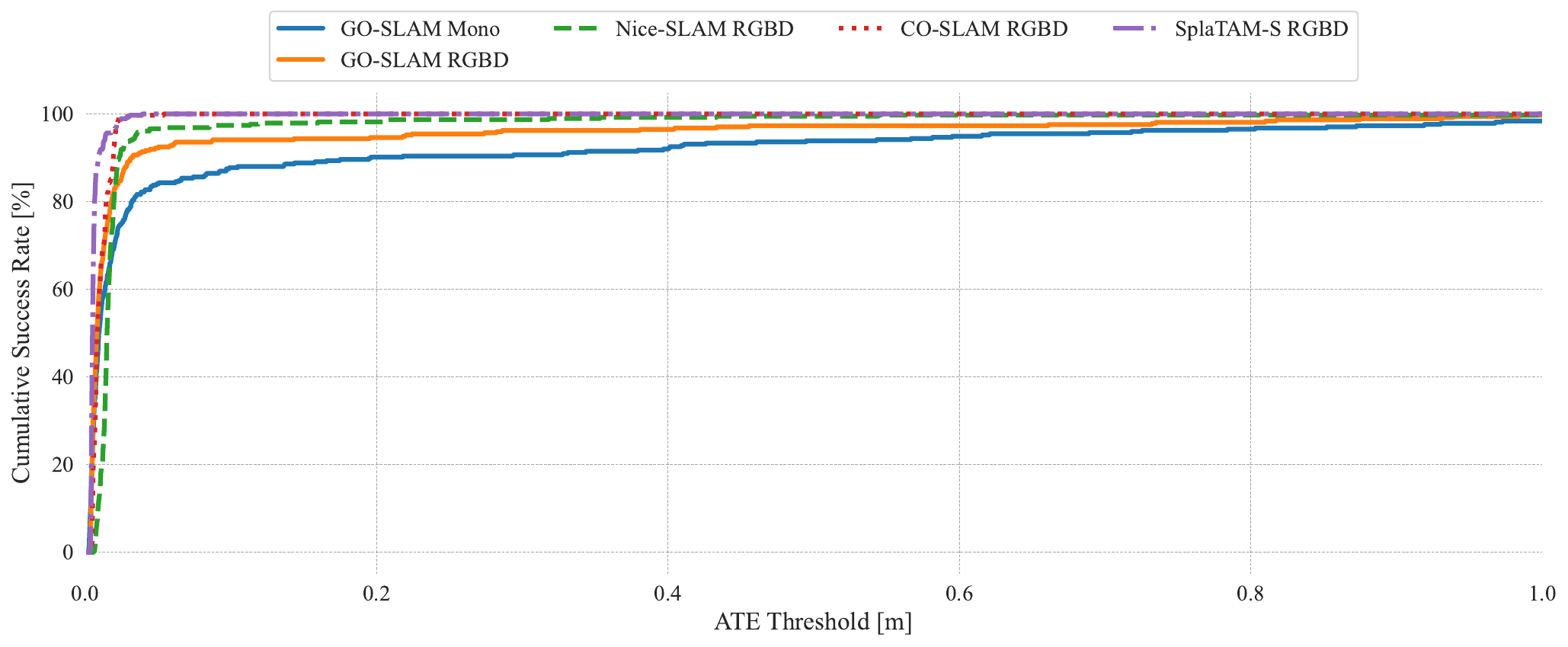}
 \vspace{4mm}
\includegraphics[width=\textwidth]{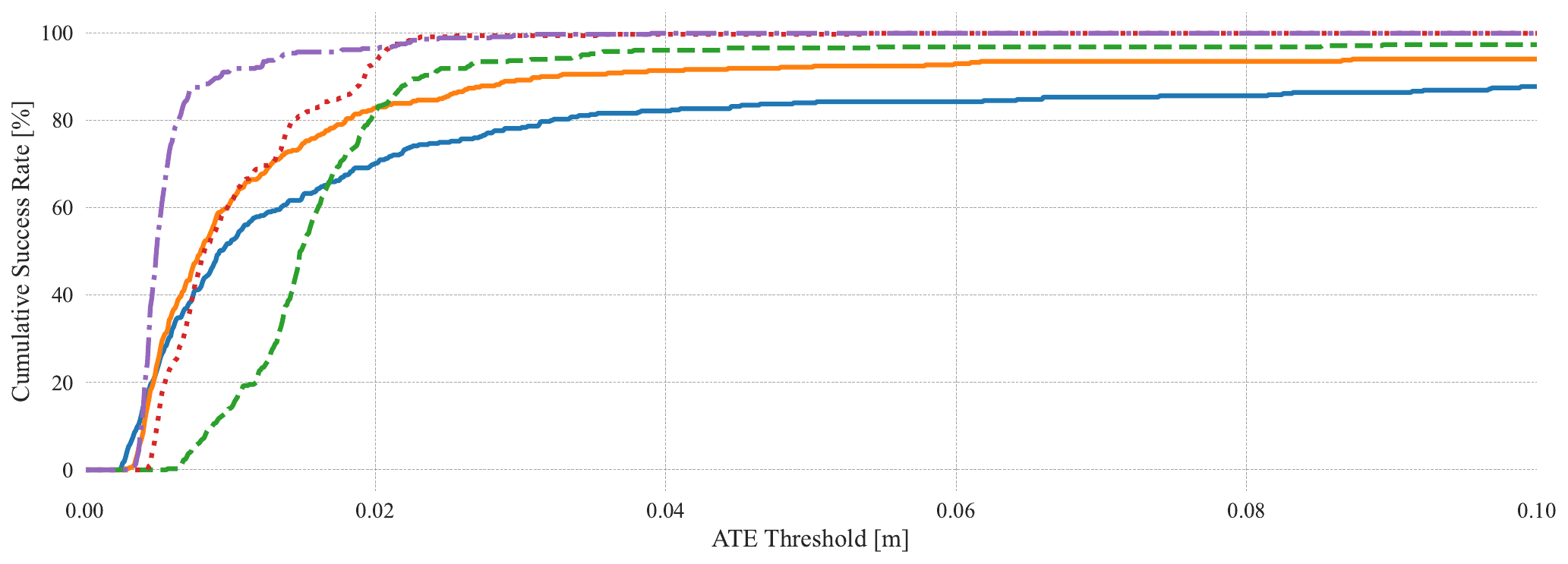} 
\vspace{4mm}
\includegraphics[width=\textwidth]{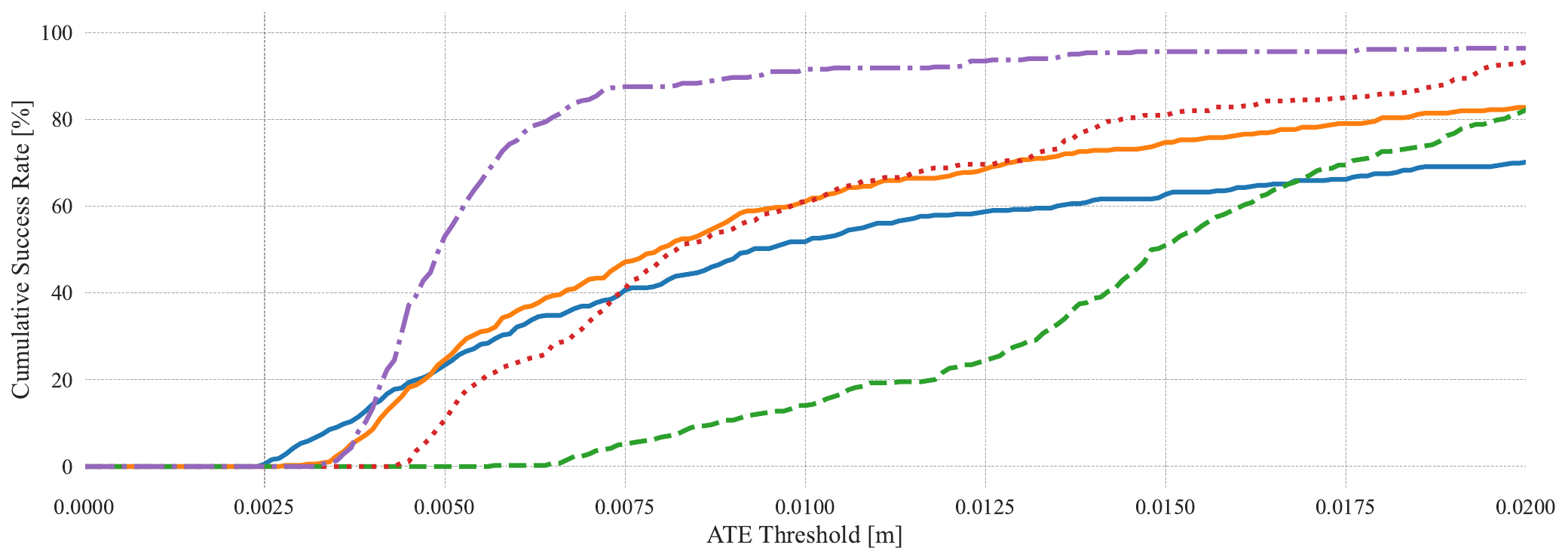} 
\caption{\textbf{Cumulative success rate measured at various ATE threshold ranges under the static mode of image-level perturbations.} We include the results within the ATE range of $\leq 1.0$ [m] (\textbf{Top}), $\leq 0.1$ [m] (\textbf{Middle}), and $\leq 0.02$ [m] (\textbf{Bottom}). We present the results of five SLAM models, including monocular\&RGBD-based GO-SLAM~\cite{zhang2023goslam}, RGBD-based Nice-SLAM~\cite{niceslam}, RGBD-based CO-SLAM~\cite{coslam}, and RGBD-based SplaTAM-S~\cite{keetha2023splatam}.}
\label{fig:cumulative_density_comparison_image_static} 
\end{figure*}

\begin{figure*}[ht!]
	\centering
 \includegraphics[width=\textwidth]{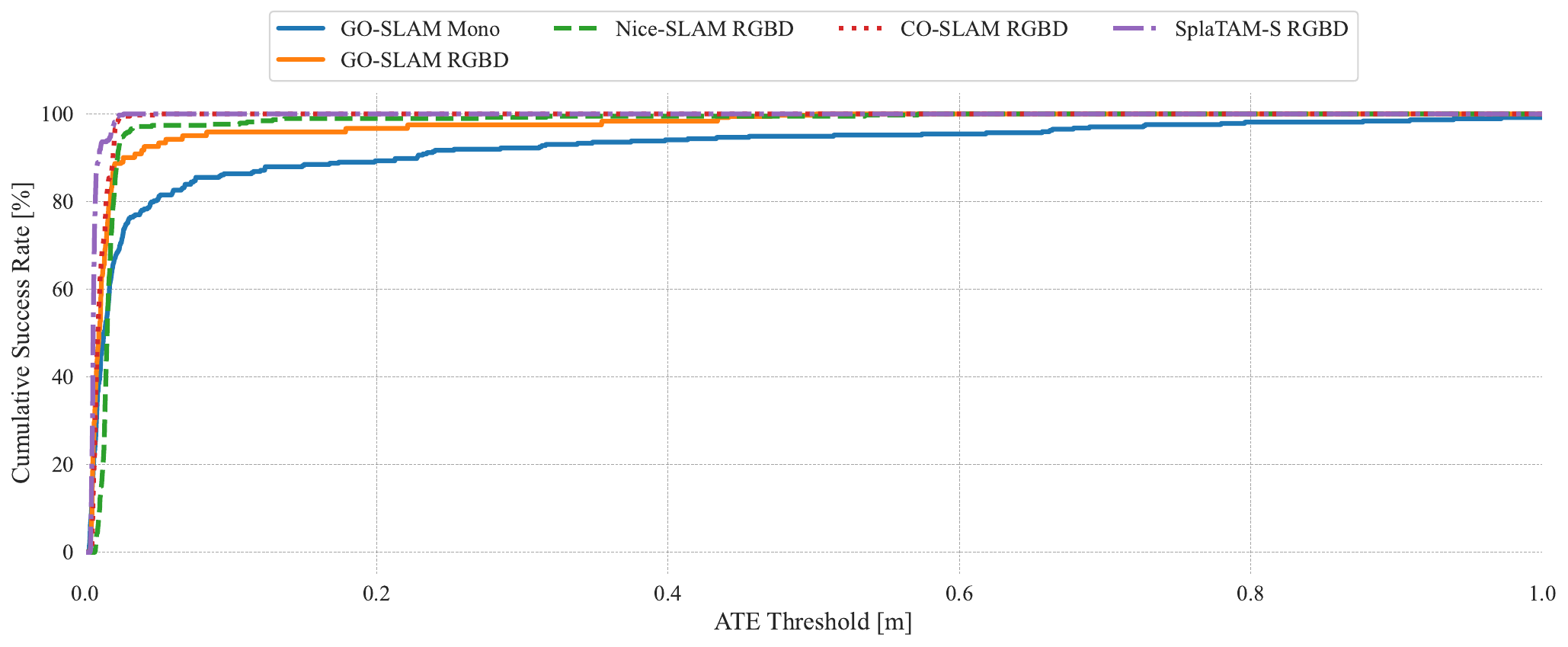}
 \vspace{4mm}
\includegraphics[width=\textwidth]{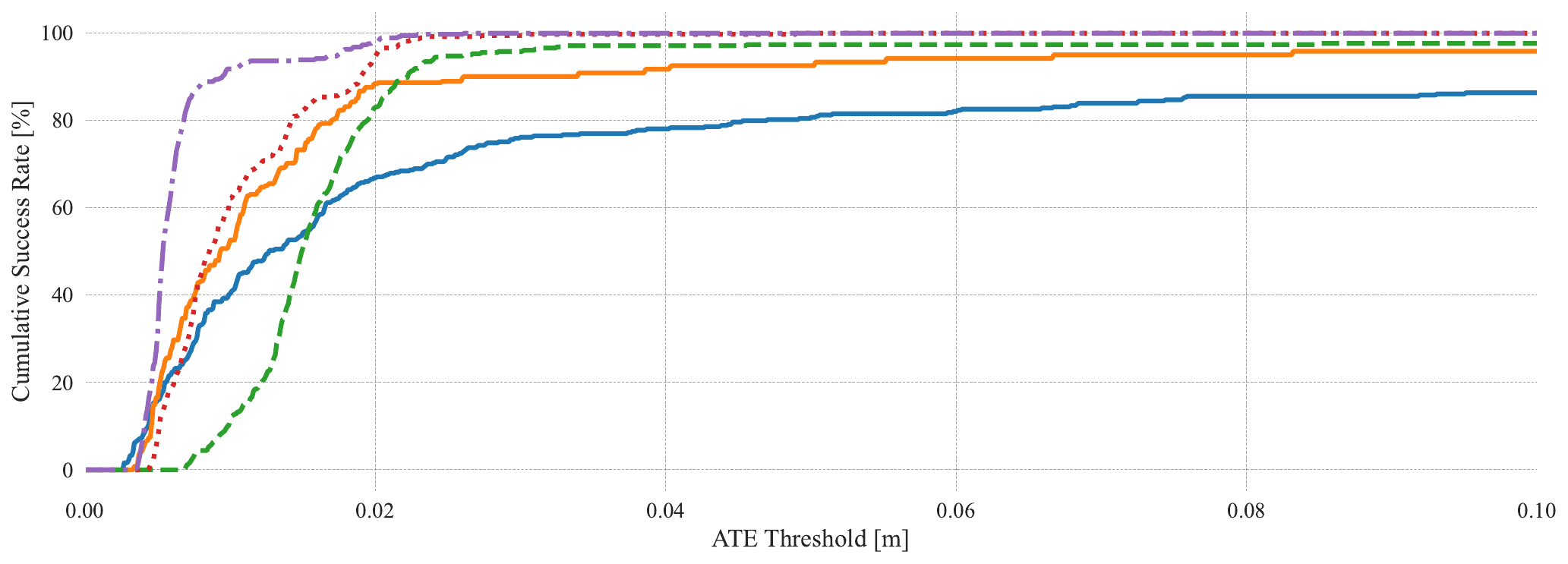} 
\vspace{4mm}
\includegraphics[width=\textwidth]{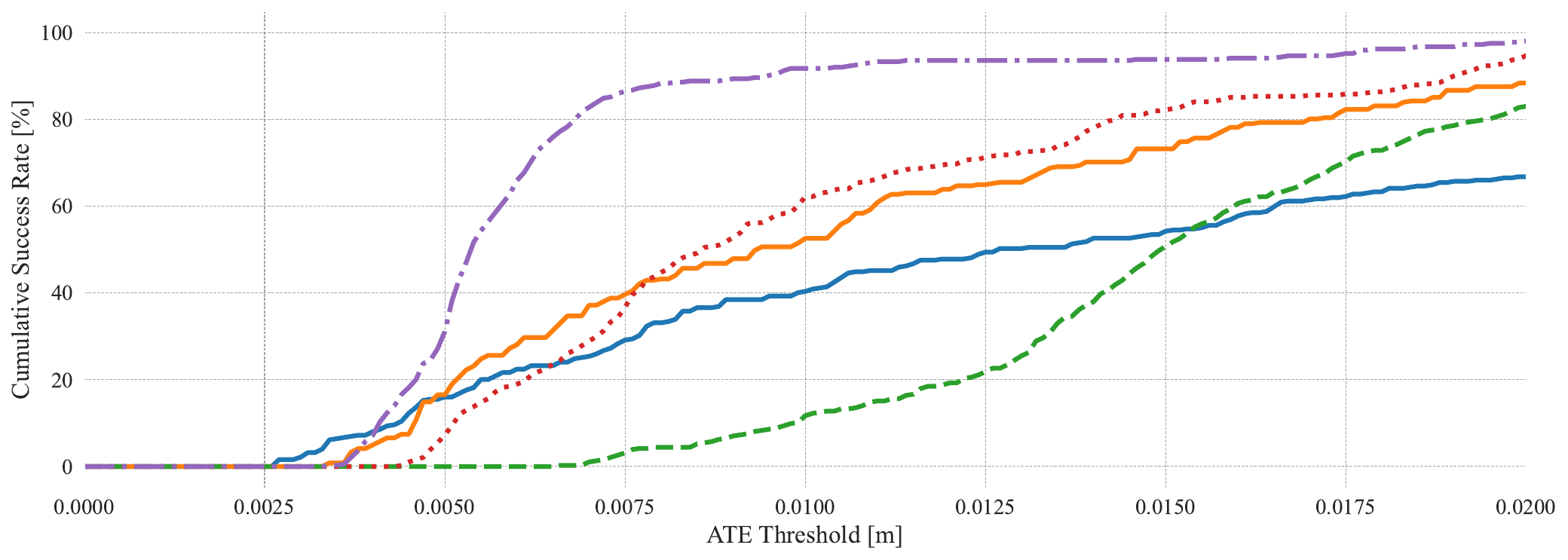} 
\caption{\textbf{Cumulative success rate measured at various ATE threshold ranges under the dynamic mode of image-level perturbations.} We include the results within the ATE range of $\leq 1.0$ [m] (\textbf{Top}), $\leq 0.1$ [m] (\textbf{Middle}), and $\leq 0.02$ [m] (\textbf{Bottom}).  We present the results of five SLAM models, including monocular\&RGBD-based GO-SLAM~\cite{zhang2023goslam}, RGBD-based Nice-SLAM~\cite{niceslam}, RGBD-based CO-SLAM~\cite{coslam}, and RGBD-based SplaTAM-S~\cite{keetha2023splatam}. }
\label{fig:cumulative_density_comparison_dynamic} 
\end{figure*}

\begin{figure*}[ht!]
	\centering
 \includegraphics[width=\textwidth]{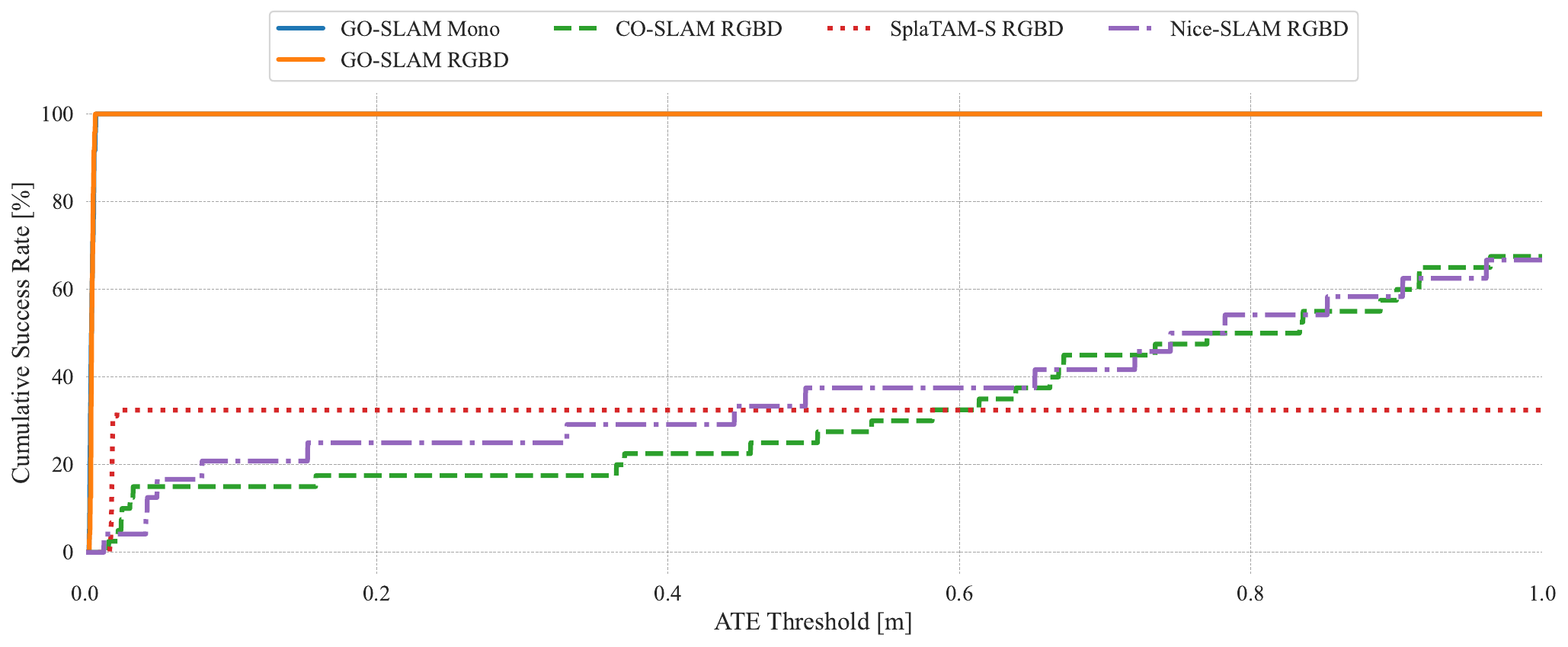}
 \vspace{4mm}
\includegraphics[width=\textwidth]{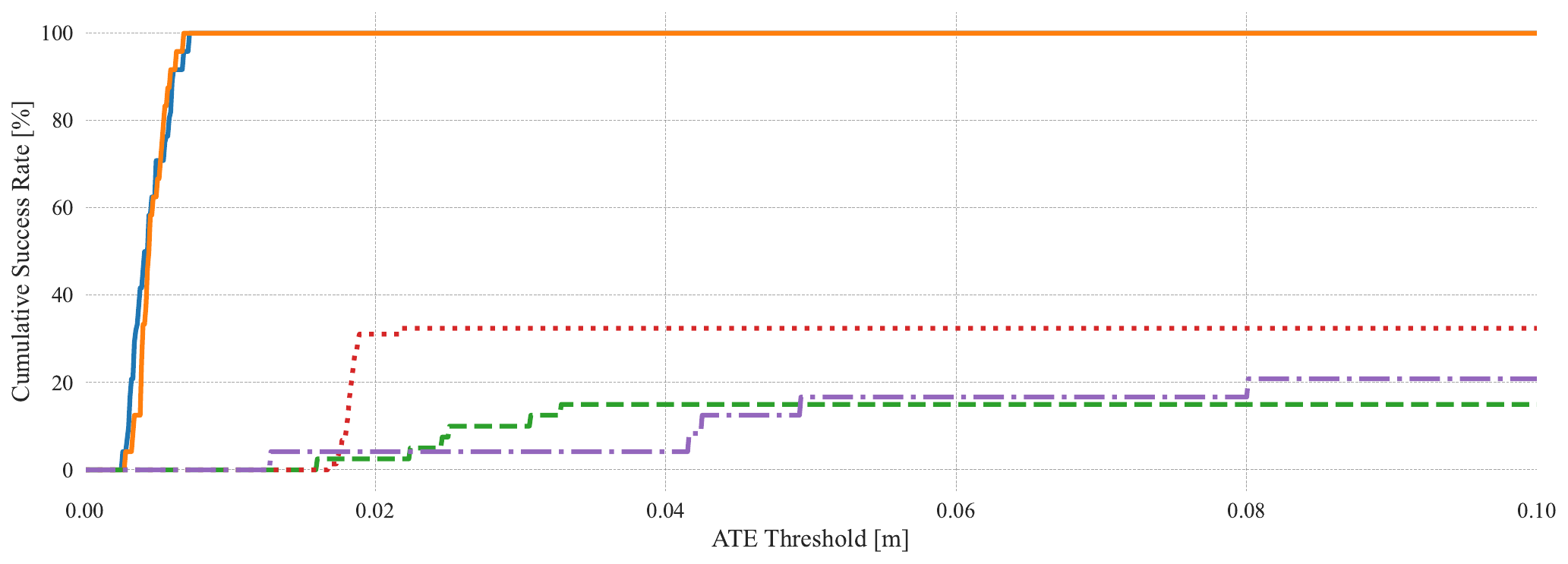} 
\vspace{4mm}
\includegraphics[width=\textwidth]{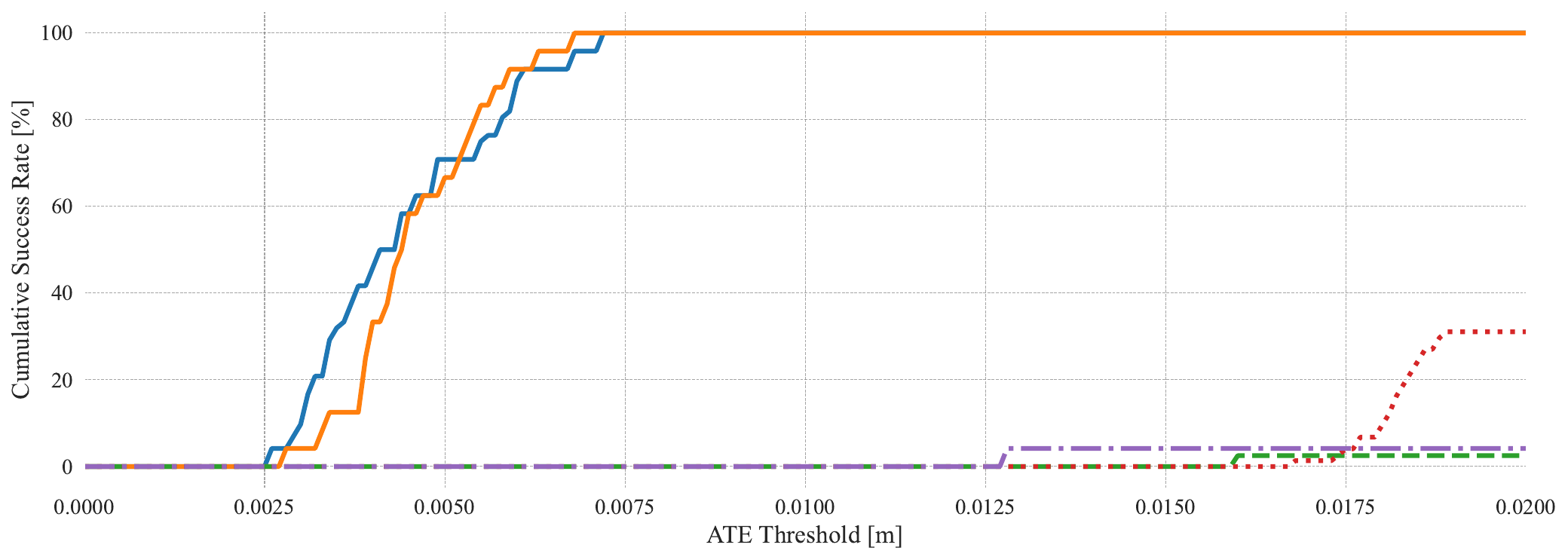} 
\caption{\textbf{Cumulative success rate measured at various ATE threshold ranges under the faster motion effect.} We include the results within the ATE range of $\leq 1.0$ [m] (\textbf{Top}), $\leq 0.1$ [m] (\textbf{Middle}), and $\leq 0.02$ [m] (\textbf{Bottom}).  We present the results of five SLAM models, including monocular\&RGBD-based GO-SLAM~\cite{zhang2023goslam}, RGBD-based Nice-SLAM~\cite{niceslam}, RGBD-based CO-SLAM~\cite{coslam}, and RGBD-based SplaTAM-S~\cite{keetha2023splatam}.}
\label{fig:cumulative_density_comparison_faster_motion} 
\end{figure*}

\begin{figure*}[ht!]
	\centering
 \includegraphics[width=\textwidth]{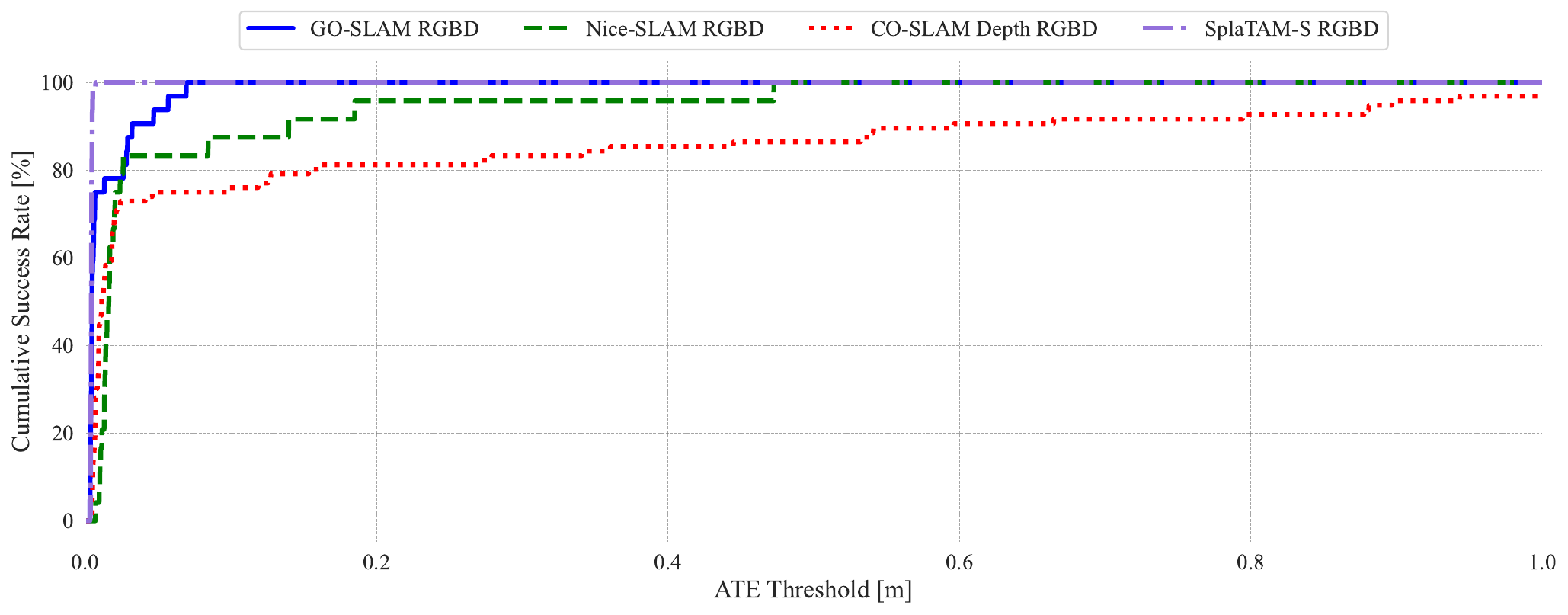}
 \vspace{4mm}
\includegraphics[width=\textwidth]{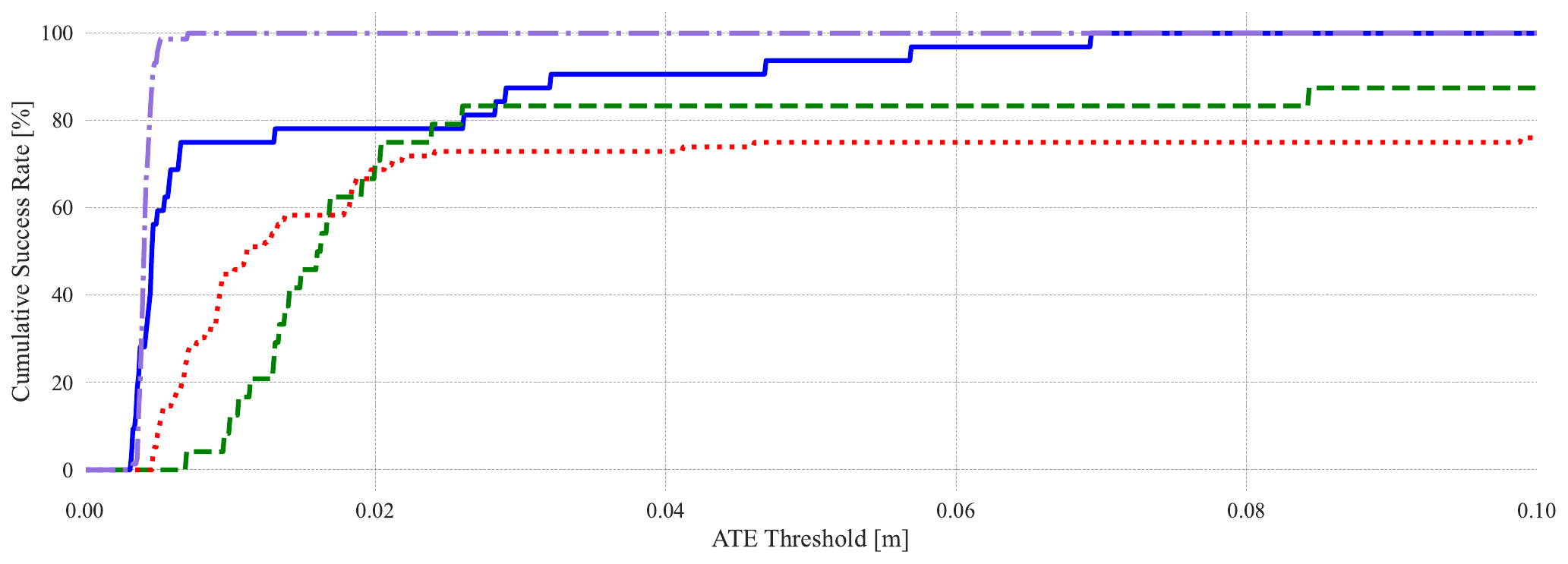} 
\vspace{4mm}
\includegraphics[width=\textwidth]{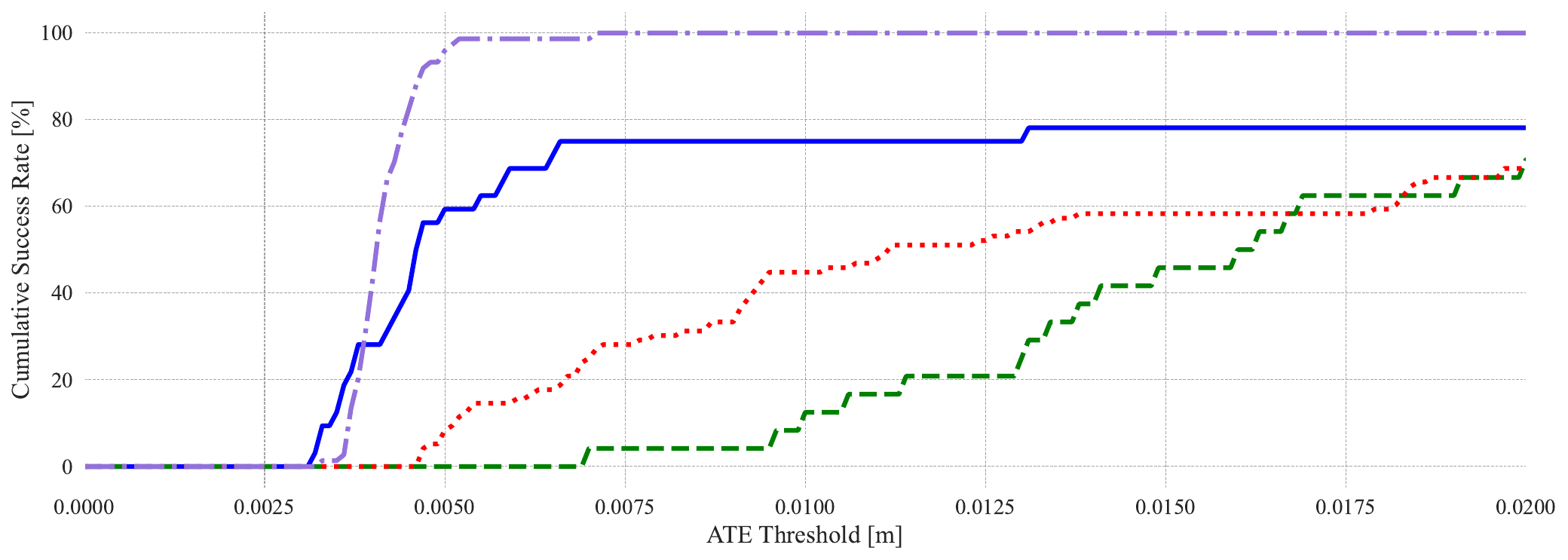} 
\caption{\textbf{Cumulative success rate measured at various ATE threshold ranges under the depth perturbation.} We include the results within the ATE range of $\leq 1.0$ [m] (\textbf{Top}), $\leq 0.1$ [m] (\textbf{Middle}), and $\leq 0.02$ [m] (\textbf{Bottom}).  We present the results of four RGBD SLAM models, including GO-SLAM~\cite{zhang2023goslam}, Nice-SLAM~\cite{niceslam}, CO-SLAM~\cite{coslam}, and SplaTAM-S~\cite{keetha2023splatam}. }
\label{fig:cumulative_density_comparison_depth} 
\end{figure*}

\begin{figure*}[ht!]
	\centering
 \includegraphics[width=\textwidth]{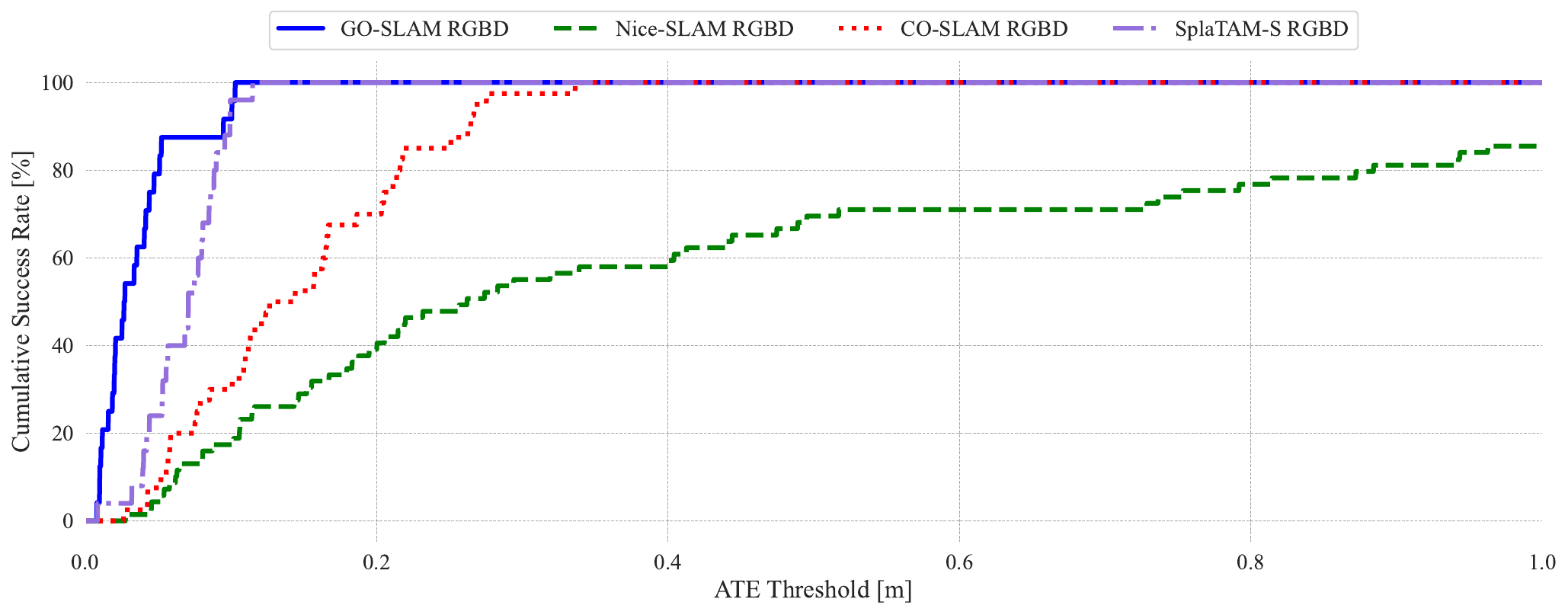}
 \vspace{4mm}
\includegraphics[width=\textwidth]{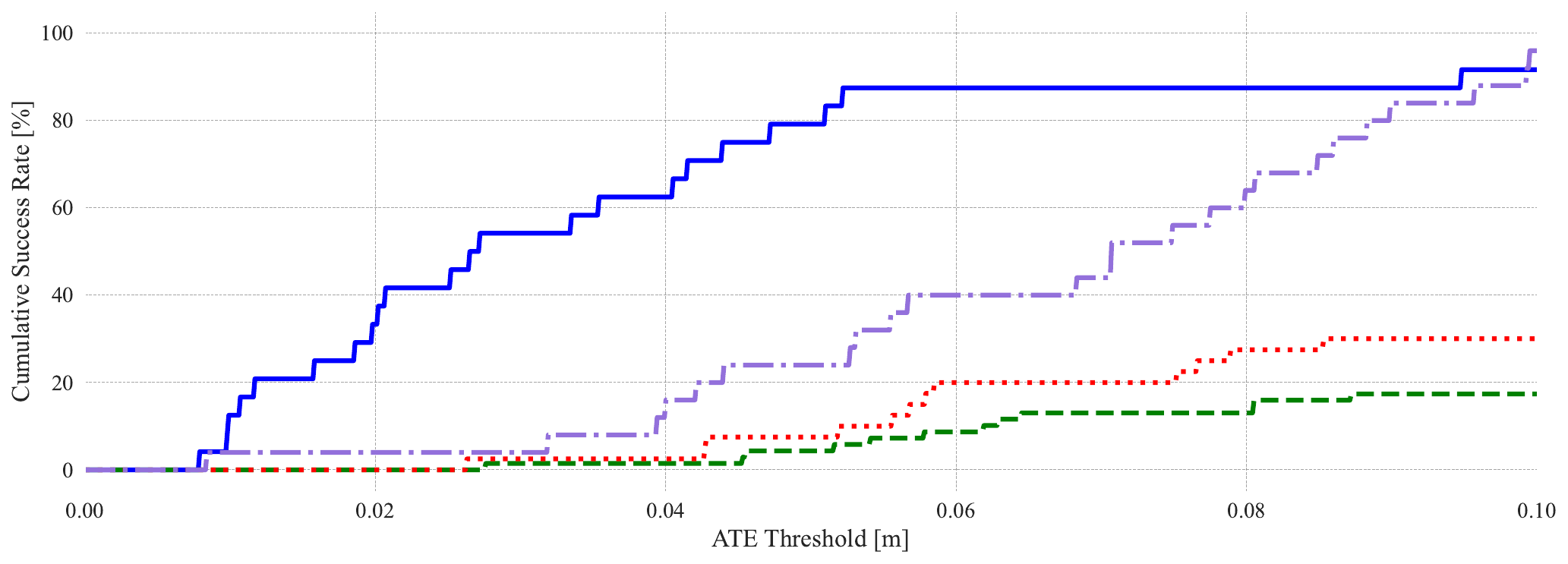} 
\vspace{4mm}
\includegraphics[width=\textwidth]{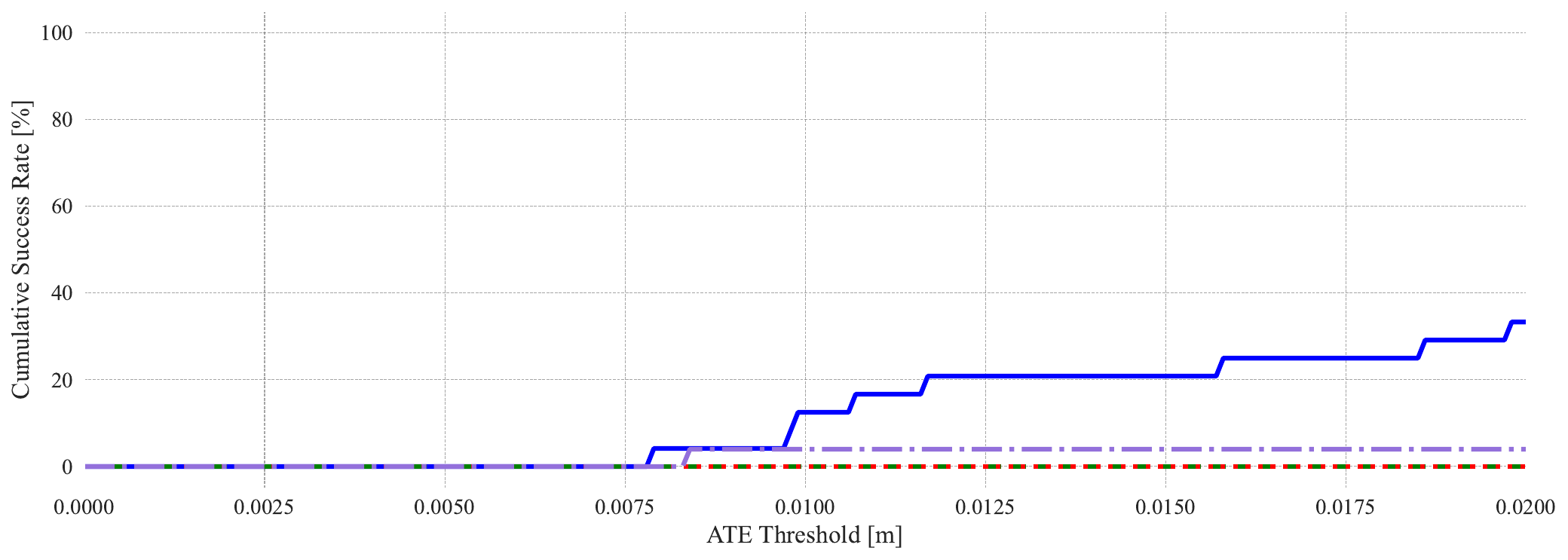} 
\caption{\textbf{Cumulative success rate measured at various ATE threshold ranges under the static mode of the sensor misalignment perturbation.} We include the results within the ATE range of $\leq 1.0$ [m] (\textbf{Top}), $\leq 0.1$ [m] (\textbf{Middle}), and $\leq 0.02$ [m] (\textbf{Bottom}).  We present the results of four RGBD-based SLAM models, including GO-SLAM~\cite{zhang2023goslam}, Nice-SLAM~\cite{niceslam}, CO-SLAM~\cite{coslam}, and SplaTAM-S~\cite{keetha2023splatam}.}
\label{fig:cumulative_density_comparison_sensor_misalign_static} 
\end{figure*}

\begin{figure*}[ht!]
	\centering
 \includegraphics[width=\textwidth]{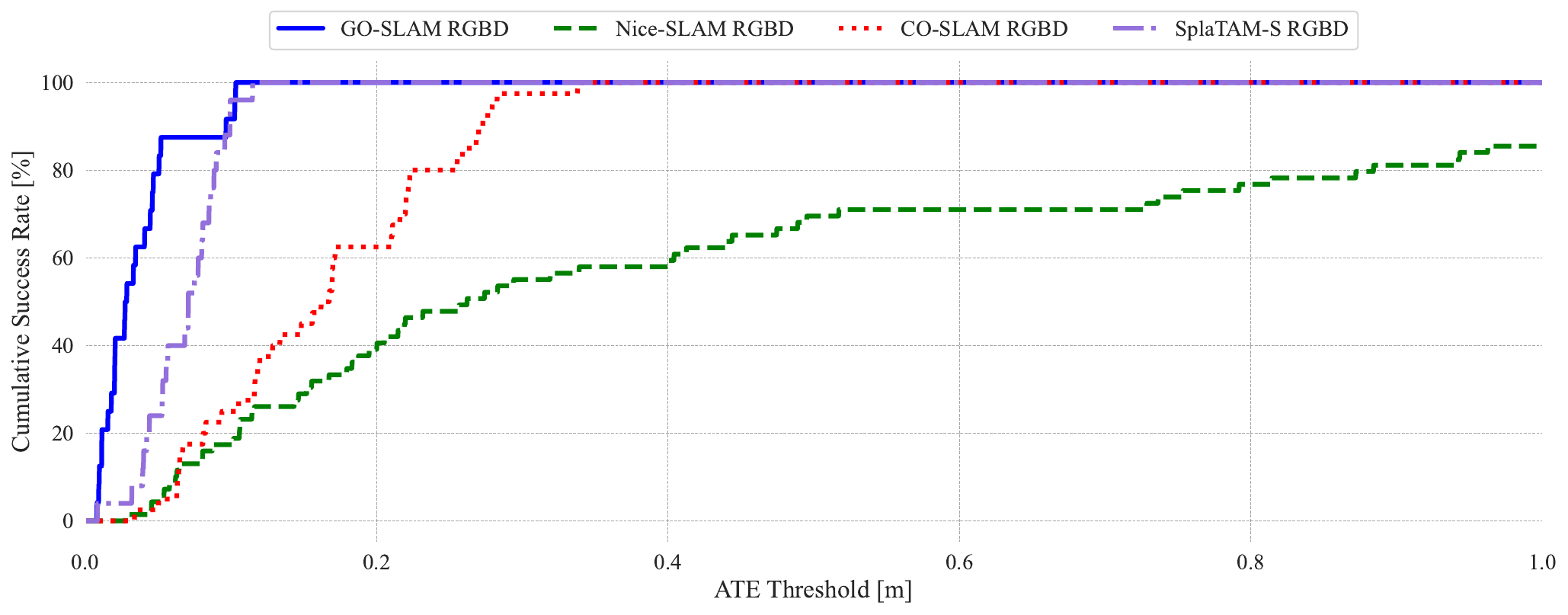}
 \vspace{4mm}
\includegraphics[width=\textwidth]{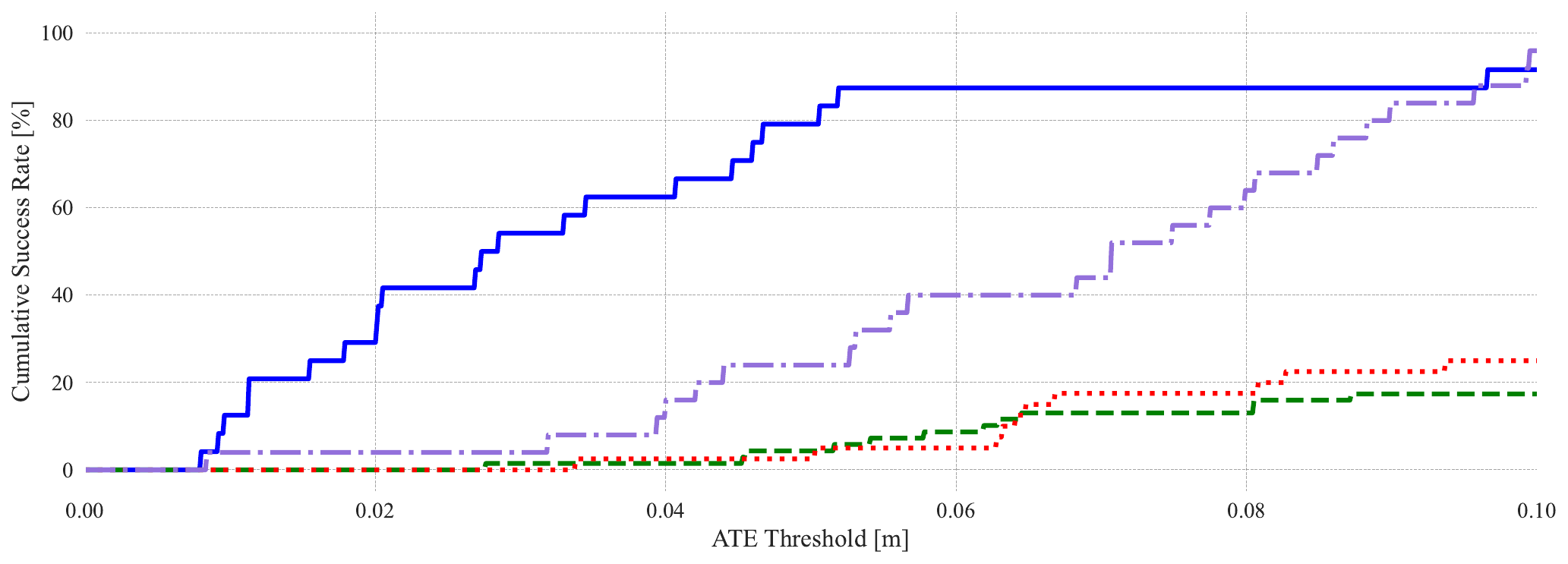} 
\vspace{4mm}
\includegraphics[width=\textwidth]{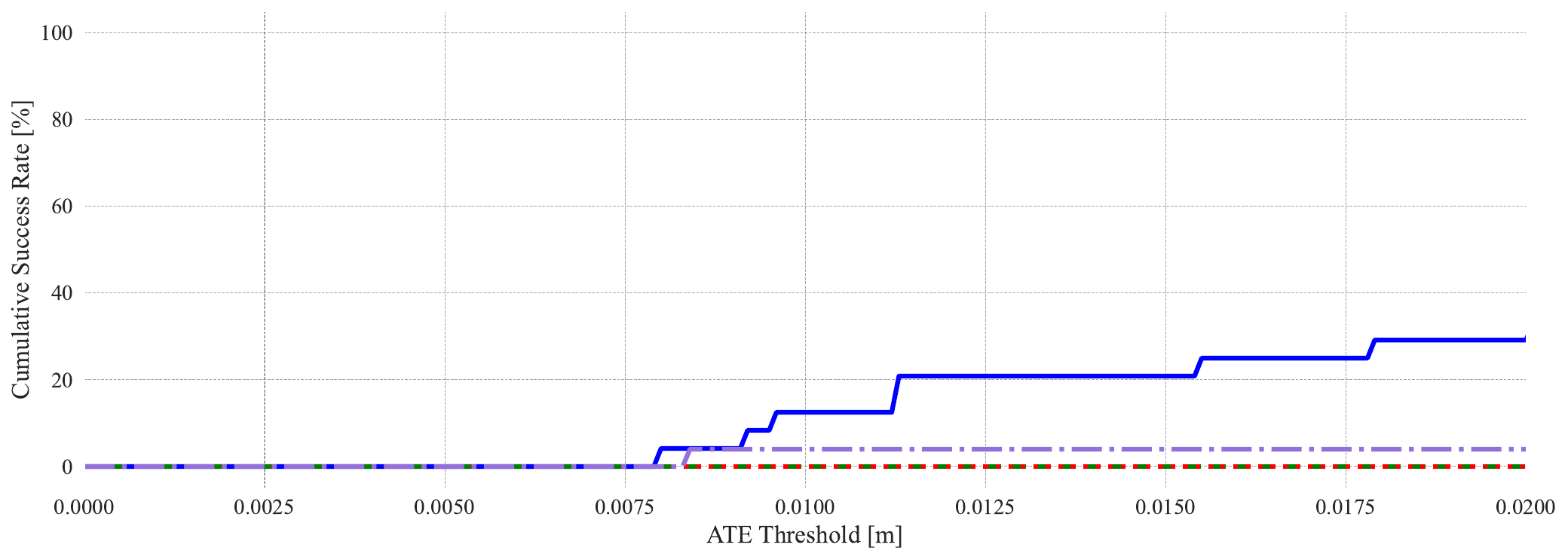} 
\caption{\textbf{Cumulative success rate measured at various ATE threshold ranges under the dynamic mode of the sensor misalignment perturbation.} We include the results within the ATE range of $\leq 1.0$ [m] (\textbf{Top}), $\leq 0.1$ [m] (\textbf{Middle}), and $\leq 0.02$ [m] (\textbf{Bottom}).  We present the results of four RGBD-based SLAM models, including GO-SLAM~\cite{zhang2023goslam}, Nice-SLAM~\cite{niceslam}, CO-SLAM~\cite{coslam}, and SplaTAM-S~\cite{keetha2023splatam}.}
\label{fig:cumulative_density_comparison_sensor_misalign_dynamic} 
\end{figure*}

\noindent\textbf{Are current SLAM models specialized or generalist?}
Fig.~\ref{fig:cumulative_density_comparison} depicts the cumulative success rate (CSR) curves of some advanced SLAM models, providing insights into their performance across various categories of perturbations. For the qualitative comparison within each perturbation type, we present separate cumulative success rate results of different SLAM methods in Fig.\ref{fig:cumulative_density_comparison_image_static}, Fig.\ref{fig:cumulative_density_comparison_dynamic}, Fig.\ref{fig:cumulative_density_comparison_faster_motion}, Fig.\ref{fig:cumulative_density_comparison_depth}, Fig.\ref{fig:cumulative_density_comparison_sensor_misalign_static}, and Fig.\ref{fig:cumulative_density_comparison_sensor_misalign_dynamic} (with the clean setting in Fig.~\ref{fig:cumulative_density_comparison_clean} as a reference). The results highlight the vulnerability of current models to `survive' under varied perturbation types. It reveals that different methods exhibit specialization in achieving robustness under specific settings. For instance, the SplaTAM-S~\cite{keetha2023splatam} model excels under image-level perturbations, while the GO-SLAM~\cite{zhang2023goslam} model demonstrates excellent robustness under trajectory-related perturbations. However, the development of a robust \textit{generalist} SLAM model that exhibits robustness against both sensor and trajectory perturbations remains relatively under-explored.

\begin{table*}[t]
\caption{\textbf{Effects of perturbing the stereo camera baseline length on trajectory estimation performance of the ORBSLAM3 model with  stereo (\textbf{Top}) and stereo-inertia (\textbf{Bottom}) SLAM setting}. We leverage three sequences, including \textit{MH01}, \textit{MH02}, and \textit{MH03}, on the EuRoC~\cite{Burri25012016} dataset. We introduce random noise $\eta$ along the axis of the stereo camera baseline. The noise is sampled from a Gaussian distribution with zero mean and variance $\sigma^2$, represented as $\eta \sim \mathcal{N}(0, \sigma^2)$ [m]. The stereo camera has a predefined baseline length of 0.1 meters.}
\centering 
{
\begin{tabular}{l|ccc|ccc|ccc}
    \toprule \toprule   
        \multirow{1}{*}{\textbf{ }}  &    \multicolumn{3}{c|}{\textit{MH01} Sequence}  &     \multicolumn{3}{c|}{\textit{MH02} Sequence}  &    \multicolumn{3}{c}{\textit{MH03} Sequence}\\ \midrule
    \multirow{1}{*}{\textbf{Metrics}}  &     {$\sigma=0.00$} &     {$\sigma=0.001$}
     &  {$\sigma=0.01$} &     {$\sigma=0.00$} &     {$\sigma=0.001$}
     &  {$\sigma=0.01$} &     {$\sigma=0.00$} &     {$\sigma=0.001$}
     &  {$\sigma=0.01$}  \\ \midrule
 \multicolumn{10}{c}{Stereo Setting}  \\ \midrule
    \textit{ATE-w/o Scale}$\downarrow$ [m]  & $\textbf{0.036}$ & $1.646$ & $8.284$ & $\textbf{0.016}$ & $2.943$ & $19.42$  & $\textbf{0.028}$ & $1.444$ & $6.211$ 
    \\
    \textit{Scale}  & ${1.006}$ & $1.129$ & $0.261$  & ${0.998}$ & $2.251$ & $0.172$  & ${0.998}$ & $0.941$ & $0.395$ 
    \\
     \textit{ATE-w/ Scale}$\downarrow$  [m] & $\textbf{0.024}$ & $1.580$ & $2.221$ & $\textbf{0.013}$ & $1.674$ & $2.415$  & $\textbf{0.027}$ & $1.429$ & $1.985$ 
    \\  \midrule
     \multicolumn{10}{c}{Stereo-inertia Setting}  \\ \midrule
   \textit{ATE-w/o Scale}$\downarrow$ [m]    & $\textbf{0.068}$ & $0.353$ & $5.470$ & $\textbf{0.050}$ & $2.370$ & $3.954$ & $\textbf{0.053}$ & $0.572$ & $673.8$ 
    \\
    \textit{Scale}   & ${1.012}$ & $0.987$ & $0.410$ & ${1.004}$ & $2.172$ & $0.491$  & ${1.004}$ & $1.018$ & $0.001$ 
    \\
     \textit{ATE-w/ Scale}$\downarrow$ [m] & $\textbf{0.046}$ & $0.349$ & $1.865$  & $\textbf{0.046}$ & $0.312$ & $1.024$ & $\textbf{0.052}$ & $0.568$ & $3.576$ 
    \\
    \bottomrule \bottomrule
    \multicolumn{10}{l}{{\textbf{1})The best performance, indicated by the least ATE, for each sequence and input modality setting is highlighted in \textbf{bold}.}}\\
  \multicolumn{10}{l}{\textbf{2}) Under the clean setting ($\sigma=0.00$), we observe that the Stereo setting has lower trajectory estimation than the stereo-inertia setting, which  } \\
    \multicolumn{10}{l}{ is aligns with the results in Table V of the ORBSLAM3~\cite{orbslam3} paper. } \\
       \multicolumn{10}{l}{ {\textbf{3}}) \textit{ATE-w/o Scale} and \textit{ATE-w/ Scale} refer to calculating the ATE metric without and with scaling the estimated trajectory using the ground truth   }\\
       \multicolumn{10}{l}{ trajectory for evaluation, respectively; \textit{Scale} refers to the relative scaling factor between the estimated and ground truth trajectories.}
    \end{tabular}
}\label{tab:stereo-slam-robustness-case-study}
\end{table*}

\begin{figure*}[t]
	\centering
  \centering

  \begin{subfigure}{0.33\textwidth}
    \centering
    \includegraphics[width=\linewidth]{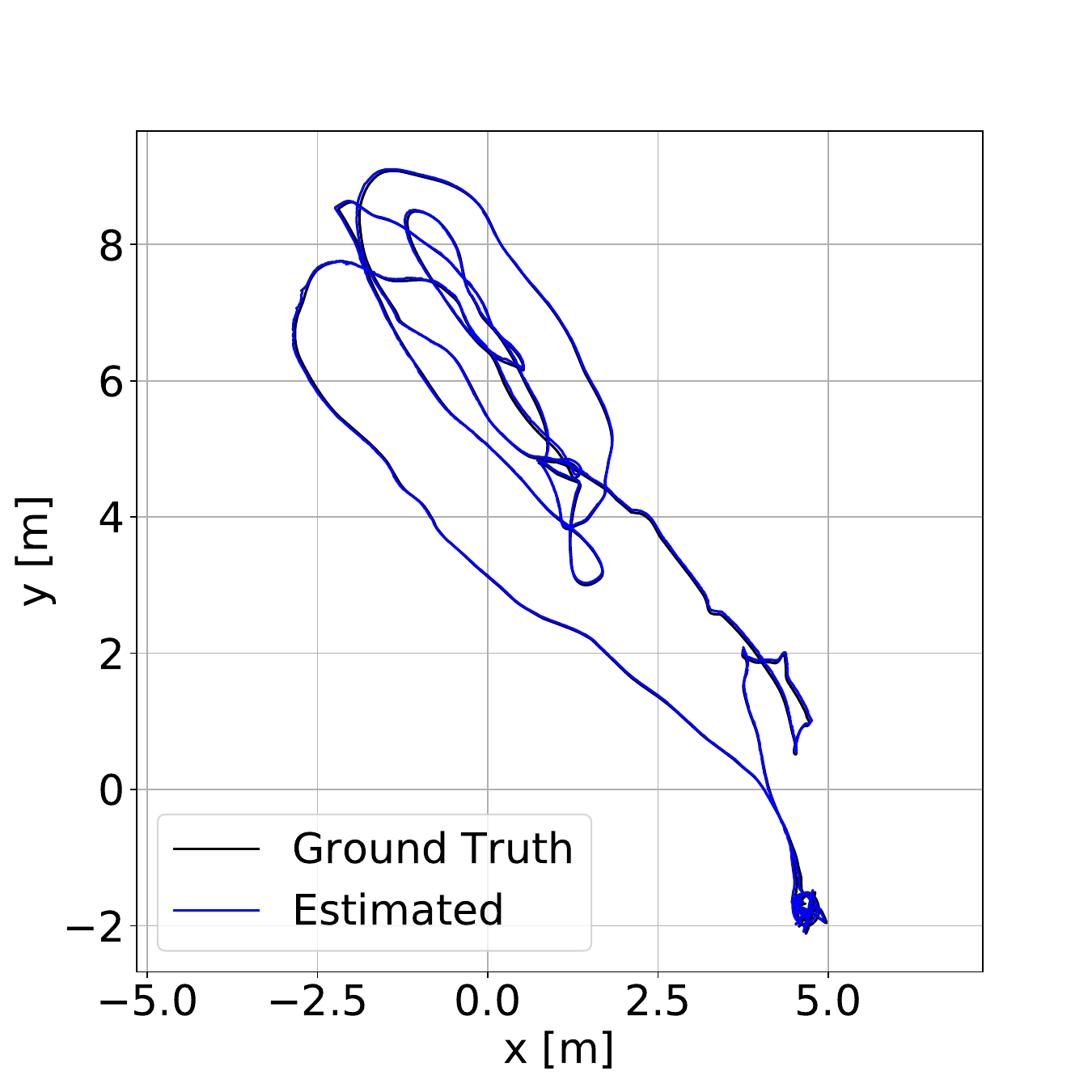}
    \caption{No noise; \textit{MH01} sequence.}
    \label{fig:sub1}
  \end{subfigure}%
  \hfill
  \begin{subfigure}{0.33\textwidth}
    \centering
    \includegraphics[width=\linewidth]{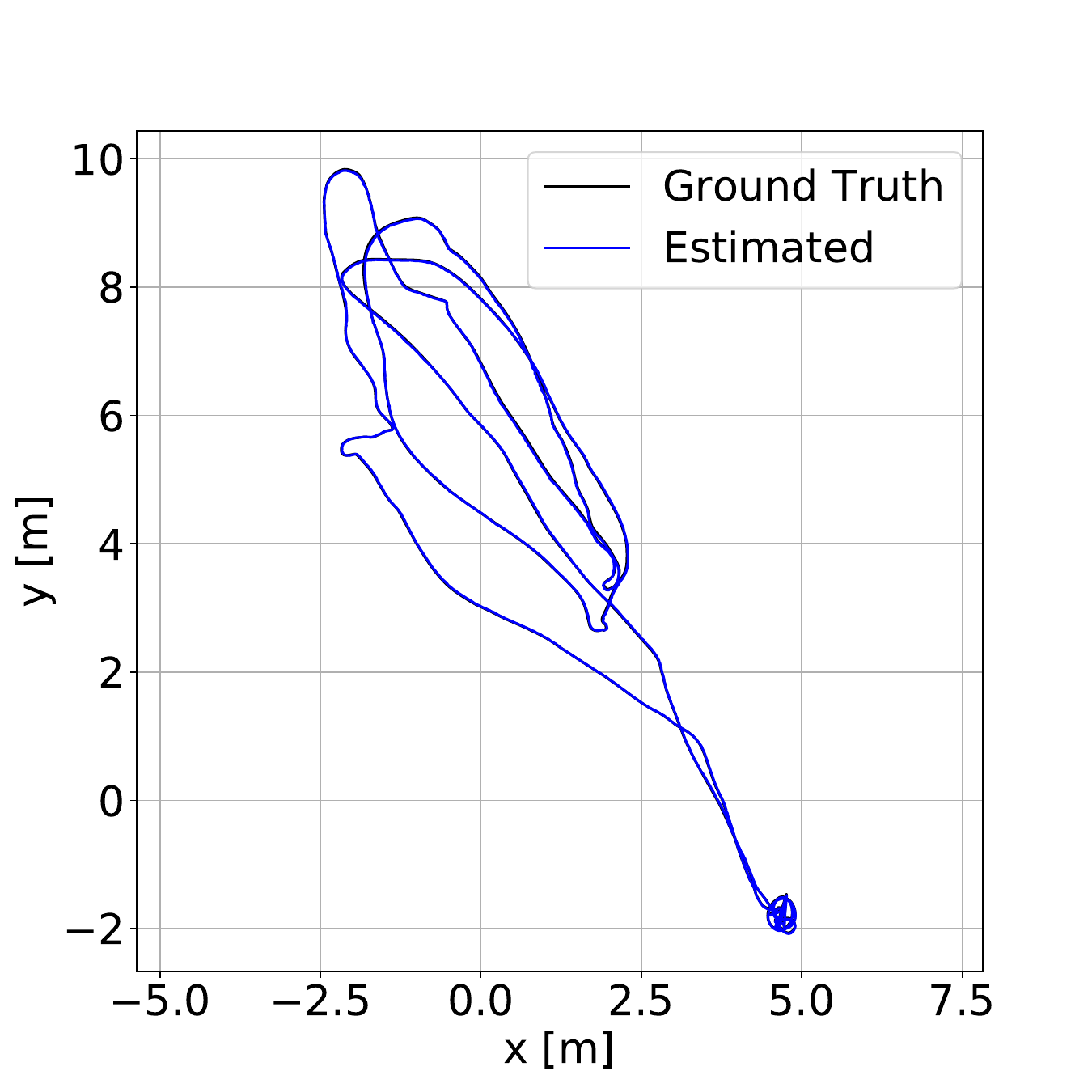}
    \caption{No noise; \textit{MH02} sequence.}
    \label{fig:sub2}
  \end{subfigure}%
  \hfill
  \begin{subfigure}{0.33\textwidth}
    \centering
    \includegraphics[width=\linewidth]{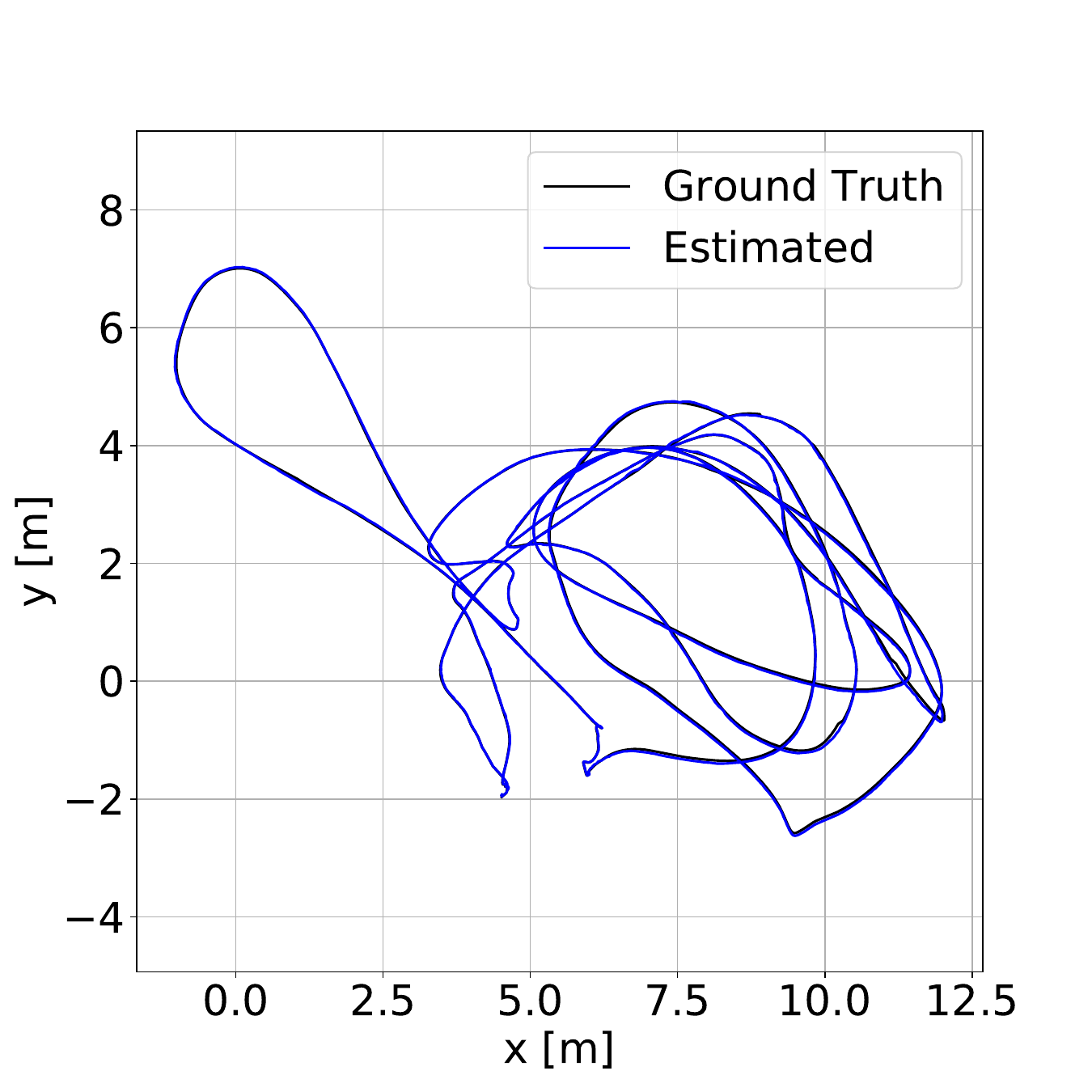}
    \caption{No noise; \textit{MH03} sequence.}
    \label{fig:sub3}
  \end{subfigure}

  \vspace{0.2cm}
  \begin{subfigure}{0.33\textwidth}
    \centering
    \includegraphics[width=\linewidth]{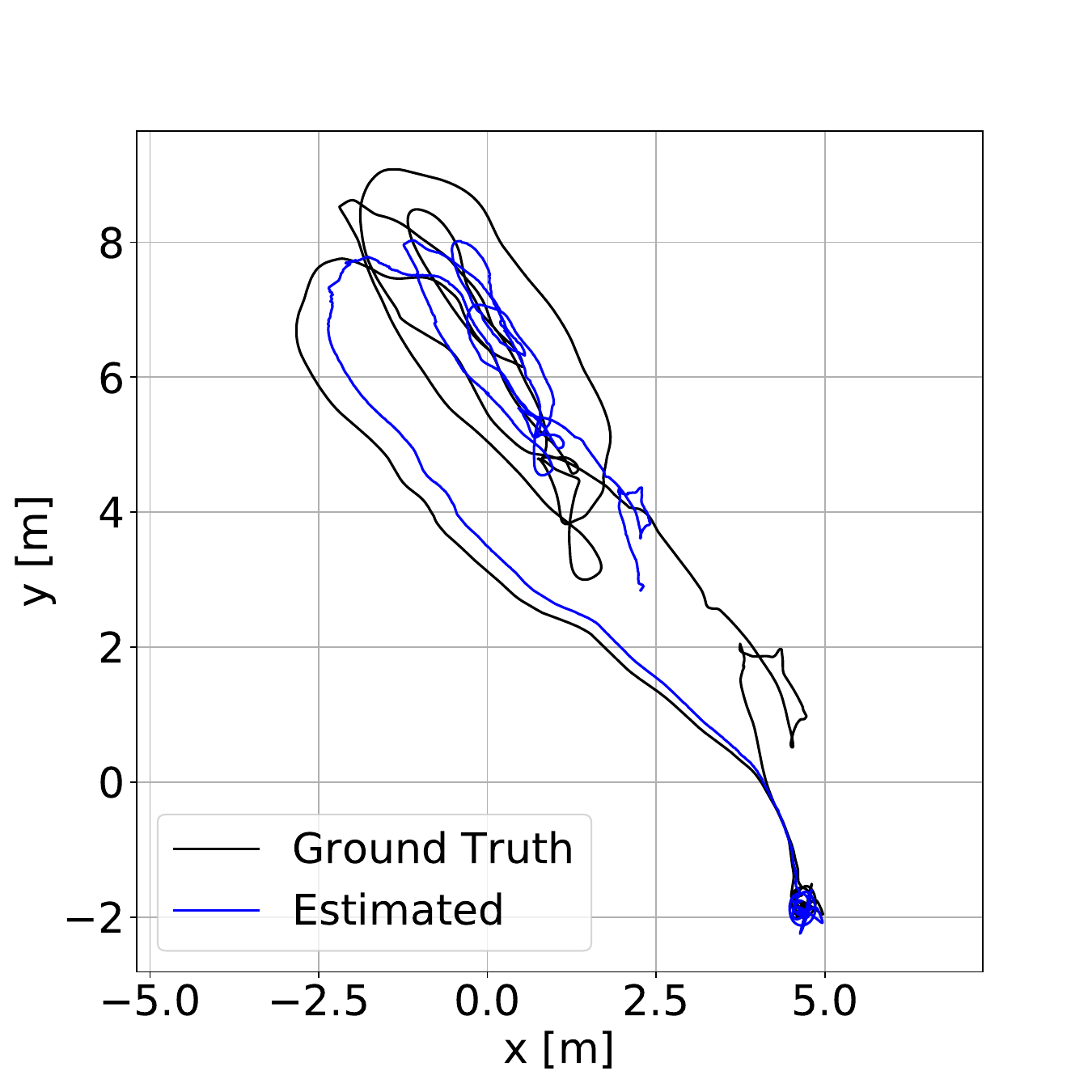}
    \caption{$\eta \sim \mathcal{N}(0, 0.001)$; \textit{MH01} sequence.}
    \label{fig:sub1}
  \end{subfigure}%
  \hfill
  \begin{subfigure}{0.33\textwidth}
    \centering
    \includegraphics[width=\linewidth]{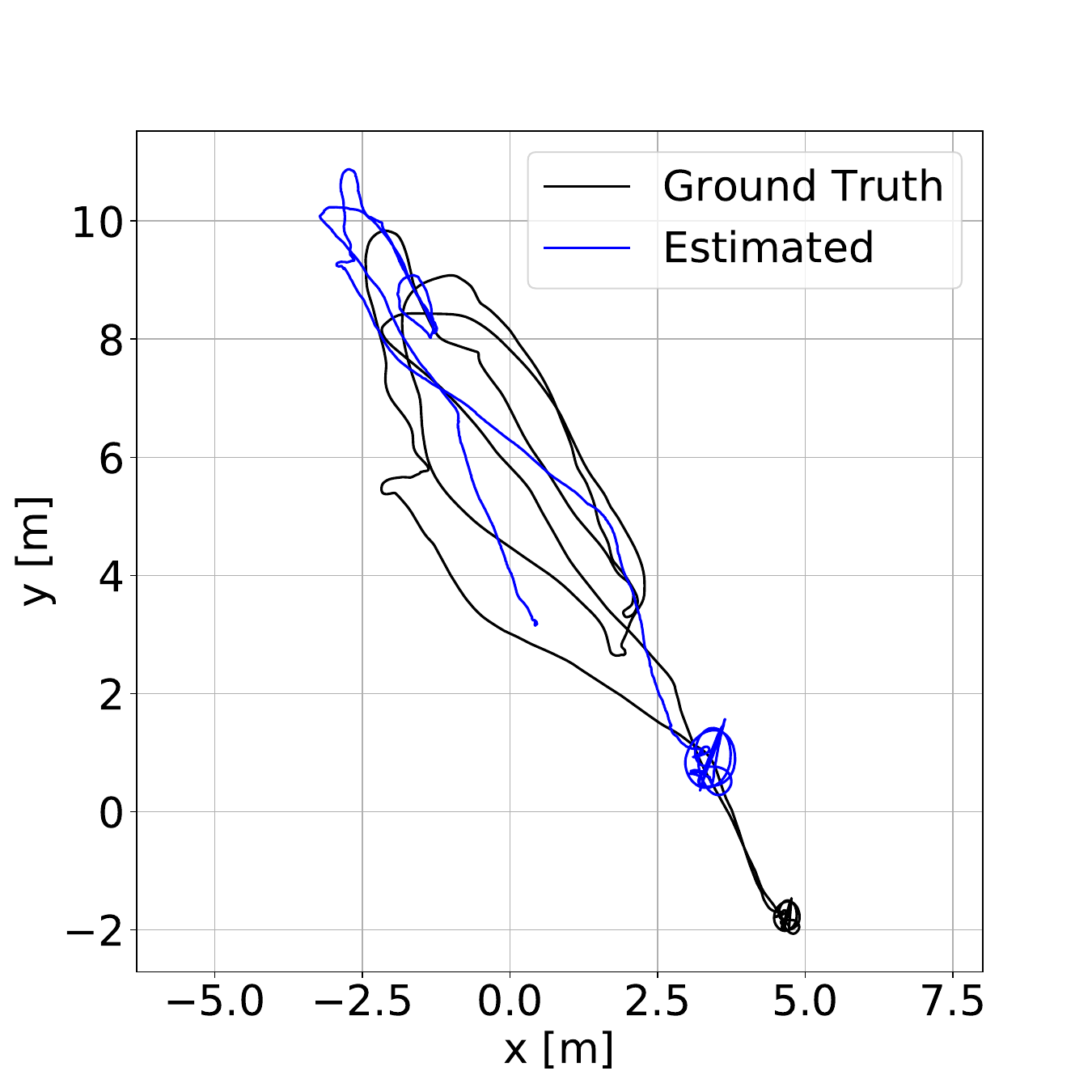}
    \caption{$\eta \sim \mathcal{N}(0, 0.001)$; \textit{MH02} sequence.}
    \label{fig:sub2}
  \end{subfigure}%
  \hfill
  \begin{subfigure}{0.33\textwidth}
    \centering
    \includegraphics[width=\linewidth]{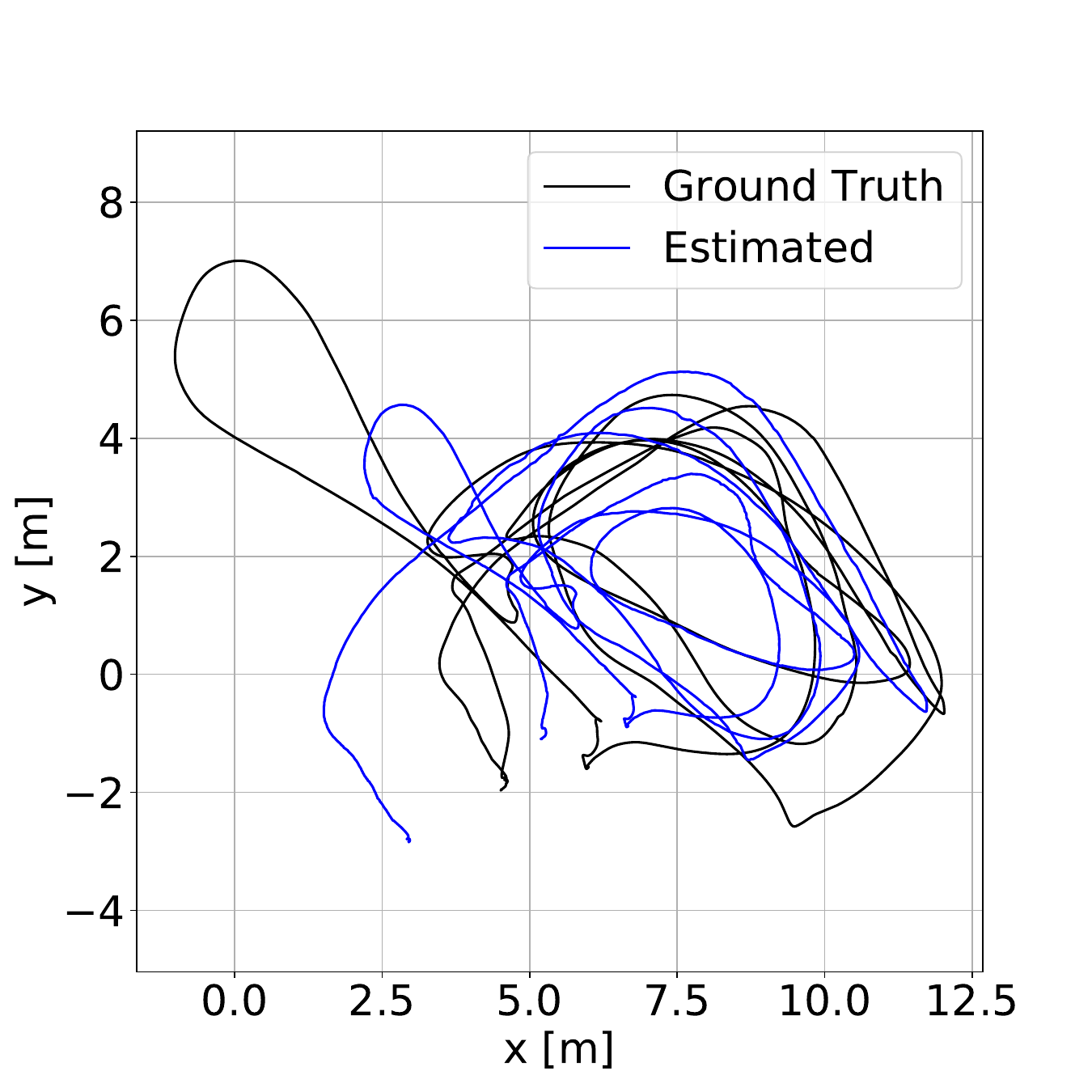}
    \caption{$\eta \sim \mathcal{N}(0, 0.001)$; \textit{MH03} sequence.}
    \label{fig:sub3}
  \end{subfigure}

  \vspace{0.2cm}

  \begin{subfigure}{0.33\textwidth}
    \centering
    \includegraphics[width=\linewidth]{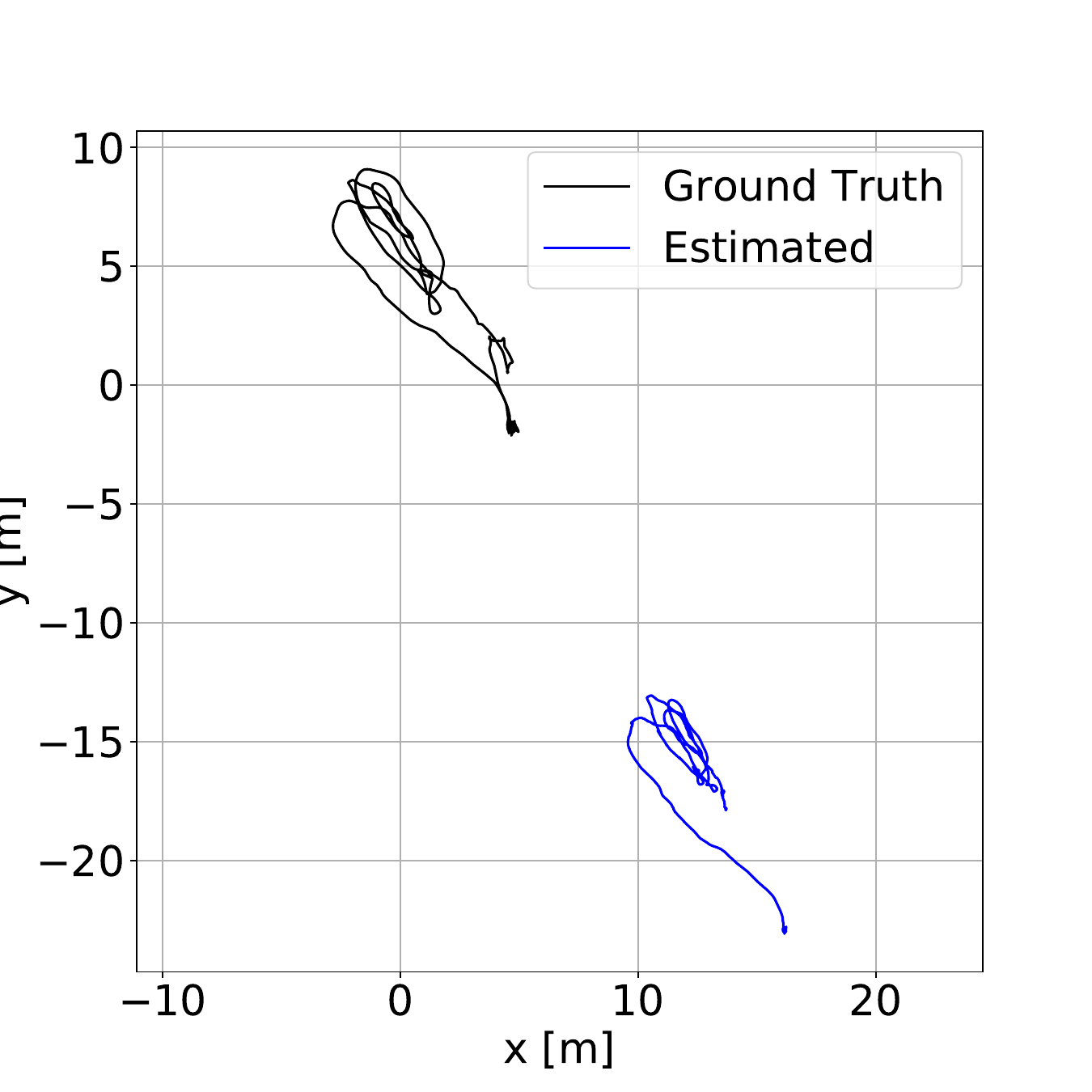}
    \caption{$\eta \sim \mathcal{N}(0, 0.01)$; \textit{MH01} sequence.}
    \label{fig:sub1}
  \end{subfigure}%
  \hfill
  \begin{subfigure}{0.33\textwidth}
    \centering
    \includegraphics[width=\linewidth]{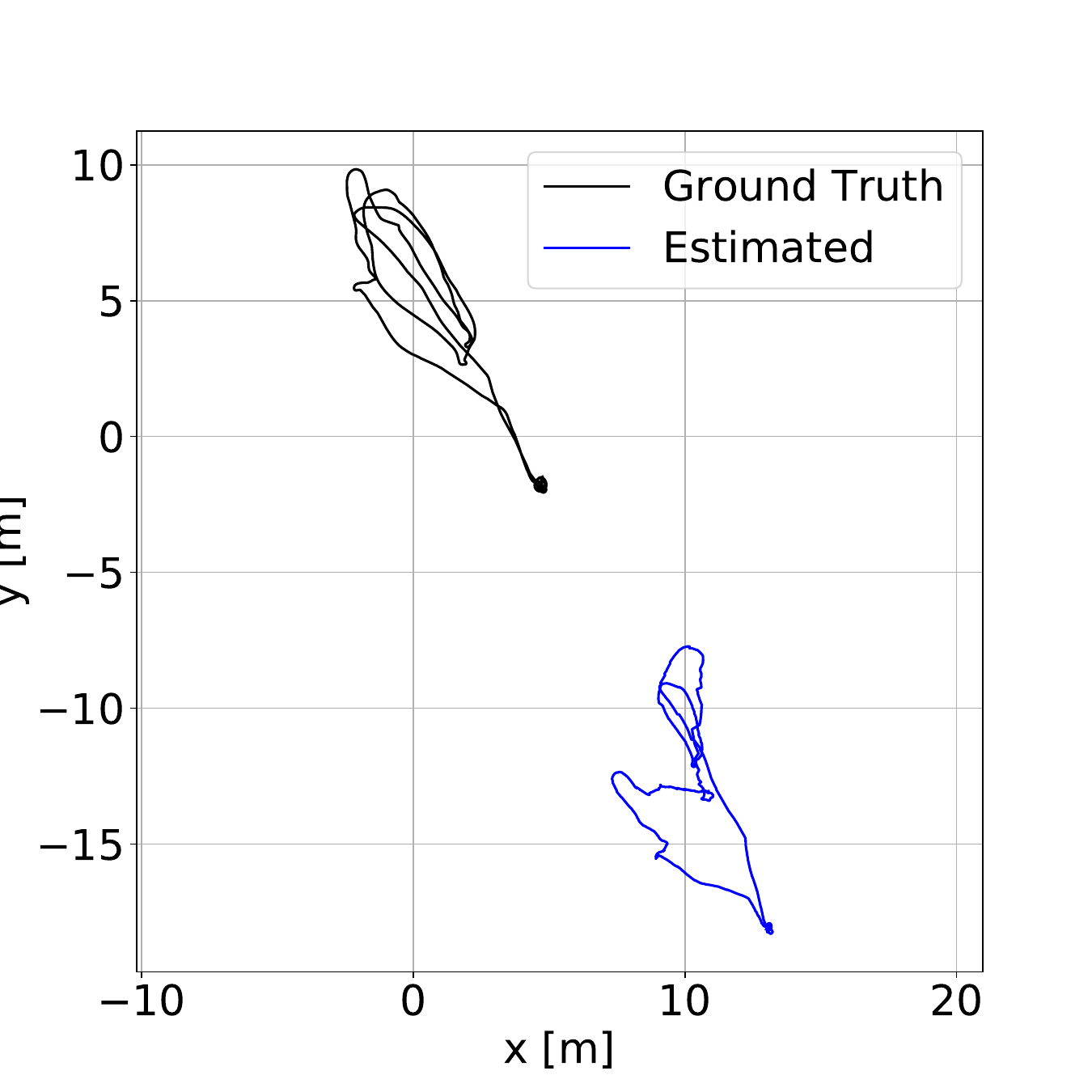}
    \caption{$\eta \sim \mathcal{N}(0, 0.01)$; \textit{MH02} sequence.}
    \label{fig:sub2}
  \end{subfigure}%
  \hfill
  \begin{subfigure}{0.33\textwidth}
    \centering
    \includegraphics[width=\linewidth]{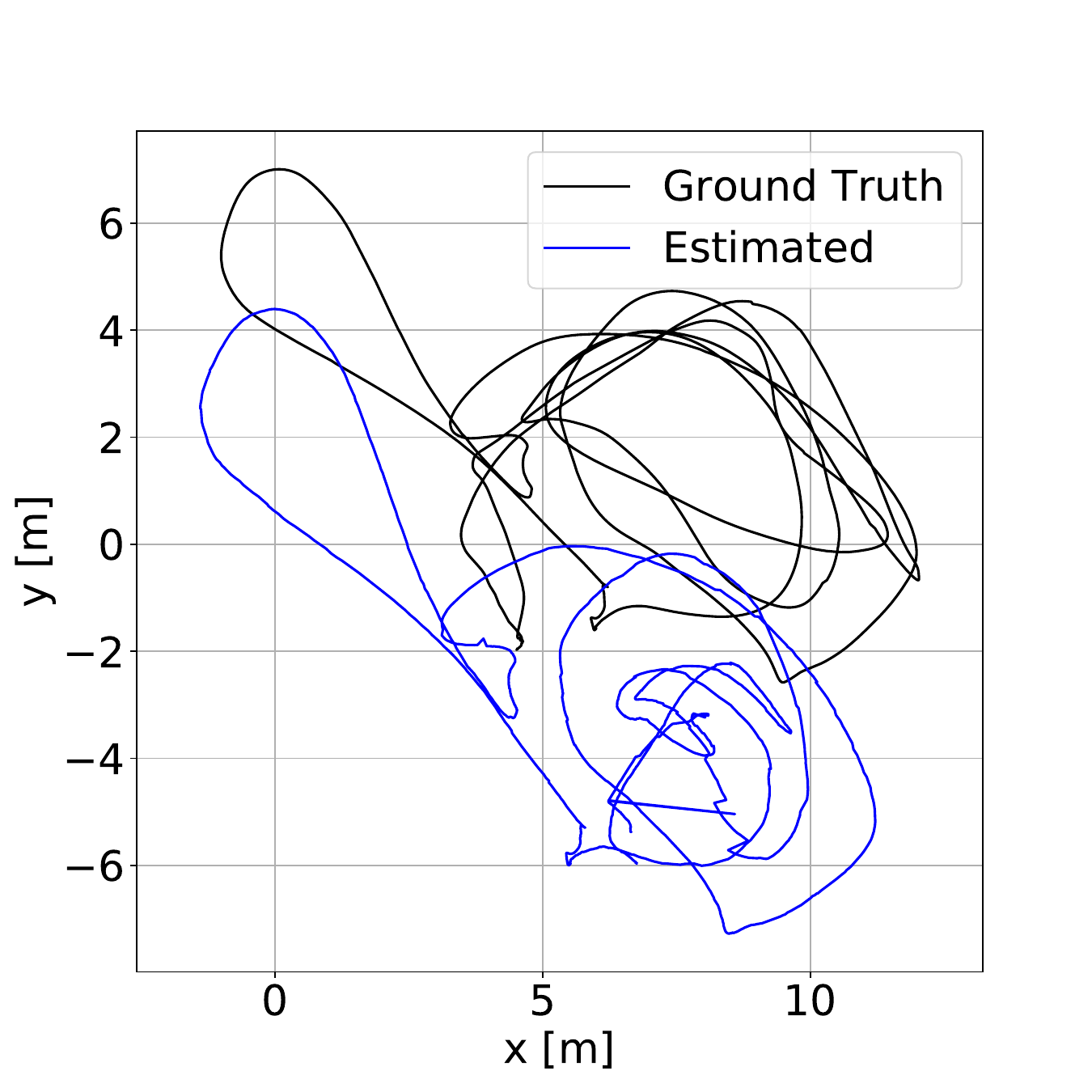}
    \caption{$\eta \sim \mathcal{N}(0, 0.01)$; \textit{MH03} sequence.}
    \label{fig:sub3}
  \end{subfigure}
	\caption{\textbf{Effect of perturbations on the stereo camera baseline for the ORBSLAM3~\cite{orbslam3} SLAM model with the stereo input setting.} We introduce a random noise $\eta$ along the axis of the stereo camera baseline. The noise is sampled from a Gaussian distribution with zero mean and variance $\sigma^2$, represented as $\eta \sim \mathcal{N}(0, \sigma^2)$. Here, we show the qualitative comparisons on three sequences of the EuRoC~\cite{Burri25012016} dataset. } 
	\label{fig:stereo-slam-baseline-perturb-more-cases}
\end{figure*}

\begin{figure*}[t]
	\centering
  \centering

  \begin{subfigure}{0.33\textwidth}
    \centering
    \includegraphics[width=\linewidth]{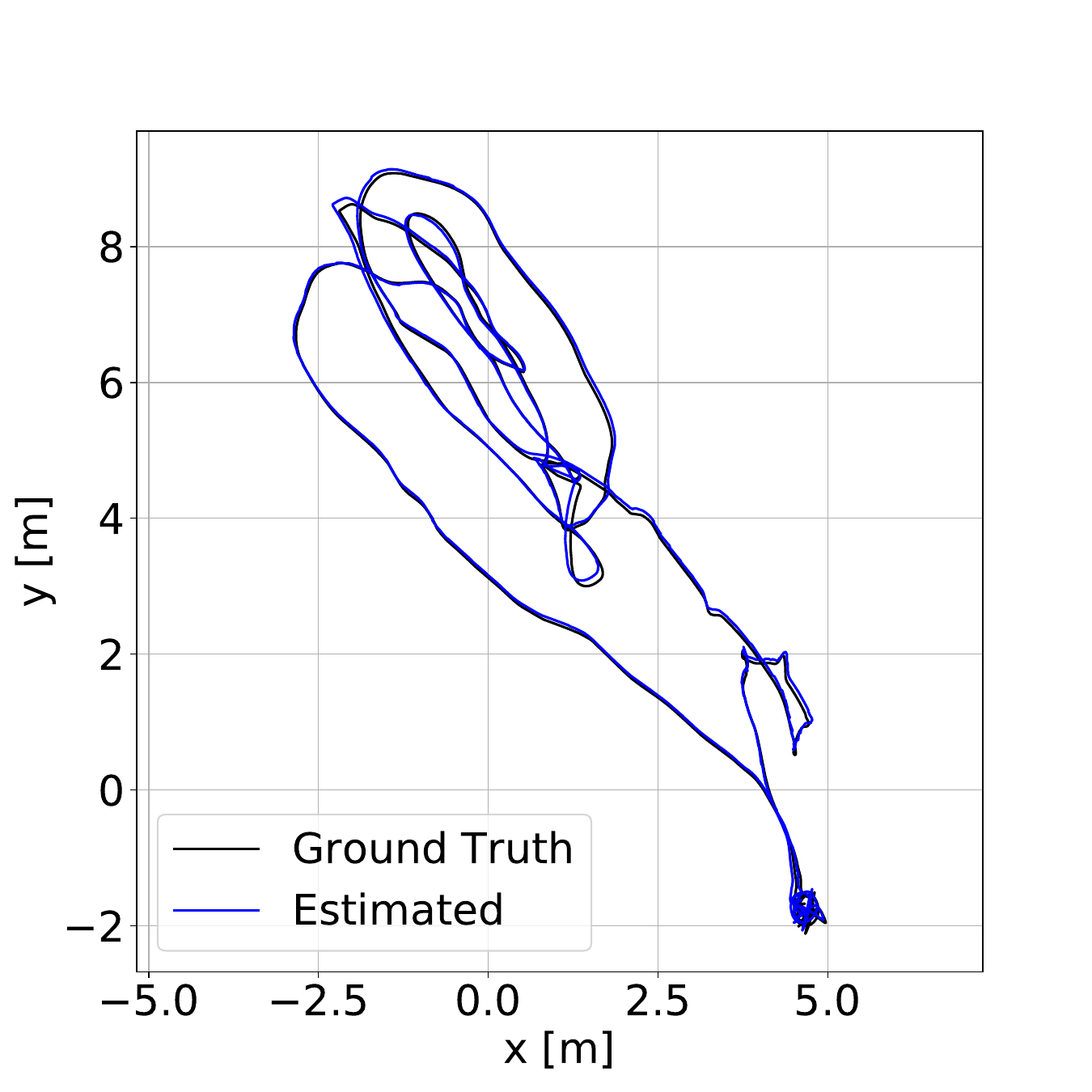}
    \caption{No noise; \textit{MH01} sequence.}
    \label{fig:sub1}
  \end{subfigure}%
  \hfill
  \begin{subfigure}{0.33\textwidth}
    \centering
    \includegraphics[width=\linewidth]{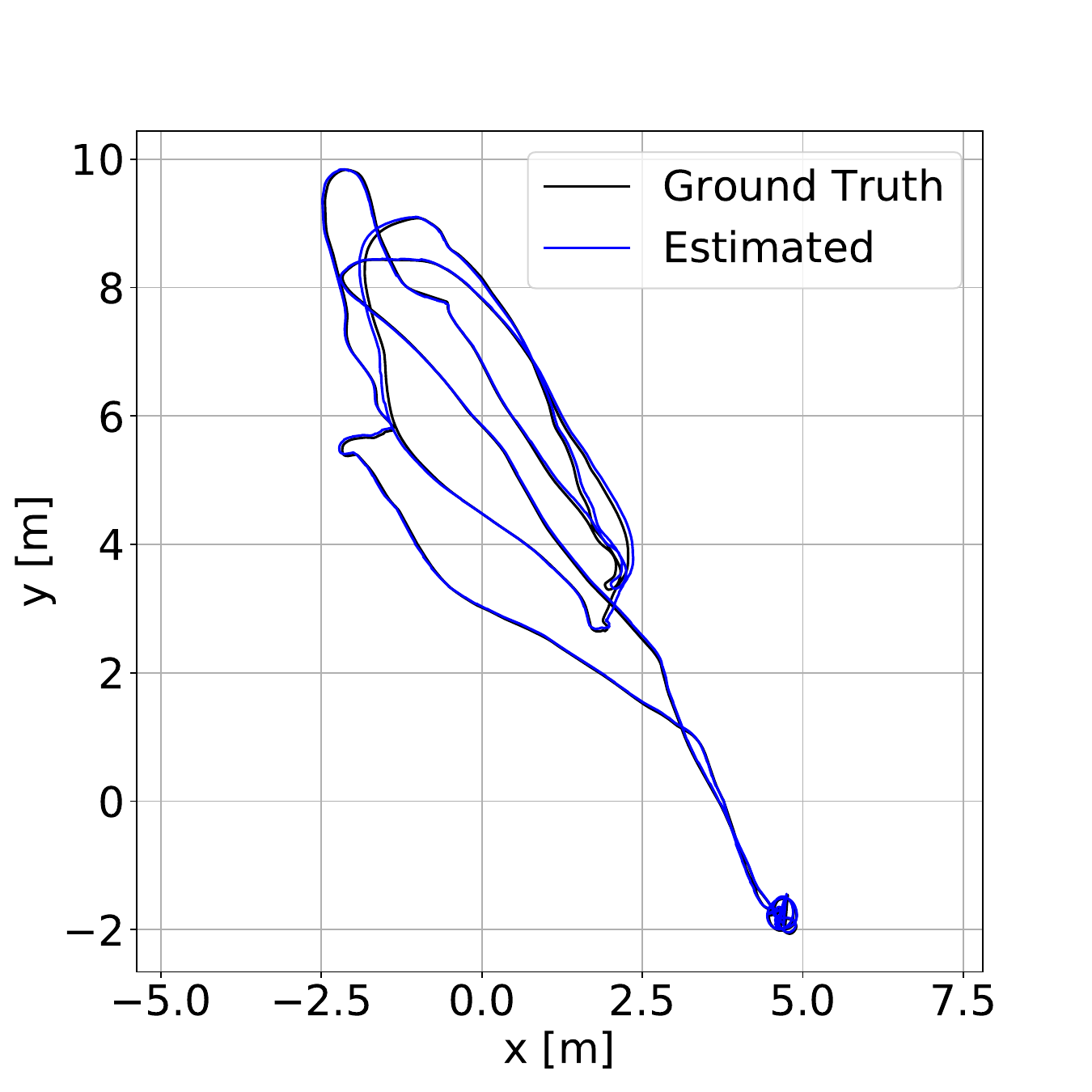}
    \caption{No noise; \textit{MH02} sequence.}
    \label{fig:sub2}
  \end{subfigure}%
  \hfill
  \begin{subfigure}{0.33\textwidth}
    \centering
    \includegraphics[width=\linewidth]{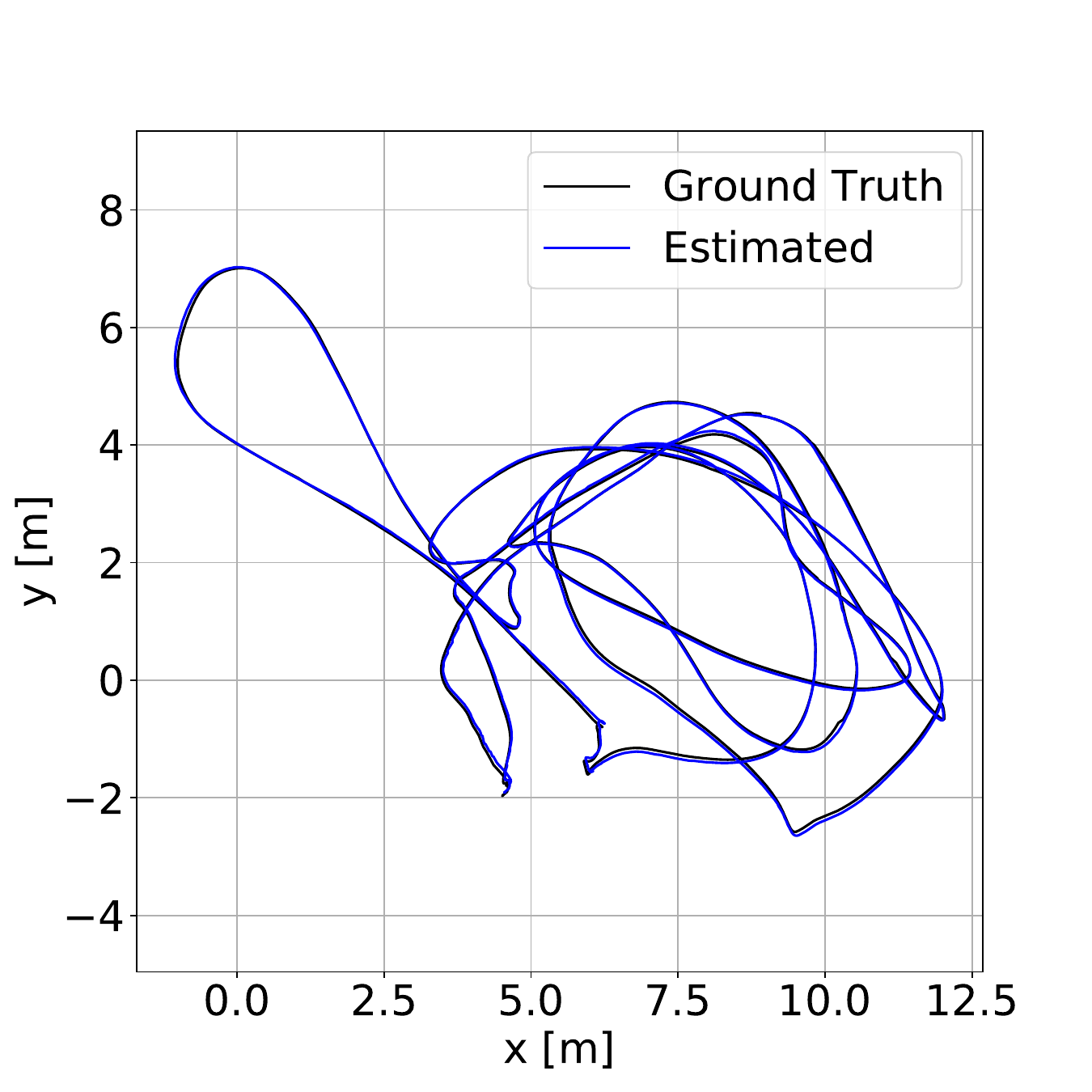}
    \caption{No noise; \textit{MH03} sequence.}
    \label{fig:sub3}
  \end{subfigure}

  \vspace{0.2cm}
  \begin{subfigure}{0.33\textwidth}
    \centering
    \includegraphics[width=\linewidth]{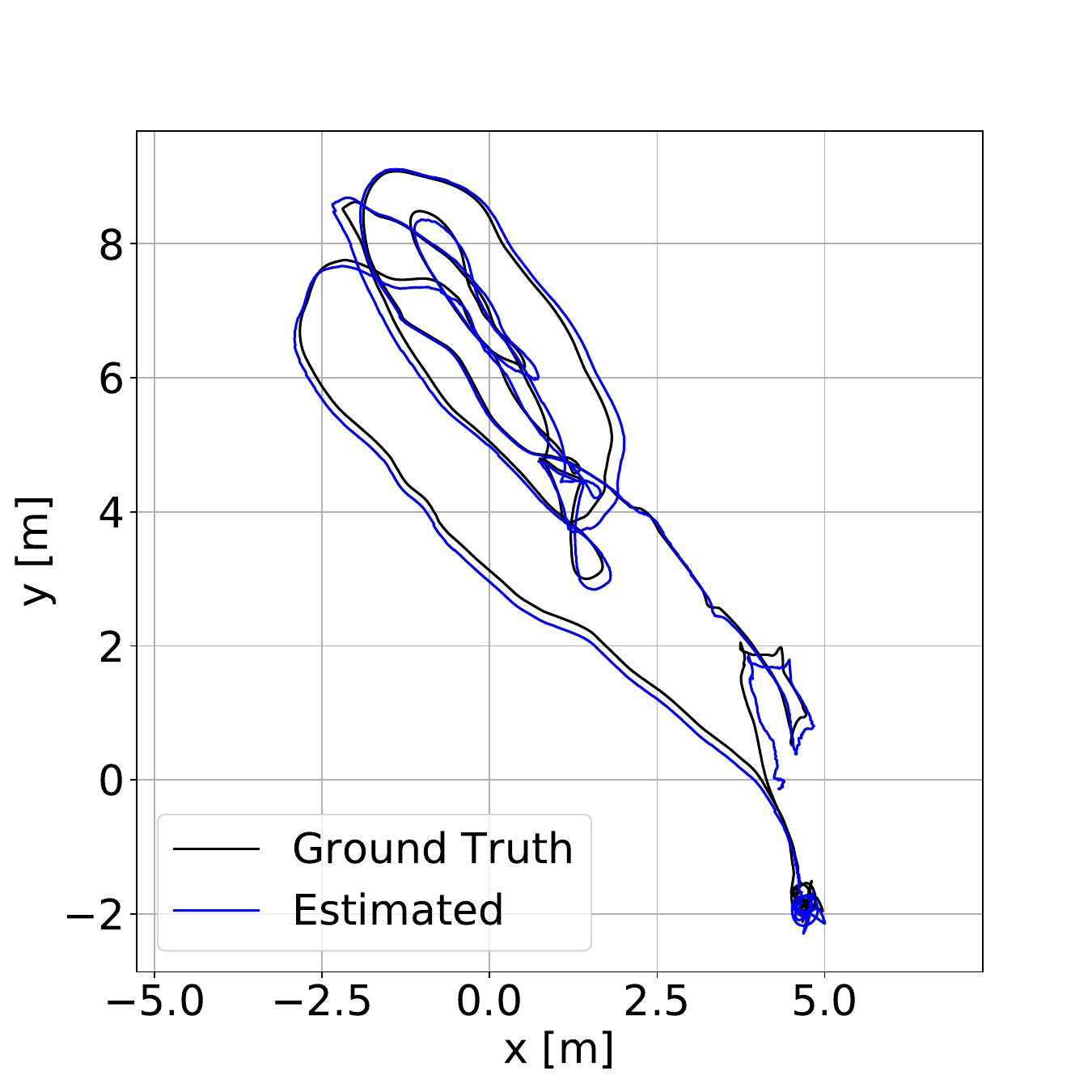}
    \caption{$\eta \sim \mathcal{N}(0, 0.001)$; \textit{MH01} sequence.}
    \label{fig:sub1}
  \end{subfigure}%
  \hfill
  \begin{subfigure}{0.33\textwidth}
    \centering
    \includegraphics[width=\linewidth]{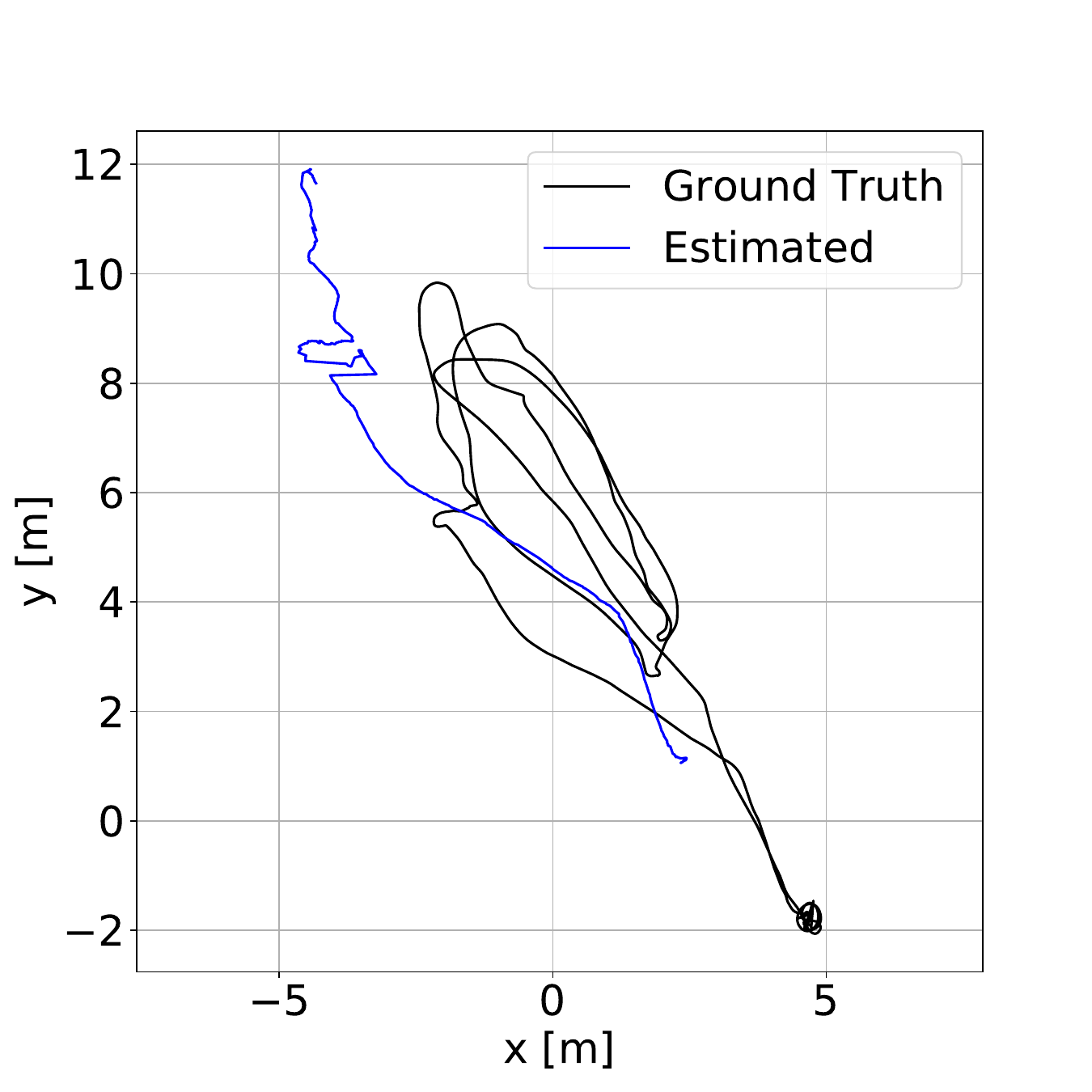}
    \caption{$\eta \sim \mathcal{N}(0, 0.001)$; \textit{MH02} sequence.}
    \label{fig:sub2}
  \end{subfigure}%
  \hfill
  \begin{subfigure}{0.33\textwidth}
    \centering
    \includegraphics[width=\linewidth]{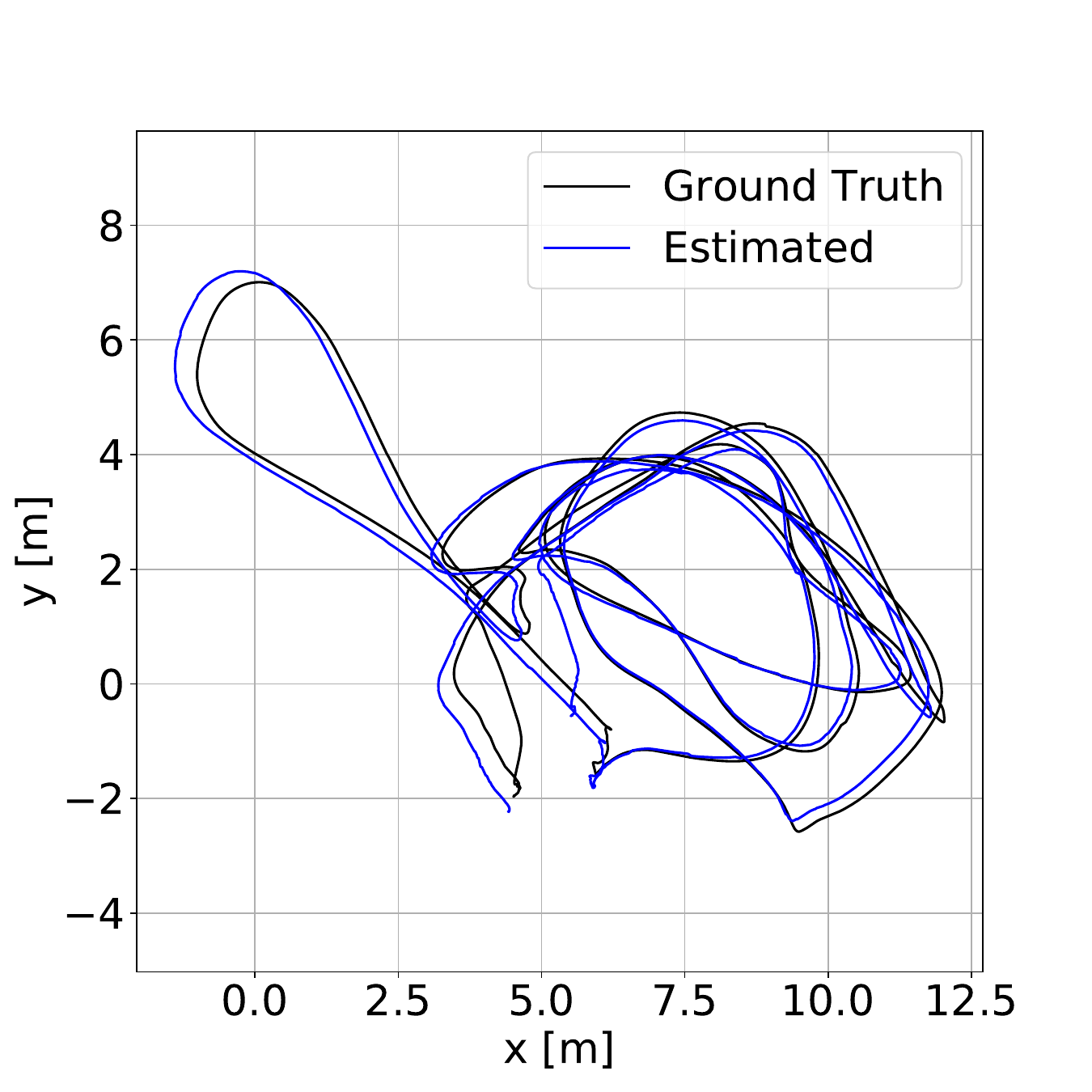}
    \caption{$\eta \sim \mathcal{N}(0, 0.001)$; \textit{MH03} sequence.}
    \label{fig:sub3}
  \end{subfigure}

  \vspace{0.2cm}

  \begin{subfigure}{0.33\textwidth}
    \centering
    \includegraphics[width=\linewidth]{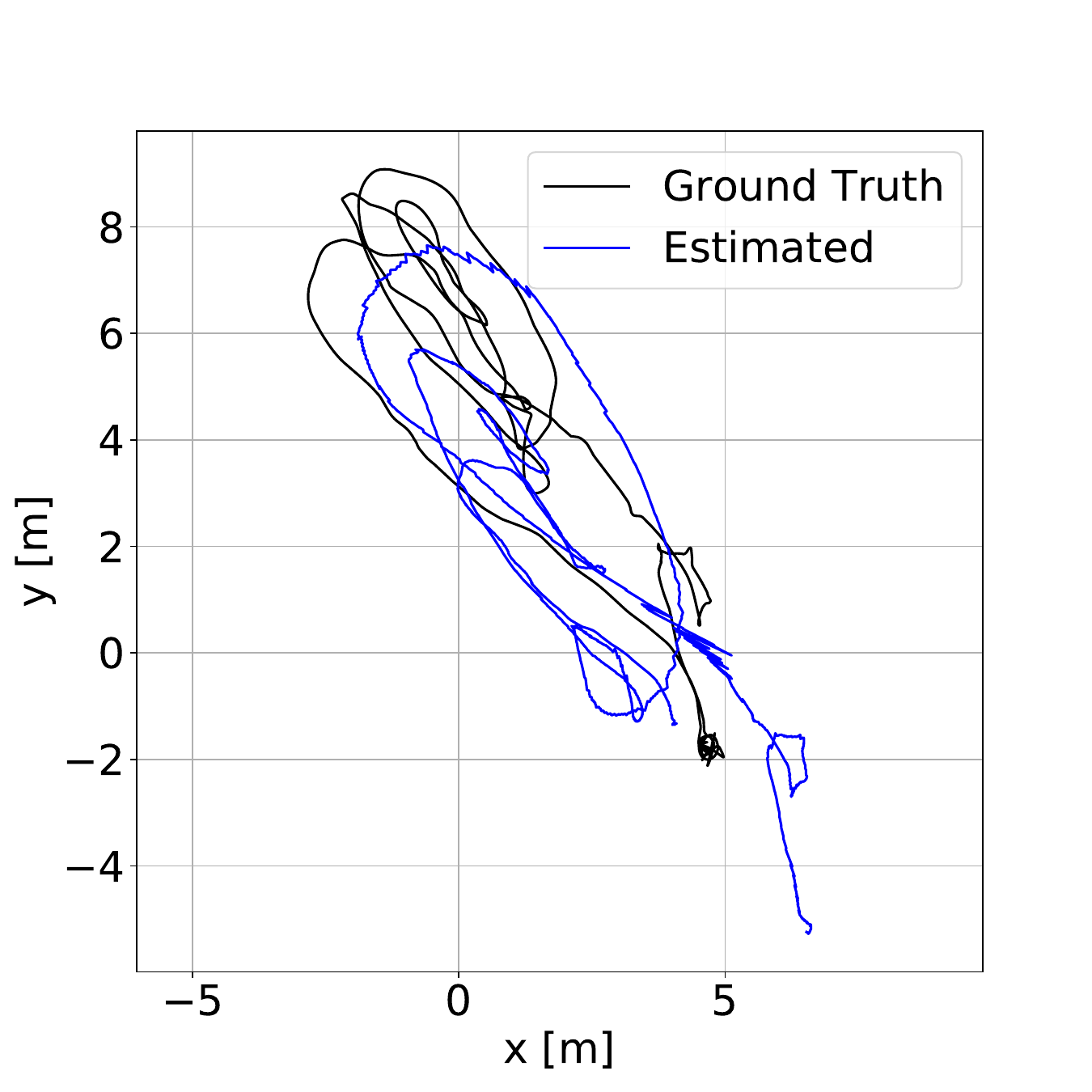}
    \caption{$\eta \sim \mathcal{N}(0, 0.01)$; \textit{MH01} sequence.}
    \label{fig:sub1}
  \end{subfigure}%
  \hfill
  \begin{subfigure}{0.33\textwidth}
\hspace{-6mm}
    \includegraphics[width=\linewidth]{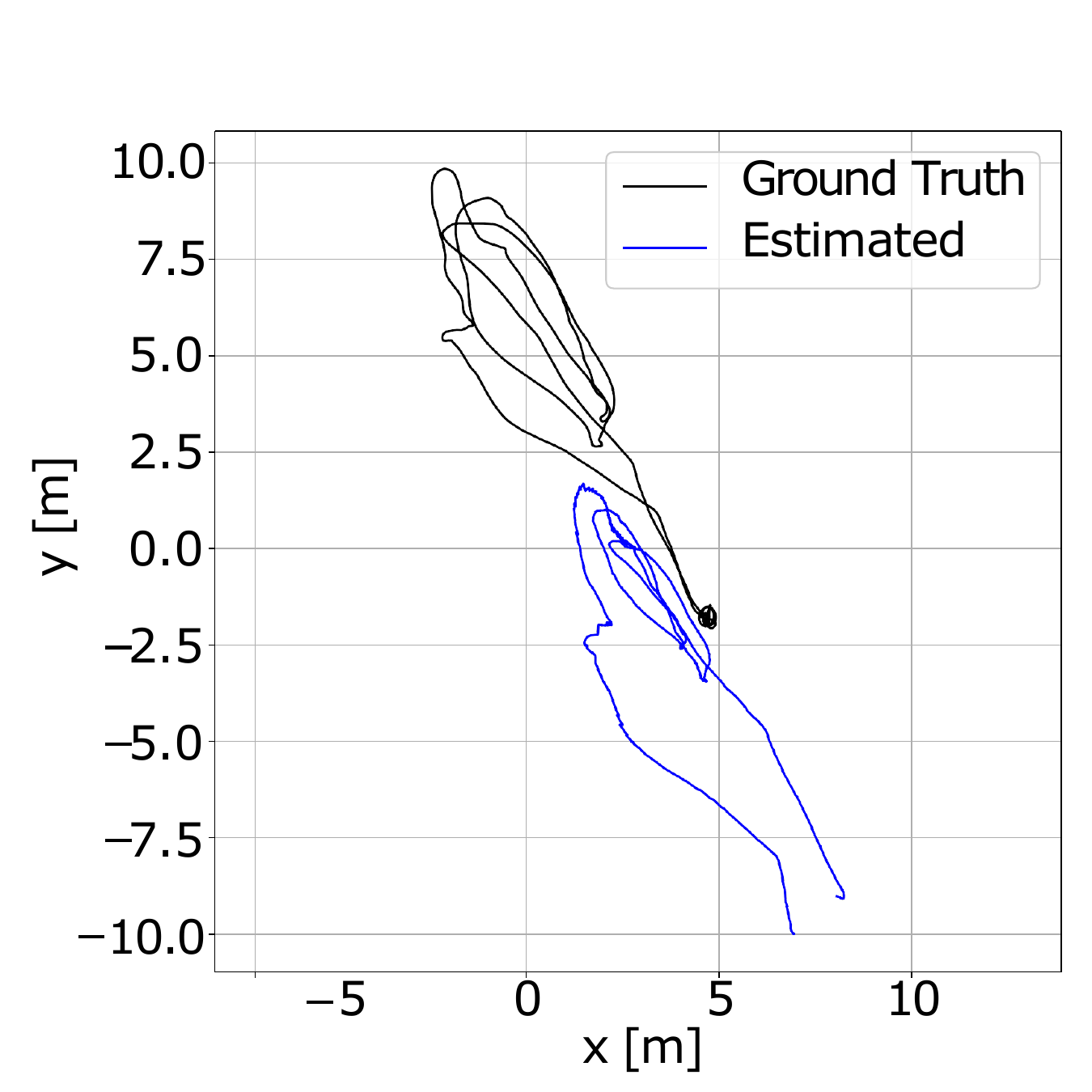}
    \caption{$\eta \sim \mathcal{N}(0, 0.01)$; \textit{MH02} sequence.}
    \label{fig:sub2}

  \end{subfigure}%
  \hfill 
  \begin{subfigure}{0.33\textwidth}
    \centering
    \includegraphics[width=\linewidth]{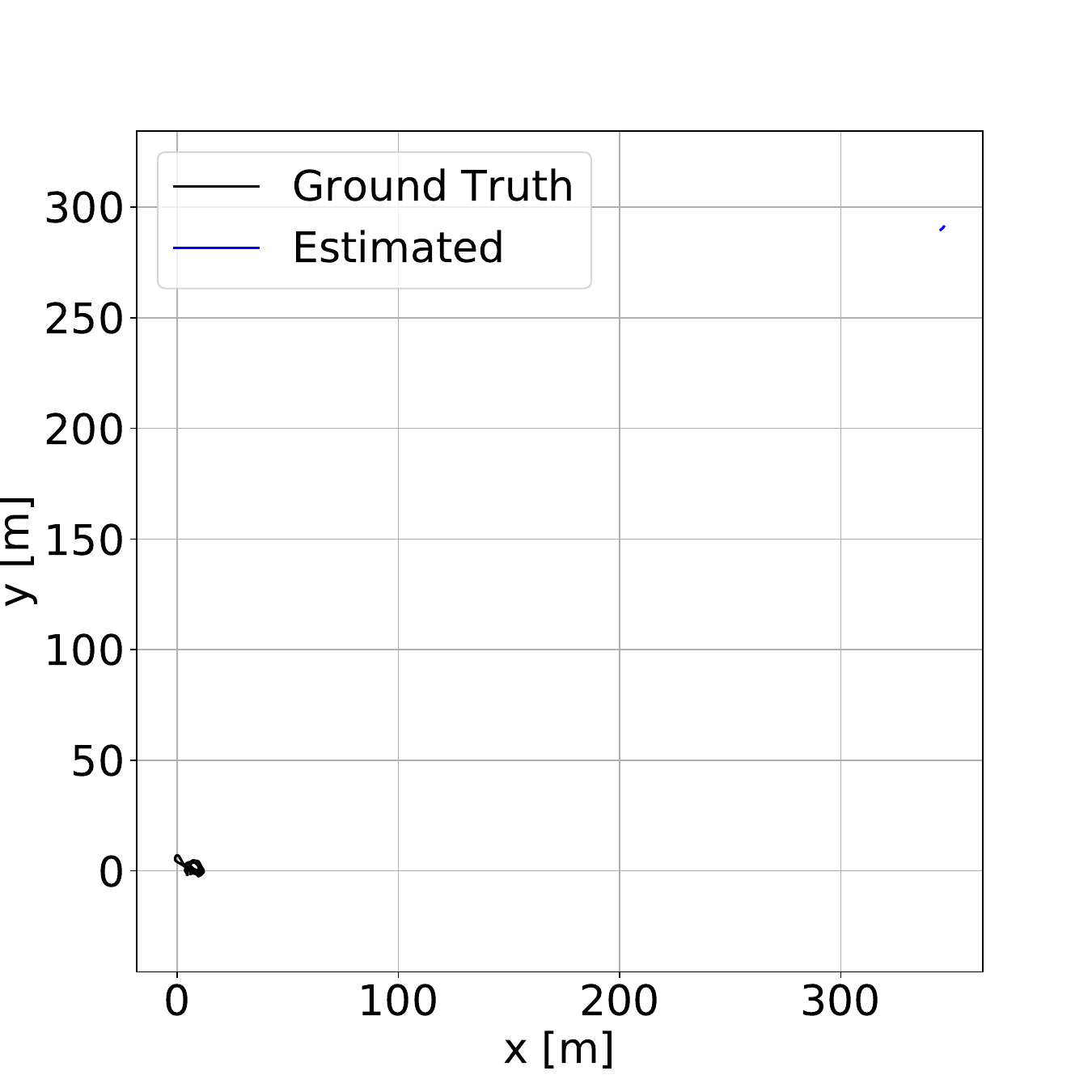}
    \caption{$\eta \sim \mathcal{N}(0, 0.01)$; \textit{MH03} sequence.}
    \label{fig:sub3}
  \end{subfigure}
	\caption{\textbf{Effect of perturbations on the stereo camera baseline for the ORBSLAM3~\cite{orbslam3} SLAM model with the stereo-inertia input setting.} We introduce a random noise $\eta$ along the axis of the stereo camera baseline. The noise is sampled from a Gaussian distribution with zero mean and variance $\sigma^2$, represented as $\eta \sim \mathcal{N}(0, \sigma^2)$. Here, we show the qualitative comparisons on three sequences of the EuRoC~\cite{Burri25012016} dataset. } 
	\label{fig:stereo-slam-baseline-perturb-more-cases-with-imu}
\end{figure*}

\section{Robustness Analysis of Stereo SLAM System Under Perturbation}\label{sec:stereo-slam-case-study}

\noindent\textbf{Quantitative analysis.} We conduct a robustness analysis of a stereo SLAM system, which utilizes two cameras to capture observations of the environment. In this study, we focus on the inter-camera transformation matrix, a crucial sensor parameter that describes the transformation between the left and right cameras. Our objective is to investigate the effects of introducing Gaussian noise as the perturbation to the inter-camera transformation matrix of the stereo SLAM system, which mimics the inaccuracy of the transformation matrix. This inaccuracy can be caused by factors such as noisy extrinsic matrix calibration, vibration due to a less rigid camera holder, or errors in camera placement or assembly. Specifically, we introduce random noise $\eta$ along the stereo camera baseline direction, sampled from a Gaussian distribution with zero mean and variance $\sigma^2$, denoted as $\eta \sim N(0, \sigma^2)$.
In our preliminary robustness study, we utilize the ORB-SLAM3 model with the stereo camera input setting and the stereo-inertia setting. We apply random perturbations to the camera baseline length for every frame in three sequences (MH01, MH02, and NM03) from the EuRoC dataset \cite{Burri25012016}. The results of our study are summarized in Table~\ref{tab:stereo-slam-robustness-case-study}, which includes three severity levels of noises: $\sigma = 0.00$, $\sigma = 0.001$, and $\sigma = 0.01$. 
Under the clean setting ($\sigma = 0.00$), 
when we introduced perturbations to the baseline axis of the inter-camera transformation matrix, increasing the noise levels (\textit{e.g.}, $\sigma = 0.001$ and $\sigma = 0.01$) resulted in a noticeable rise in trajectory estimation error, which is quantified by \textit{ATE-w/o Scale}  and  \textit{ATE-w/ Scale}  metrics. This indicates a higher susceptibility to errors and deviations from the ground truth trajectory during trajectory estimation. These findings highlight the significance of taking a holistic approach when considering the robustness of the whole SLAM system. To ensure a comprehensive evaluation of the SLAM model's robustness, it is crucial to consider not only the software (SLAM model) but also the hardware, specifically the inaccuracies in predefined or estimated sensor configurations. These inaccuracies can significantly affect the overall system's robustness.

\noindent\textbf{Qualitative analysis.} In Fig.\ref{fig:stereo-slam-baseline-perturb-more-cases} and Fig.\ref{fig:stereo-slam-baseline-perturb-more-cases-with-imu}, we present qualitative analyses illustrating the impact of noises on the stereo camera baseline. We evaluate the trajectory estimation performance for both the stereo and stereo-inertia SLAM settings using three sequences from the EuRoC dataset.
The qualitative comparisons reveal that the estimated trajectory becomes noisier as the level of noise is increased. Notably, even a slight random noise (Gaussian noise with zero mean and $\sigma = 0.001$ [m]) added to the camera baseline leads to a significant degradation in trajectory estimation for all evaluated sequences. In addition, our analysis reveals that when the variance of random deviations of the stereo camera baseline reaches $\sigma = 0.01$ [m], the stereo SLAM system exhibits a complete failure in both the stereo and stereo-inertia settings. 
These findings emphasize the vulnerability of the stereo SLAM system to inaccuracies in the sensor/camera extrinsics. The observed degradation in trajectory estimation highlights the need for precise calibration, robust camera holder design, and accurate camera calibration to ensure reliable performance of the stereo SLAM system.

\section{Robustness Analysis of Multi-agent SLAM System Under Perturbation}\label{sec:multi-agent-slam-robustness}

We investigate the robustness of the multi-agent SLAM framework COVINS-G~\cite{covins-g} under Gaussian noise image-level perturbation. We evaluate the performance of the model on the EuRoC~\cite{Burri25012016}  dataset using the ORB-SLAM3~\cite{orbslam3} model with monocular-inertia input as the front-end module. The experimental setup involves introducing varying severity levels of perturbation, measured by the parameter $\sigma$, on all the agents of a multi-agent system. Two metrics, Average Trajectory Error (ATE) and Sum of Squared Errors (SSE), are used to assess the multi-agent SLAM system's performance. These metrics provide a comprehensive evaluation of trajectory estimation accuracy and the accumulation of errors, respectively.
As is shown in Table~\ref{tab:metric-covinsg}, the multi-agent SLAM system demonstrates consistent performance in the clean setting ($\sigma = 0.00$) and low perturbation severity ($\sigma = 0.02$), indicating the system's ability to tolerate low random noise. 
However, as the perturbation severity increases to $\sigma = 0.04$, the system's performance degrades noticeably. The ATE experiences a significant increase, suggesting the vulnerability of the model to higher levels of perturbation. Similarly, the SSE metric rises to 20.354 cm, reflecting the accumulation of errors caused by the higher noise level.

\begin{table}[t]
\caption{\textbf{Effects of Gaussian noise perturbation on trajectory estimation performance} for the multi-agent SLAM framework COVINS-G~\cite{covins-g} with the ORBSLAM3~\cite{orbslam3} model as the front-end are presented.  The noise is sampled from a Gaussian distribution with zero mean and variance $\sigma^2$, represented as $\eta \sim \mathcal{N}(0, \sigma^2)$. The experiments were conducted using a mono-inertia multi-modal input setting for each agent.  The results are reported based on running five agents in sequential order on the \textit{MH} scene of the EuRoC~\cite{Burri25012016} dataset.}
\label{tab:metric-covinsg}
\centering 
\setlength{\tabcolsep}{3.0mm}
\resizebox{0.48\textwidth}{!}{
\begin{tabular}{l|c|cc}
    \toprule \toprule   
    \multirow{1}{*}{\textbf{Metrics}}  &     {$\sigma^2=0.00$} &     {$\sigma^2=0.02$}
     &  {$\sigma^2=0.04$}  \\ \midrule
   ATE$\downarrow$ [cm] & $\textbf{0.071}$ & $0.071$ & $0.088$ 
    \\
    SSE$\downarrow$ [cm] & $\textbf{11.545}$ & $12.517$ & $20.354$ 
    \\
    \bottomrule \bottomrule
    \multicolumn{4}{l}{{\textbf{1}) The best performance for each metric is in \textbf{bold}.}}\\
    \multicolumn{4}{l}{\textbf{2}) SSE indicates sum of squared errors metric. } \\
        \multicolumn{4}{l}{  Please refer to COVINS-G~\cite{covins-g} paper for more details.}
    \end{tabular}
}
\end{table}

\section{Qualitative Results of SLAM Model Performance under Perturbations}\label{sec:qualitative_results}

This section presents qualitative results of trajectory estimation and 3D reconstruction in SLAM models, highlighting both successful and failure cases under specific perturbations.

\begin{figure*}[ht!]
	\centering
\includegraphics[width=\textwidth]{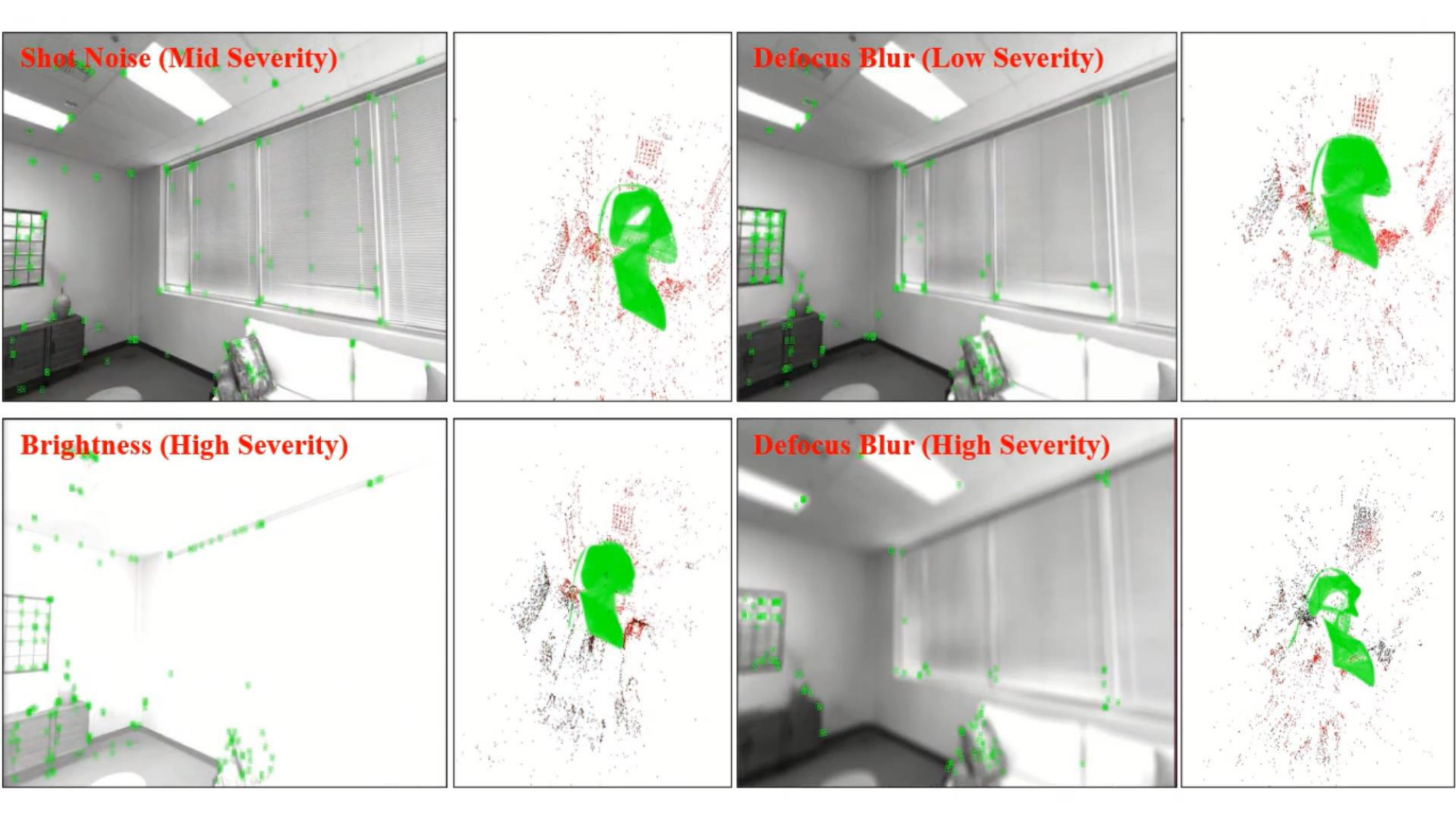} 
\caption{\textbf{Qualitative results of the successful cases of ORBSLAM3 model~\cite{orbslam3} with the RGBD input.}}
\label{fig:orbslam3-succeed} 
\end{figure*}

\begin{figure*}[ht!]
	\centering
\includegraphics[width=\textwidth]{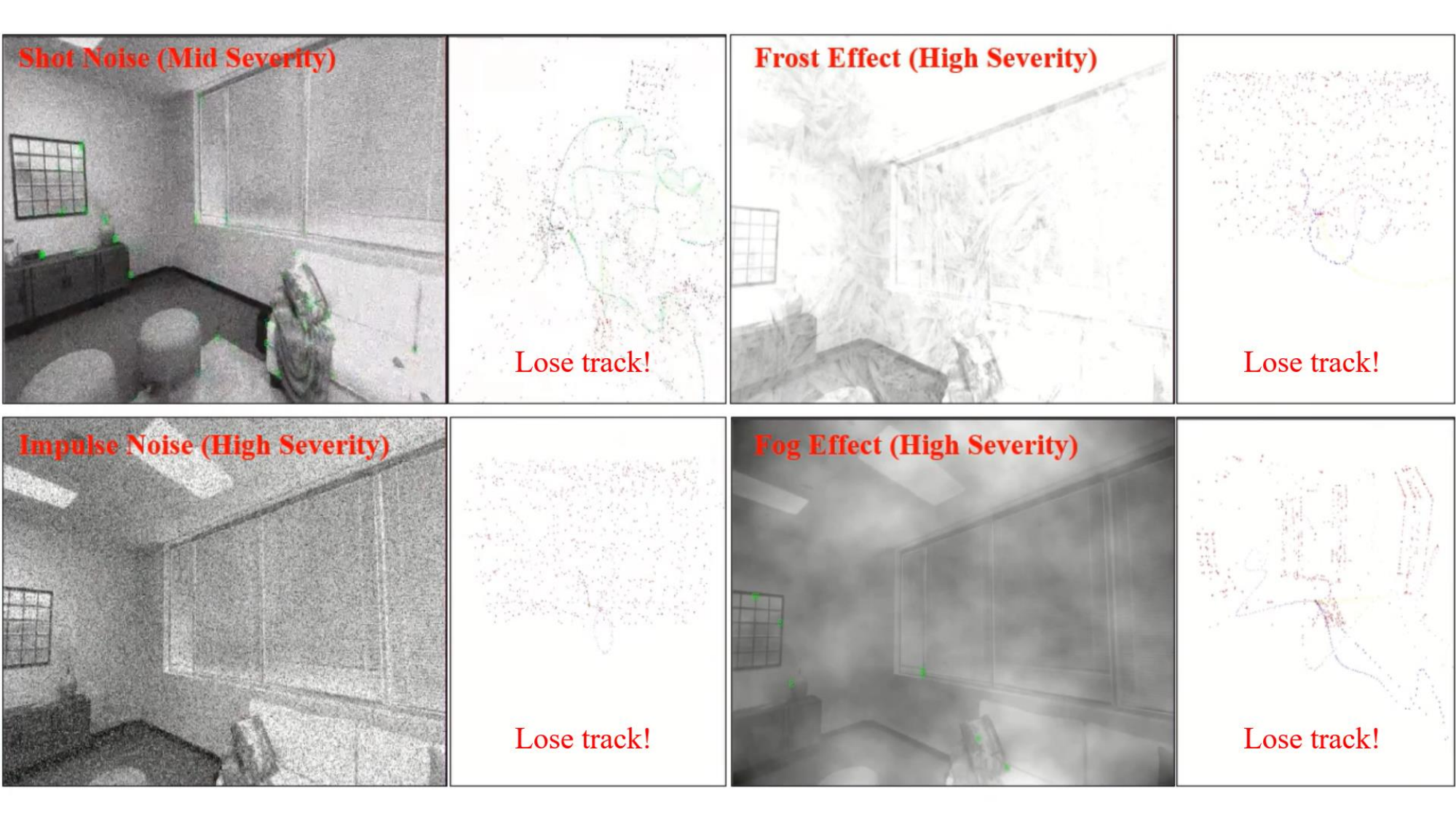} 
\caption{\textbf{Qualitative results of the failure cases of ORBSLAM3 model~\cite{orbslam3} with the RGBD input.}}
\label{fig:orbslam3-fail} 
\end{figure*}

\noindent\textbf{ORB-SLAM3.} Fig.~\ref{fig:orbslam3-succeed} demonstrates the resilience of the ORB-SLAM3~\cite{orbslam3} model to certain image corruptions, \textit{e.g.}, brightness changing and defocus blur. However, we observed that noise-related perturbations can cause the failure of ORB feature detection of the ORB-SLAM3 model, resulting in complete loss of tracking, as depicted in Fig.~\ref{fig:orbslam3-fail}.

\begin{figure*}[ht!]
	\centering
\includegraphics[width=\textwidth]{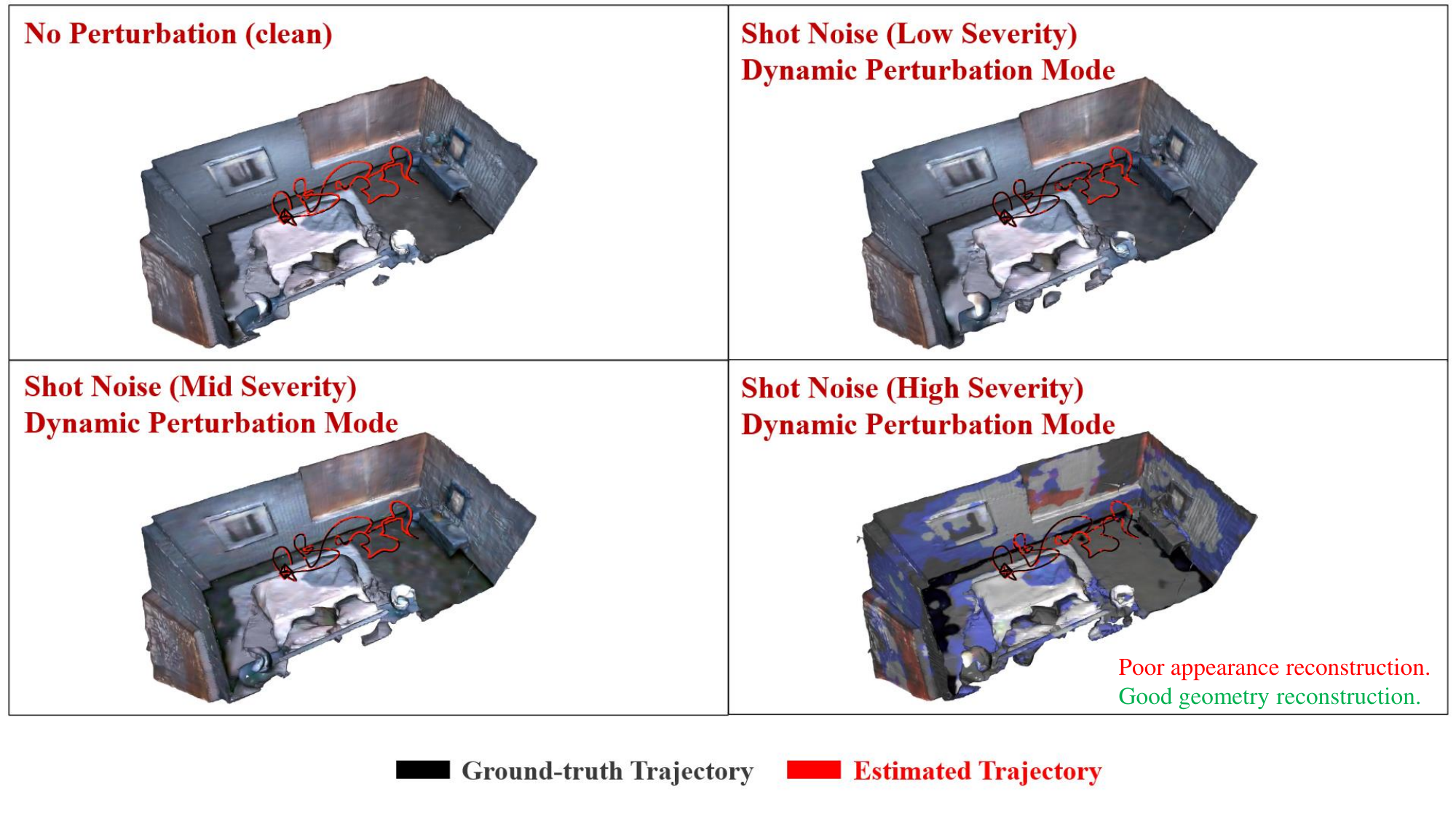} 
\caption{\textbf{Qualitative results of successful cases of Nice-SLAM model~\cite{niceslam} with the RGBD input.}}
\label{fig:niceslam-succeed} 
\end{figure*}

\begin{figure*}[ht!]
	\centering
\includegraphics[width=\textwidth]{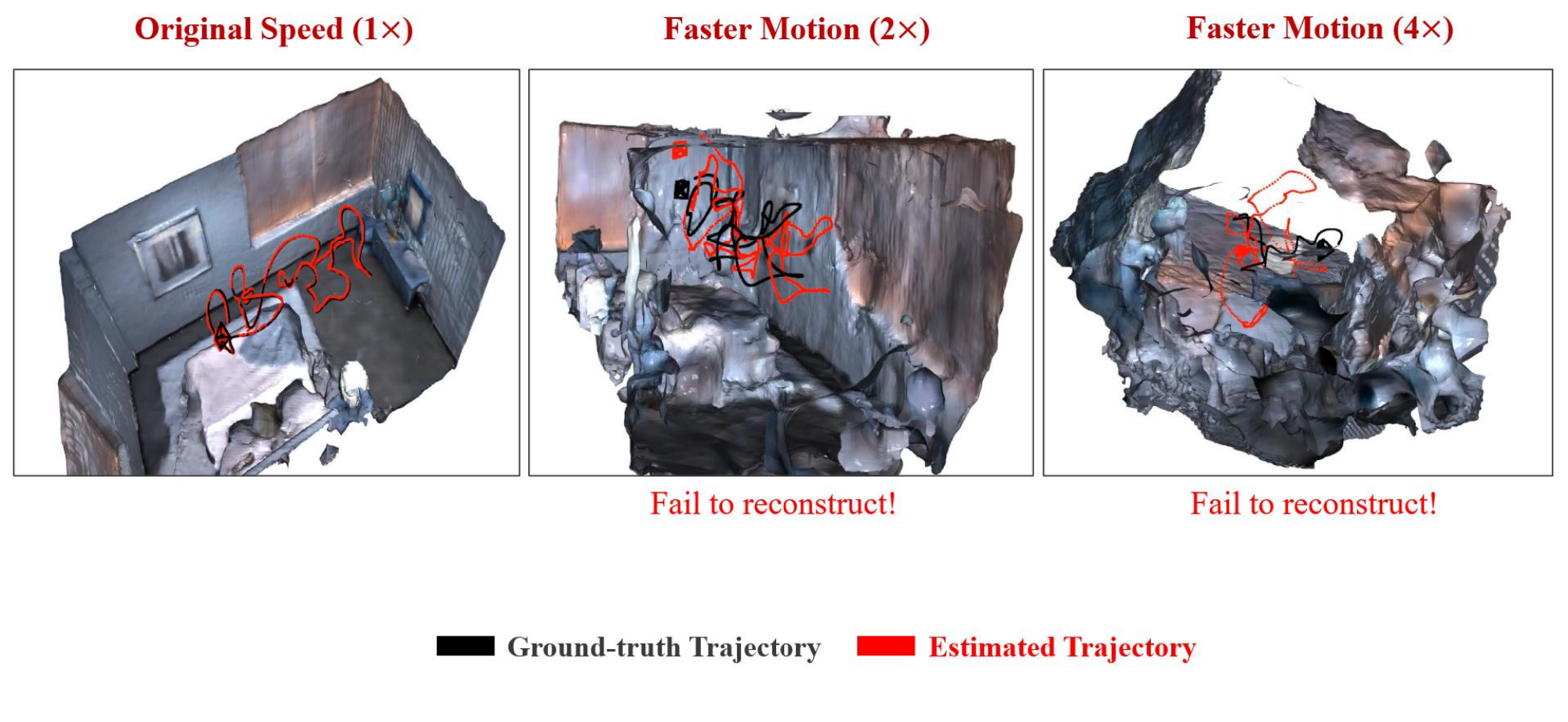} 
\caption{\textbf{Qualitative results of the failure cases of Nice-SLAM model~\cite{niceslam} with the RGBD input.}}
\label{fig:niceslam-fail} 
\end{figure*}

\noindent\textbf{Nice-SLAM.} Fig.~\ref{fig:niceslam-succeed} showcases the 3D reconstruction and trajectory estimation results of the Nice-SLAM~\cite{niceslam} model under varying levels of shot noise perturbation on the RGB image. Nice-SLAM consistently produces high-quality geometry reconstructions even when subjected to high severity levels of shot noise, which we attribute to the nearly error-free, unperturbed depth map aiding geometry reconstruction. However, we observe that the model struggles to accurately predict and reconstruct appearance details. Consequently, as the noise in the RGB images intensifies, color reconstruction quality diminishes. Furthermore, Fig.~\ref{fig:niceslam-fail} highlights the complete failure of the Nice-SLAM model in reconstructing 3D geometry and maintaining tracking when exposed to rapid motion.

\begin{figure*}[ht!]
	\centering
\includegraphics[width=\textwidth]{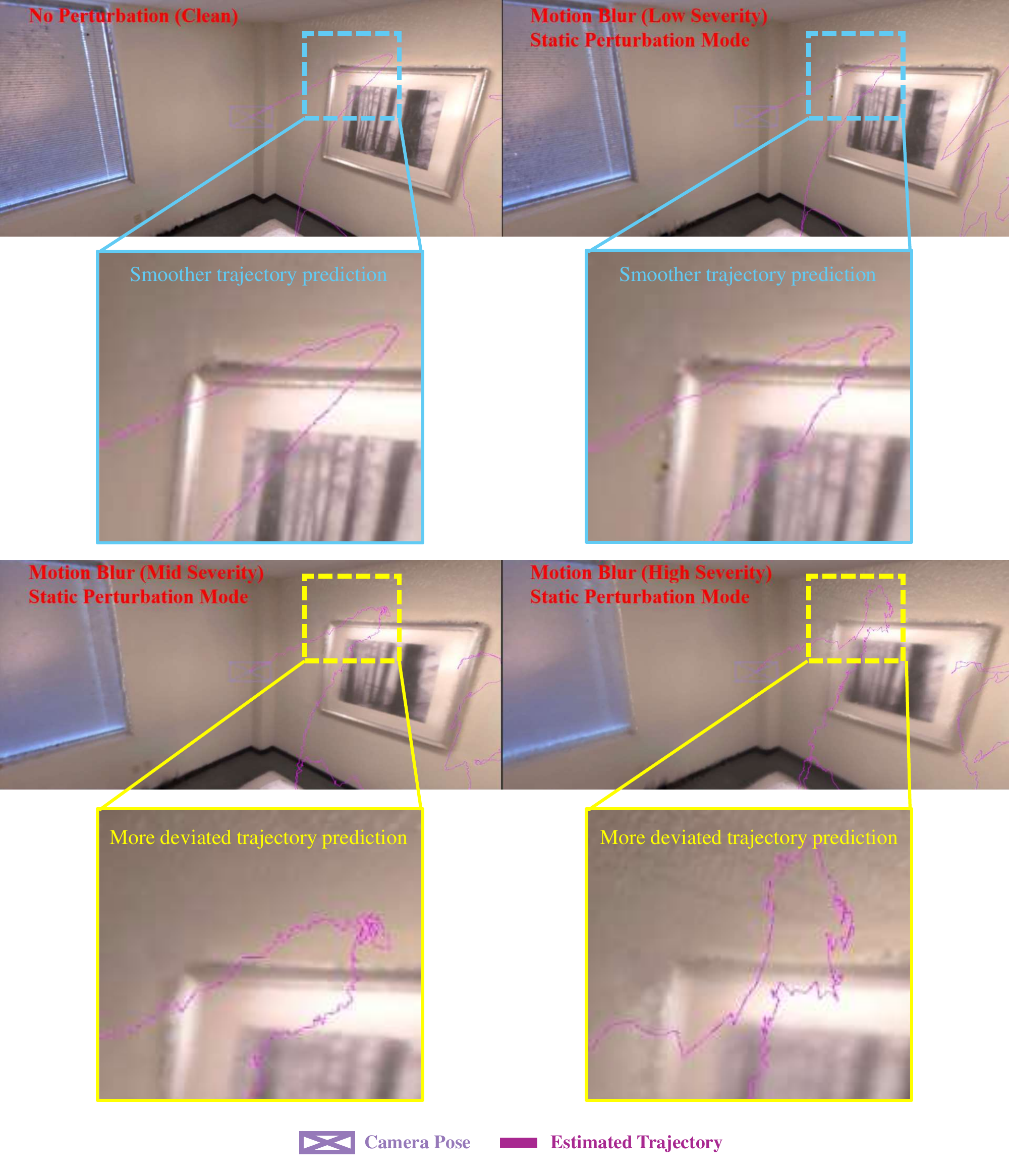} 
\caption{\textbf{Qualitative results of successful cases of SplaTAM-S model~\cite{keetha2023splatam} with the RGBD input.}}
\label{fig:splatam-succeed} 
\end{figure*}

\begin{figure*}[ht!]
	\centering
\includegraphics[width=\textwidth]{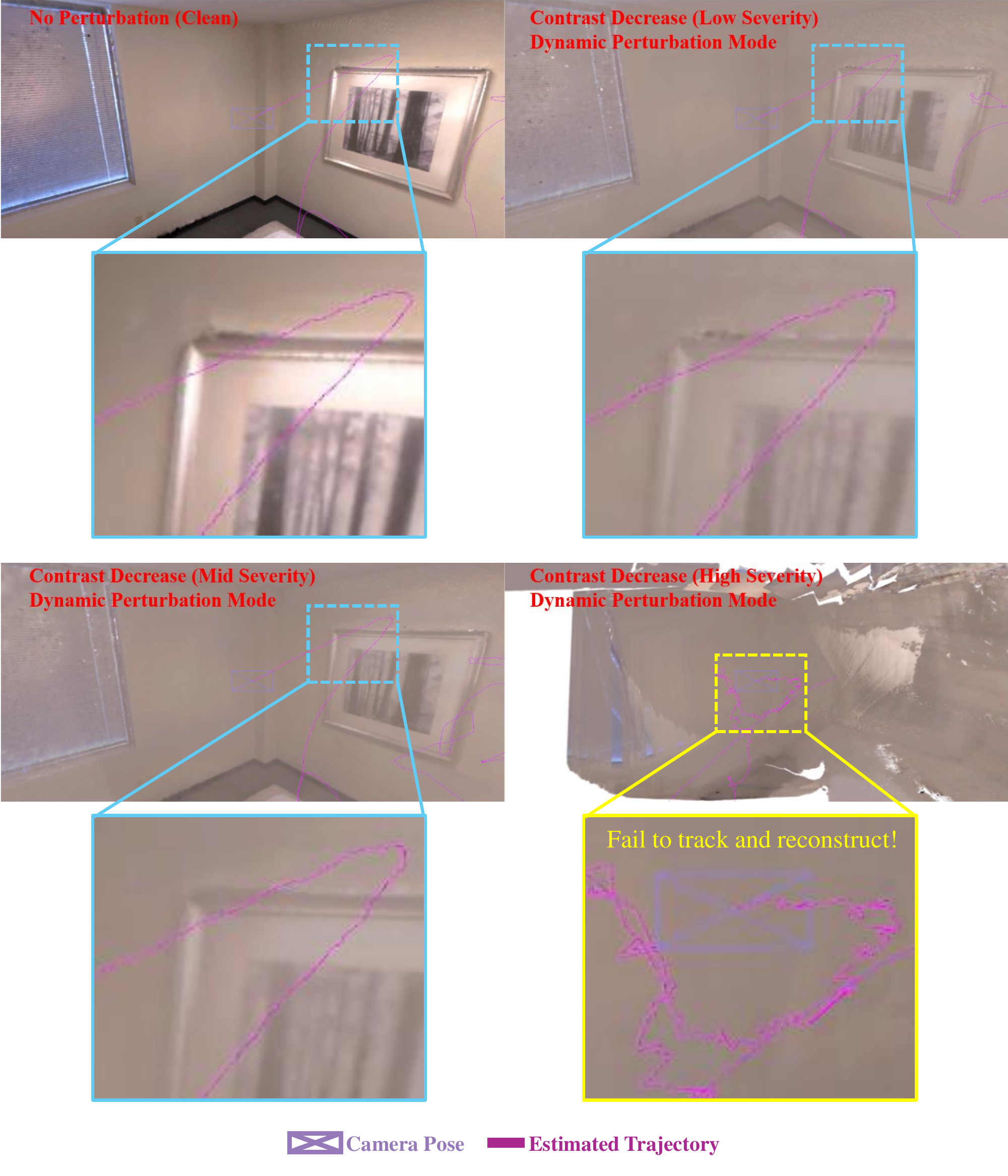} 
\caption{\textbf{Qualitative results of the failure cases of SplaTAM-S model~\cite{niceslam} with the RGBD input.}}
\label{fig:splatam-fail} 
\end{figure*}

\begin{figure*}[ht!]
	\centering
\includegraphics[width=\textwidth]{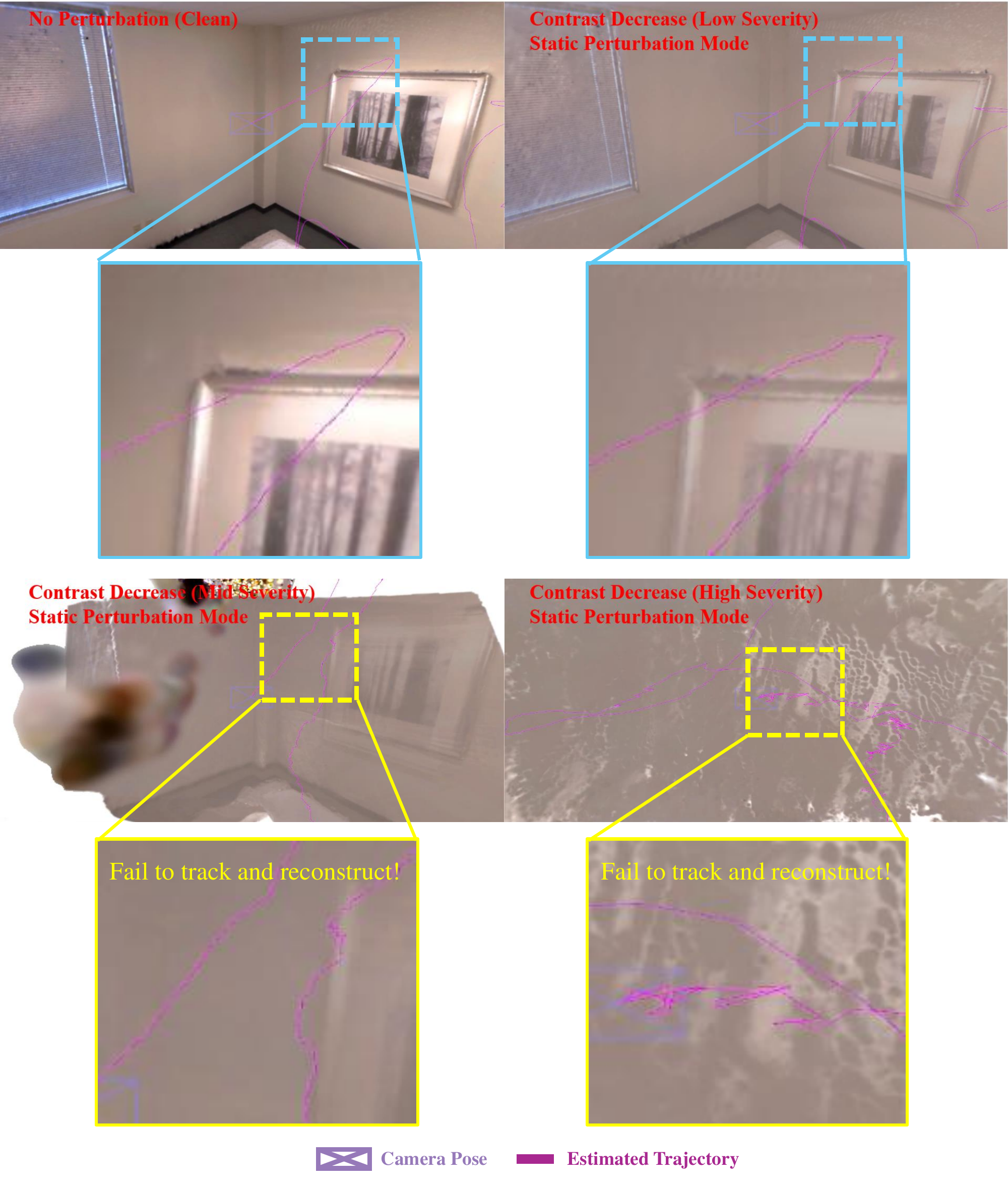} 
\caption{\textbf{Qualitative results of the failure cases of SplaTAM-S model~\cite{niceslam} with the RGBD input.}}
\label{fig:splatam-fail-2} 
\end{figure*}

\noindent\textbf{SplaTAM-S.} Fig.~\ref{fig:splatam-succeed} presents the qualitative results of the SplaTAM-S~\cite{keetha2023splatam} model under different severity levels of motion blur image-level perturbations. The trajectory estimation reveals that, in the absence of perturbation or with low levels of motion blur, the model produces smooth trajectories. However, as perturbation severity increases to a moderate or high level, the predicted trajectory exhibits more deviations. Notably, the 3D reconstruction consistently maintains high quality despite the increasing blurring caused by observation degradation. In addition, Fig.\ref{fig:splatam-fail} and Fig.\ref{fig:splatam-fail-2} depict failure instances of the SplaTAM-S~\cite{keetha2023splatam} model. These failures occur when subjected to varying levels of contrast decrease image-level perturbation under both static and dynamic perturbation modes. Higher severity levels of perturbation result in complete tracking loss and reconstruction failure.

To provide a better viewing experience, we kindly refer you to the attached video demo.

\section{More Future Directions for Exploration}\label{sec:more future work}

\noindent\textbf{Robustness evaluation in unbounded 3D scene.} Our work does not encompass the robustness of SLAM systems in unbounded scenes, such as outdoor environments~\cite{dosovitskiy2017carla,zhao2023subt}. Investigating the robustness of SLAM systems in such scenarios holds significant potential for future research. It can contribute to a better understanding of the practical applicability and generalizability of SLAM systems in complex scenes.

\noindent\textbf{Computationally-efficient robustness evaluation.} Our findings have identified discernible indicators within certain SLAM models that can reflect degraded observations, \textit{i.e.}, the reconstruction quality of RGB images and depth maps for the SplaTAM-S~\cite{keetha2023splatam} model. Future research could explore leveraging and designing robustness indicators to evaluate the robustness of SLAM systems more efficiently. By incorporating such indicators, we have the potential to enable unsupervised performance evaluation of SLAM systems, especially in scenarios where obtaining ground-truth annotations is challenging or costly.

\noindent\textbf{Real-world robustness evaluation.} While our work primarily focuses on synthesis-based robustness analysis, we recognize the value of real-world verification and validation for SLAM systems. Conducting extensive field tests in more challenging environments~\cite{SubT}, where SLAM systems are subjected to agile locomotion types~\cite{kaufmann2023champion}, would provide empirical validation of simulation results and uncover additional challenges. This real-world evaluation would bridge the gap between simulated environments and actual deployment conditions, ensuring the practical reliability and robustness of SLAM systems.

\ifCLASSOPTIONcaptionsoff
  \newpage
\fi
\bibliographystyle{plainnat}
\bibliography{main}
\end{document}